\documentclass[lettersize,journal]{IEEEtran}

\usepackage{times}
\usepackage{graphicx}

\usepackage{siunitx}
\usepackage{amsmath, amsfonts}
\usepackage{amssymb}
\usepackage{algorithm}
\usepackage{algorithmic}
\usepackage{multirow}
\usepackage{bm}
\usepackage{float}
\usepackage{lettrine}
\usepackage{stfloats}
\usepackage{cite}
\usepackage[utf8]{inputenc} 
\usepackage[T1]{fontenc}    
\usepackage{hyperref}       
\usepackage{url}            
\usepackage{booktabs}       
\usepackage{amsfonts}       
\usepackage{nicefrac}       
\usepackage{microtype}      
\usepackage{xcolor}         
\usepackage{url}
\usepackage{graphicx}
\usepackage{amsthm}
\usepackage{booktabs}
\usepackage{makecell}
\usepackage[normalem]{ulem}
\usepackage{listings}
\usepackage{pifont}
\usepackage{subcaption}
\usepackage{enumitem}

\usepackage{wrapfig}
\usepackage[numbers]{natbib}

\newcommand{\rvtwo}[1]{#1}

\begin{document}
\title{PatchAD: A Lightweight Patch-Based MLP-Mixer for Time Series Anomaly Detection}

%


\author{Zhijie Zhong, Zhiwen Yu\IEEEauthorrefmark{1},~\IEEEmembership{Senior Member~IEEE}, Yiyuan Yang, Weizheng Wang~\IEEEmembership{Member~IEEE},\\
Kaixiang Yang,~\IEEEmembership{Member~IEEE}, C. L. Philip Chen,~\IEEEmembership{Fellow IEEE}
\IEEEcompsocitemizethanks{
\IEEEcompsocthanksitem Zhijie Zhong is with the School of Future Technology, South China University of Technology, Guangzhou, Guangdong 510650, China, and also with the Pengcheng Laboratory, Shenzhen, Guangdong 518066, China.
\IEEEcompsocthanksitem  Zhiwen~Yu is with the School of Computer Science and Engineering, South China University of Technology, Guangzhou, Guangdong 510650, China, and also with the Pengcheng Laboratory, Shenzhen, Guangdong 518066, China. Email: zhwyu@scut.edu.cn. Telephone number: 86-20-62893506. Fax number: 86-20-39380288. 
\IEEEcompsocthanksitem Yiyuan Yang, from the Department of Computer Science at the University of Oxford, Oxford, UK.
\IEEEcompsocthanksitem Weizheng Wang is currently a Postdoctoral Research Fellow in the Department of Electrical and Electronic Engineering at the Hong Kong Polytechnic University, Hong Kong SAR, China.
\IEEEcompsocthanksitem Kaixiang Yang are C. L. Philip Chen are with the School of Computer Science and Engineering, South China University of Technology, Guangzhou, Guangdong 510650, China.
\IEEEcompsocthanksitem \IEEEauthorrefmark{1}Corresponding author: Zhiwen Yu.
}
}

\markboth{IEEE Transactions on Big Data}
{XXX \MakeLowercase{\textit{et al.}}: IEEE Transactions on Big Data}





\maketitle

\begin{abstract}
%
  Time series anomaly detection is a pivotal task in data analysis, yet it poses the challenge of discerning normal and abnormal patterns in label-deficient scenarios. While prior studies have largely employed reconstruction-based approaches, which limit the models' representational capacities. Moreover, existing deep learning-based methods are not sufficiently lightweight. Addressing these issues, we present PatchAD, our novel, highly efficient multiscale patch-based MLP-Mixer architecture that utilizes contrastive learning for representation extraction and anomaly detection. With its four distinct MLP Mixers and innovative dual project constraint module, PatchAD mitigates potential model degradation and offers a lightweight solution, requiring only \textbf{0.403M} parameters. Its efficacy is demonstrated by state-of-the-art results across \textbf{8} datasets sourced from different application scenarios, outperforming over \textbf{30} comparative algorithms. PatchAD significantly improves the classical F1 score by \textbf{6.84\%}, the Aff-F1 score by \textbf{4.27\%}, and the V-ROC by \textbf{2.49\%}. 
  Simultaneously, an in-depth analysis of the mechanisms underlying PatchAD has been conducted from both theoretical and experimental perspectives, validating the design motivations of the model.
  The code is publicly available at   \url{https://github.com/EmorZz1G/PatchAD}.
    
\end{abstract}

\begin{IEEEkeywords}
Time series, anomaly detection, data mining, lightweight, MLP.
\end{IEEEkeywords}

\IEEEdisplaynotcompsoctitleabstractindextext

%
\IEEEpeerreviewmaketitle


\section{Introduction}
\IEEEPARstart{T}ime series anomaly detection (TSAD) is a pivotal aspect of data mining, focused on identifying abnormal patterns within time series data that significantly deviate from expected behaviour \cite{tbd1,tbd2,tbd3,adamembls}. 
In the era of big data, with the rapid advancement of large-scale sensing technologies and the continuous enhancement of storage capabilities, TSAD has been widely applied in diverse sectors. 
For instance, the Internet of Things (IoTs) serve to monitor abnormal events in sensor data \cite{yang2021early,yang2021pipeline}. 
Unsupervised algorithms, capable of operating without labels, have gained wide popularity \cite{bls2}. These include reconstruction-based \cite{LSTM_VAE,OmniAnomaly}, density-based \cite{DAGMM}, contrastive learning \cite{dc_detector,cont1}, autoregression-based \cite{VAR,LSTM_RNN,timesnet}, large language models (LLMs) \cite{gpt2} \textit{etc}.
However, extracting anomalies from extensive and complex temporal data presents significant challenges for different algorithms. 
\textbf{Challenge 1:} Accurately defining anomaly representations is challenging due to real-world anomalies' diverse and dynamic nature. Effective anomaly detection requires models that can learn the inter-relationships between various data channels. 
\textbf{Challenge 2}: There is still significant potential for improvement in terms of training and inference speed, model complexity, and the lack of lightweight features.
\textbf{Challenge 3}: In an unsupervised setting, the challenge lies in establishing a reliable basis to model normal and anomalous data. 
Specifically, several algorithms \cite{anomaly_trans,dc_detector,v1_cont_trans,v1_USAD} have begun incorporating contrastive learning to enhance model representational capabilities. Nonetheless, these approaches often overlook the design of network structures and model degradation issue of contrastive learning. 
For instance, \citet{anomaly_trans} devised a contrastive learning approach that leverages prior correlation and series correlation mining to elucidate the relationship between normal and anomalous patterns. However, this necessitated the employment of a maximization-minimization optimization strategy to train the model and prevent its degradation. Similarly, \citet{dc_detector} extended the paradigm of image contrastive learning established by BYOL \cite{v1_byol} by designing different temporal views to circumvent the challenges associated with negative sample construction in contrastive learning. Nonetheless, there exists the potential for performance degradation of the model when the patch settings are not appropriately configured.

Additionally, they either do not explicitly model inter-channel dependencies \cite{dc_detector} or rely solely on basic MLPs or Transformers that focus on temporal information \cite{v1_cont_trans}, mainly considering point-level features \cite{v1_USAD}. 
However, with the advancement of TSAD evaluation metrics, an increasing number of interval-based evaluation indicators \cite{VUS,AFFI,PATE,TSB-AD} have garnered research attention. Nevertheless, prior studies predominantly concentrated on the F1 with point ajustment \cite{anomaly_trans}, thereby neglecting the detection of interval anomalies. Consequently, there remains considerable potential for enhancing the detection capabilities for interval anomalies in current research \cite{TSB-AD}.

Furthermore, the previous TSAD algorithms based on contrastive learning \cite{anomaly_trans,dc_detector,cont1,v1_cont_trans} lack in-depth exploration from a theoretical perspective and fail to provide more valuable intuitions for future research.
In contrast, we find that normal time series data points exhibit reduced information entropy, indicating higher similarity and correlation among them, whereas anomalies show increased entropy due to difficulties in establishing connections with other points. This difference aids our model in detecting anomalies, ensuring that normal data maintains consistent representations across different views, whereas anomalies do not. This fundamental principle forms the basis of our proposed anomaly detection method.
Simultaneously, to further explore this concept, we conducted both experimental and theoretical analyses to investigate the underlying principles of this approach, thereby addressing the challenges from various perspectives.

In response to the above challenges and principles, we introduce \textbf{\underline{PatchAD}}, a lightweight \textbf{\underline{Patch}}-based MLP architecture for time series \textbf{\underline{A}}nomaly \textbf{\underline{D}}etection.
Specifically, proposed \textit{Multi-scale Patching and Embedding} decomposes the input data along the temporal dimension into non-overlapping patches, extracting semantically richer features beyond simple point-level characteristics. Besides, various submodels address various time scales, each focusing on different patch sizes. 
\textit{Patch Mixer Encoders} are introduced to learn the complex relationships within and between patches and channels. A shared \textit{MixRep Mixer} uniformly represents the feature space, while a \textit{Dual Project Constraint} prevents the model from adopting overly simplistic solutions, thereby enhancing its representational capabilities. 
At the same time, the lightweight model configuration enables PatchAD to better tackle Challenge 2, while the intricate encoder design enhances the model's feature learning capacity to confront Challenge 1.

In detail, this work makes the following contributions:
\begin{figure*}[t]
    \centering
    \begin{minipage}[b]{0.7\linewidth}
        \centering
        \includegraphics[width=\linewidth]{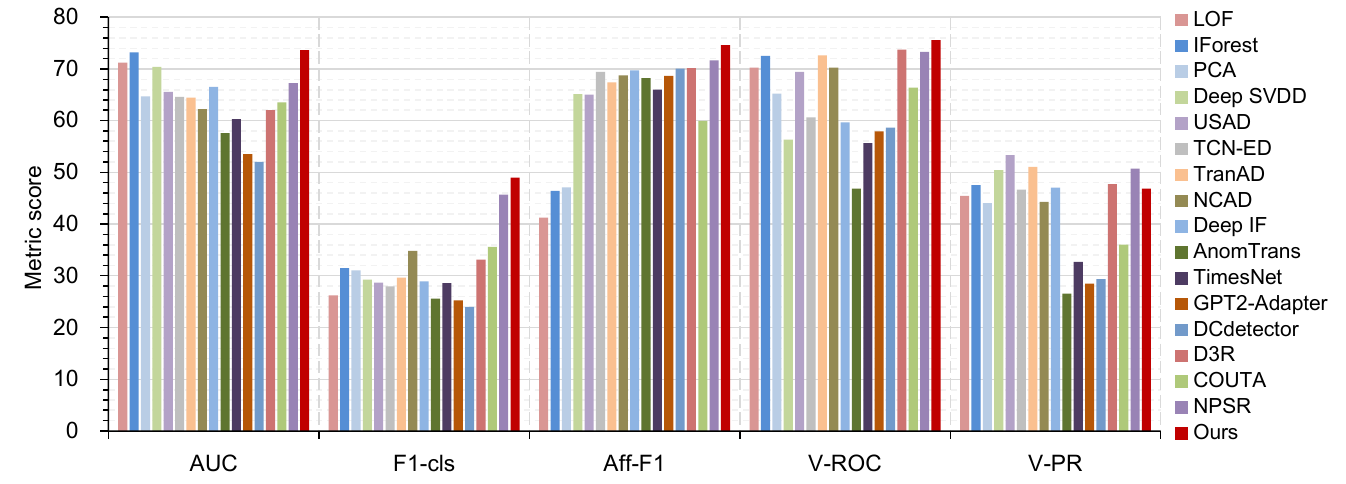}
        \caption{Overall performances under different metrics.}
        \label{sfig:overall_metric}
    \end{minipage}
    \begin{minipage}[b]{0.29\linewidth}
        \centering
        \includegraphics[width=\linewidth]{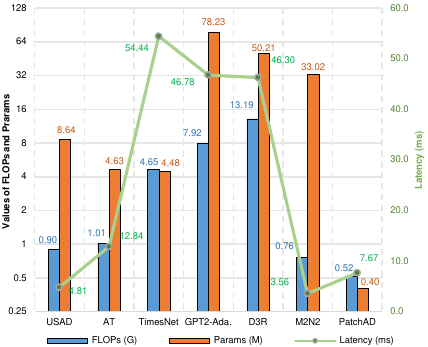}
        \caption{Model Params, FLOPs, and Latency.}
        \label{sfig:model_size}
    \end{minipage}
\end{figure*}

(1) \textbf{State-of-the-art Anomaly Detection} (Figure\ref{sfig:overall_metric}): We propose a novel contrastive learning-based time series anomaly detection method. PatchAD demonstrates exceptional performance across eight benchmark datasets, exceeding 30 baselines across a range of evaluation metrics, including commonly used metrics and interval anomaly indicators.

(2) \textbf{Lightweight and Efficient Design} (Figure\ref{sfig:model_size}): PatchAD employs a multi-scale patch-based MLP-Mixer architecture with different novel strategies, achieving remarkable efficiency and compactness compared to Transformer-based models.
Moreover, we conducted an analysis of the model's scalability from both experimental and theoretical perspectives.

(3) \textbf{Comprehensive Analysis}: An extensive ablation study validates the effectiveness of each component of PatchAD, supported by practical and theoretical analyses that elucidate the model's underlying principles and design choices.

\section{Preliminaries}
\subsection{Problem Formulation}
In this paper, we consider a multivariate time series of length $T$ in an unsupervised setting:
\(\mathcal{X} = (x_1, x_2, \cdots, x_T)\),
$x_t \in \mathbb{R}^C$ represents a C-dimensional channel feature, such as data from C sensors or machines.
The objective is to train a model using $\mathcal{X}_{train}$ and provide predictions $\mathcal{Y}_{test} = (y_{1}, y_{2}, \cdots, y_{T^{\prime}})$ on another sequence $\mathcal{X}_{test}$ of length $T^{\prime}$. Here, $y_t \in \{ 0, 1\}$, where 1 signifies an anomaly at that time point and 0 indicates a normal point.
\subsection{Time Series Anomaly Detection}
\begin{figure*}[t]
    \centering
\includegraphics[width=0.9\linewidth]{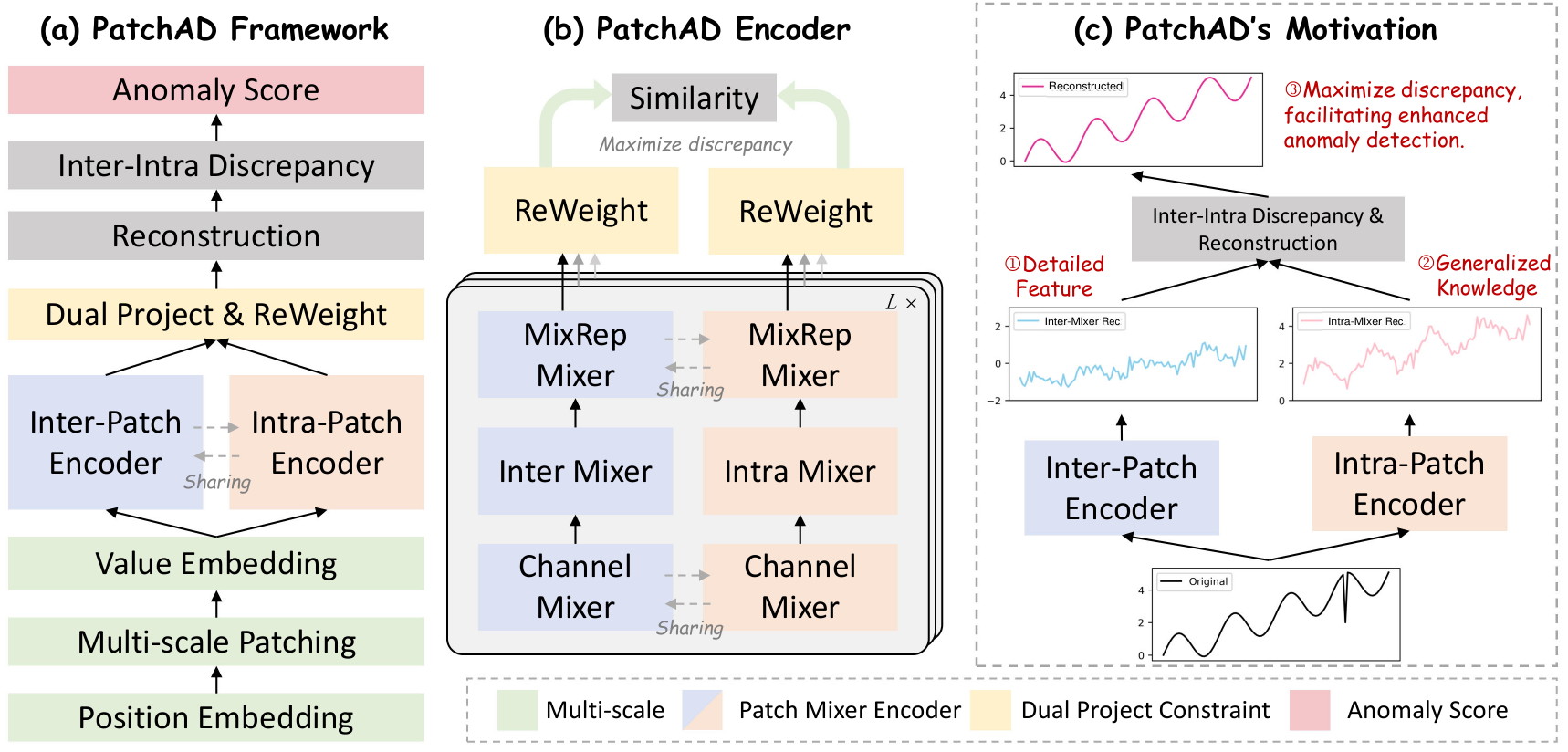}
    \caption{The workflow of the proposed PatchAD framework.}
    \label{fig:overall}
\end{figure*}
The time series anomaly detection methods encompass statistical approaches, classical machine learning methods, and deep learning techniques \cite{v1_GRELEN}. Statistical methods involve moving averages and the autoregressive integrated moving average (ARIMA) model. Classical machine learning methods encompass classification-based approaches like One-Class Support Vector Machine (OCSVM) \cite{OCSVM} and Support Vector Data Description (SVDD) \cite{SVDD}. 
However, machine learning methods often struggle to extract meaningful features and detect anomalies when confronted with high-dimensional, complex, and voluminous time series data, ultimately resulting in suboptimal performance \cite{adamembls,patch_bls}.

Deep learning methods include OmniAnomaly \cite{OmniAnomaly}, USAD \cite{v1_USAD}, LSTM-VAE \cite{LSTM_VAE}, and DeepAnT \cite{v1_DeepAnT}, which are trained on normal data and detect anomalies by inference using reconstruction errors. 
They often rely on reconstruction assumptions, wherein the model's ability to restore anomalous data diminishes after being trained on normal data. Previous studies \cite{uniad,membls} have indicated that this prior becomes less applicable when the model's reconstruction capability is excessively strong, leading to a decline in anomaly detection performance.

Some techniques have leveraged self-attention mechanisms and have demonstrated promising detection results \cite{trans1}. While \cite{v1_couta} designed six different anomaly injection methods to train the model in a self-supervised approach, \cite{v1_ncad} utilized outlier exposure and mixup to construct a self-supervised task. 
However, due to the distributional discrepancies between real and synthetic data, this results in subpar performance when confronted with real-world data, as observed in \cite{v1_couta,v1_ncad}.

Recognizing the limitations imposed by the reconstruction-based assumption, research efforts like \cite{sisvae} introduced sequence variational autoencoder model based on smoothness constraints, whereas \cite{anomaly_trans} proposed prior and sequence associations for temporal modeling, both achieving enhanced performance. 
Recently approaches based on diffusion models and Large language model have also been proposed \cite{yang2024surveyDiffusion,gpt2}.
Although significant advancements have been made in the performance of time series anomaly detection methods, most of these approaches are based on Transformers or sophisticated attention models \cite{anomaly_trans,dc_detector,yang2024surveyDiffusion,gpt2}. This reliance results in models that are not sufficiently lightweight and exhibit slower computational efficiency, rendering them unsuitable for certain real-world scenarios that require rapid detection.
\subsection{MLP Mixer \& Contrastive Learning for Time Series Analysis}
\textbf{The MLP Mixer} initially proposed as a novel architecture for computer vision \cite{v1_mlpmix}, has gained significant attention in recent years. Recently, there has been analysis indicating that MLP Mixers can effectively handle sequential data and find applications in other domains, such as time series prediction. Specifically, works like \cite{tsmixer,patchtst,v1_patchmixer,yi2024frequency} have focused on prediction tasks, particularly for multi-step time series prediction, leveraging the capabilities of MLP Mixers to achieve improved forecasting results. 
For example, PatchTST \cite{patchtst} employs a single-scale patch-based MLP as its core architecture, whereas the work in \cite{yi2024frequency} applies an MLP to capture frequency-domain information in time series data for predictive modeling.
Their experiments demonstrate that the MLP Mixer surpasses existing Transformer-based and LSTM-based methods in long sequence time series forecasting (LSTF). 
Constrained by space, further detailed descriptions about Mixer are provided in Appendix \ref{app:mixer}.



\textbf{Contrastive learning} aims to drive the model to bring positive pairs closer together in the learned representation space while pushing apart negative pairs \cite{v1_byol,v1_simsiam,cont1}. Additionally, methods like BYOL \cite{v1_byol} and SimSiam \cite{v1_simsiam} have demonstrated advanced performance without explicitly constructing negative pairs. Recently, studies such as \cite{cont1} and \cite{cont2} have explored integrating contrastive learning into time series representation learning. TS2Vec \cite{TS2Vec} performs contrastive learning in a hierarchical way over augmentation views. \cite{anomaly_trans,dc_detector}, on the other hand, combine self-attention with contrastive learning, achieving state-of-the-art results in time series anomaly detection.
In consideration of space limits, technical details are provided in Appendix \ref{app:contrastive}.

\section{Proposed Method}
\subsection{Overall Architecture}
Figure \ref{fig:overall} illustrates the overall structure of PatchAD, which adopts patch sizes at multiple scales. We exemplify its architecture using a single scale. PatchAD is an $L$-layer network, where each layer comprises distinct modules: MLP Mixers, Project head, and ReWeight, forming its structure. We have devised four distinct types of MLP Mixers: Channel Mixer, Inter Mixer, Intra Mixer, and MixRep Mixer, each dedicated to learning various channels, inter-patch, intra-patch relationships, and a unified representation space. The Project head serves to prevent the model from converging to trivial solutions, while the ReWeight module manages the weights across different layers.

The focal point of our design lies in the four distinct MLP Mixers. We utilize various MLP Mixers to learn different representations from the input. Normal points share a common latent space across different views, while anomalies, being scarce and lacking specific patterns, struggle to share a coherent representation with normal points. Consequently, normal points exhibit minor discrepancies across different views, whereas anomalies showcase larger variations. PatchAD leverages inter-intra discrepancy to model this relationship.

Our anomaly detection approach, PatchAD, leverages the principle that normal time series points, characterized by low information entropy, exhibit consistent representations across diverse perspectives. In contrast, anomalies with higher entropy struggle to maintain such consistency (Figure \ref{fig:overall}(c)). PatchAD employs an Inter-Patch Encoder to capture detailed features and an Intra-Patch Encoder to learn generalized trends. This combined approach enables a comprehensive understanding of normal and anomalous representations, achieving effective anomaly detection.

\subsection{Multi-Scale Patching and Embedding}
To enhance the temporal contextual representation in time series, we incorporated Transformer positional encoding. Acknowledging the significance of different sequence lengths in the time series analysis, we introduced \textit{Multi-scale Patching} in PatchAD. Patching divides the time series window into smaller patches \cite{patchtst}. Multi-scale Patching enables PatchAD to focus on features of varying sequence lengths. Moreover, incorporating multi-scale information helps compensate for information loss during the patching process. The input data with $C$ channels and length $T$ is transformed into patches of size $P$. It can be seen as dividing the time series data into $N$ non-overlapping patches blocks, called $\operatorname{Patching}(\cdot)$ \cite{patchtst}. Thus, the original data of dimensions $C \times T$ is transformed into $C \times N \times P$.
\begin{equation}
    \mathcal{X} = \operatorname{Patching}(\operatorname{PE}(\mathcal{X})),\ (\mathbb{R}^{C \times T} \rightarrow \mathbb{R}^{C \times N \times P}),
\end{equation}
where $\operatorname{PE}(\cdot)$ is positional embedding \cite{anomaly_trans}.
PatchAD employs two value embeddings, \textit{i.e.}, $\operatorname{VE}(\cdot)$, facilitating the features from two distinct perspectives within the framework. 
To make feature extraction and learning easier, these are used to embed the intra-patching and inter-patching information into the \(D\)-dimensional space, respectively.
And $\operatorname{VE}(\cdot)$ is a single linear layer.
\begin{equation}
\begin{aligned}
    \mathcal{N}&= \operatorname{VE}(\mathcal{X}),\ (\mathbb{R}^{C \times N \times P} \rightarrow \mathbb{R}^{C \times N \times D}), \\
    \mathcal{P}&= \operatorname{VE}(\mathcal{X}),\ (\mathbb{R}^{C \times N \times P} \rightarrow \mathbb{R}^{C \times P \times D}).
\end{aligned}
\end{equation}
For the sake of brevity and clarity, the above describes the process of single-scale patching. In the last part of this chapter, we describe how multi-scale patching is implemented (see Sec. \ref{sec:multi_scale}).


\subsection{Patch Mixer Encoder}
PatchAD comprises $L$ layers of Patch Mixer layers. This encoder incorporates four distinct MLP Mixers to extract features across different dimensions. 
The Multi-Layer Perceptron (MLP) consists of two fully connected (FC) layers, a normalization layer and a ReLU activation, and utilizes a residual connection from input to output. The MLP can be represented as:
\begin{equation}
    \operatorname{MLP}(\mathcal{X}) = \operatorname{FC}(\operatorname{ReLU}(\operatorname{FC}(\operatorname{Norm}(\mathcal{X}))))+\mathcal{X}.
    \label{eq:mlp}
\end{equation}
The MLP Mixers of different types utilize MLP to model the feature variations along different dimensions, which can be represented as:
\begin{equation}
\begin{aligned}
    &\operatorname{Mixer}(\mathcal{X},dim = i)\\
    &=\operatorname{Transpose}(\operatorname{MLP}(\operatorname{Transpose}(\mathcal{X},\text{dim = i})), \text{dim = i}), 
\end{aligned}
\end{equation}
where $\operatorname{Transpose}(\cdot, \text{dim = i})$ denotes the operation of swapping the $i$-th dimension with the last dimension. The purpose of this is to enable the MLP to effectively capture the features along different dimensions.
The following are the different types of MLP Mixers utilized in PatchAD:
\begin{enumerate}[itemsep=0pt,parsep=3pt]
    \item \textit{Channel Mixer} captures correlations among different channels.
    \item \textit{Inter Mixer} captures inter-patch representations, improving detailed expressions.
    \item \textit{Intra Mixer} captures intra-patch representations, improving generalized expressions.
    \item \textit{MixRep Mixer} embeds two distinct views into the same representation space.
\end{enumerate}
The diagram depicts PatchAD comprised of an inter-patch encoder and an intra-patch encoder, generating representations denoted as $\mathcal{N} \in \mathbb{R}^{C \times N \times D}$ and $\mathcal{P} \in \mathbb{R}^{C \times P \times D}$, representing the inter-patch and intra-patch views, respectively. It's important to note that \textit{The Channel Mixer} and \textit{The MixRep Mixer} share weights, while \textit{The Inter Mixer} and \textit{The Intra Mixer} have their own weights. This design facilitates PatchAD in learning differences between views and prevents model degradation.
Specifically, for each layer of the Patch Mixer Encoder, there exist two types of input features: inter-patch, denoted as \(\mathcal{N}\), and intra-patch, represented as \(\mathcal{P}\).

Next, we elaborate on the workflow of four types of mixers in detail.
Initially, these features are processed by the shared-parameter Channel Mixer, expressed as follows:
\begin{equation}
\begin{aligned}
    \mathcal{N}=\operatorname{ChannelMixer}(\mathcal{N},\text{dim=1}),\\
    \mathcal{P}=\operatorname{ChannelMixer}(\mathcal{P},\text{dim=1}),
\end{aligned}
\end{equation}
where \(\text{dim}=1\) indicates that the channel mixer captures the correlations among channels across dimension \(C\). 

Similarly, the Inter Mixer and Intra Mixer utilize the outputs of the channel mixer as their respective inputs, yielding:
\begin{equation}
\begin{aligned}
    \mathcal{N}= \operatorname{InterMixer}(\mathcal{N},\text{dim=2}),\\
    \mathcal{P}=\operatorname{IntraMixer}(\mathcal{P},\text{dim=2}).
\end{aligned}
\end{equation}
Here, \(\operatorname{InterMixer}\) and \(\operatorname{InraMixer}\) denote the functions of the Inter Mixer and Intra Mixer, respectively. It is important to note that although both operate on feature variation at \(\text{dim=2}\), the differing shapes of the input tensors result in \(\mathcal{N}\in \mathbb{R}^{C\times N \times D}\) and \(\mathcal{P}\in \mathbb{R}^{C\times P \times D}\). Consequently, \(\operatorname{InterMixer}\) processes \(\mathcal{N}\) along dimension \(N\) to learn detailed features of patches, whereas \(\operatorname{IntraMixer}\) processes \(\mathcal{P}\) along dimension \(P\) to acquire global knowledge of patches. Secs. \ref{sec:mech} and \ref{sec:theory} provide a detailed experimental and theoretical analysis of the nature of learning in both cases.

Subsequently, to enable the model to learn low-dimensional invariant information from different types of features and thereby enhance the generalization capability of PatchAD, inspired by \cite{v1_simclr}, the MixRep Mixer is employed to embed the two distinct types of features into a unified representation space, expressed as follows:
\begin{equation}
\begin{aligned}
    \mathcal{N}=\operatorname{MixRepMixer}(\mathcal{N},\text{dim=3}),\\
    \mathcal{P}=\operatorname{MixRepMixer}(\mathcal{P},\text{dim=3}),
\end{aligned}
\end{equation}
Analogous to previous instances, \(\text{dim}=3\) signifies that the MixRep Mixer learns information from two views along the final dimension. In order to facilitate the embedding of both types of information into the same space, a shared-parameter Mixer is employed to achieve this objective.
Ultimately, the two branches of the Patch Mixer Encoder can be expressed as:
\begin{equation}
\begin{aligned}
    \mathcal{N} = \operatorname{MixRepMixer}(\operatorname{InterMixer}(\operatorname{ChannelMixer}(\mathcal{N}))),\\
    \mathcal{P} = \operatorname{MixRepMixer}(\operatorname{IntraMixer}(\operatorname{ChannelMixer}(\mathcal{N}))).\\
\end{aligned}
\end{equation}
In summary, the four mixer workflows correspond to Figure \ref{fig:overall}(b).
In the end, two separate simple MLPs, \textit{e.i.}, \(\operatorname{Rec_1}\) and \(\operatorname{Rec_2}\) are utilized to reconstruct the original data, which are then combined.
\begin{equation}
\begin{aligned}
    \hat{\mathcal{X}}_1 = \operatorname{Rec_1}(\mathcal{N}), \quad \hat{\mathcal{X}}_2 = \operatorname{Rec_2}(\mathcal{P}), \quad
    \hat{\mathcal{X}} = \hat{\mathcal{X}}_1 + \hat{\mathcal{X}}_2.
\end{aligned}
\end{equation}
Simultaneously, the structures of \(\operatorname{Rec_1}\) and \(\operatorname{Rec_2}\) are identical. Initially, the last two channels of \(\mathcal{N}\) and \(\mathcal{P}\) are flattened using \(\operatorname{Flatten}(\cdot)\). Subsequently, the features are reconstructed into the original time series \(\mathcal{X}\) through a MLP. This process can be represented as:
\begin{equation}
\begin{aligned}
\operatorname{Rec_{1}}(\mathcal{N})&=\operatorname{MLP}(\operatorname{Flatten}(\mathcal{N})),\\
    \operatorname{Rec_{2}}(\mathcal{P})&=\operatorname{MLP}(\operatorname{Flatten}(\mathcal{P})).
\end{aligned}
\end{equation}

\subsection{Dual Project Constraint}
To prevent the model from converging to trivial solutions, each layer's output requires to be constrained by \textit{Dual Project Head}. Prior research suggests that employing a similar structure effectively prevents models from converging to trivial solutions. For instance, SimCLR \cite{v1_simclr} uses a projection head that maps augmented views of an image into a latent space and promotes smoother convergence during training. The projection head comprises two layers of MLP connections, excluding non-linear and normalization layers. Hence, for both the inter-patch and intra-patch encoders, we can derive the projected representations $\mathcal{N}^{\prime} \in \mathbb{R}^{C \times N \times D}$ and $\mathcal{P}^{\prime} \in \mathbb{R}^{C \times P \times D}$. This computation can be represented as:
\begin{equation}
        \mathcal{N}^{\prime} = \operatorname{FC}(\operatorname{FC(\mathcal{N})}),\ 
        \mathcal{P}^{\prime} = \operatorname{FC}(\operatorname{FC(\mathcal{P})}).
\end{equation}
To amalgamate outputs from different layers and prevent model degradation, we employ a simple ReWeight module to assign distinct weights to various layers. Considering representations from $L$ layers, $\{\mathcal{N}_1,\mathcal{N}_2,\cdots,\mathcal{N}_L\}$ and $\{\mathcal{P}_1,\mathcal{P}_2,\cdots,\mathcal{P}_L\}$, we can derive the final result:
\begin{equation}
\begin{aligned}
    \mathcal{N}_l &= \alpha_l \cdot \mathcal{N}_l,\ \alpha_l = \operatorname{Softmax}(\alpha), \\
    \mathcal{P}_l &= \beta_l \cdot \mathcal{P}_l,\ \beta_l = \operatorname{Softmax}(\beta). \\
\end{aligned}
\end{equation}
Therefore, the outputs of the encoder can be represented as:
\begin{equation}
\begin{aligned}
\mathcal{N}=\alpha_1\cdot\mathcal{N}_1+\cdots+\alpha_L\cdot \mathcal{N}_L,\\
\mathcal{P}=\alpha_1\cdot\mathcal{P}_1+\cdots+\alpha_L\cdot \mathcal{P}_L.\\
\end{aligned}
\end{equation}
The introduction of ReWeight allows for a better balance between outputs from different layers, resulting in improved performance.
\subsection{Objective Function}
The inter-patch encoder and intra-patch encoder can represent two distinct views, denoted as $\mathcal{N}$ and $\mathcal{P}$. To quantify the dissimilarity between these representations, we utilize a comparative loss function based on Kullback–Leibler divergence (KL divergence), termed as Inter-Intra Discrepancy. Given the scarcity of anomalies in the data and the abundance of normal samples sharing hidden patterns, similar inputs should yield similar representations for both.
In Sec. \ref{sec:ablation}, the reasons for choosing the KL divergence as the basic loss are analyzed.
The loss function for $\mathcal{N}$ and $\mathcal{P}$ can be defined as follows:
\begin{equation}
\small
    \begin{aligned}
        \mathcal{L}_{\mathcal{N}}\{\mathcal{P},\mathcal{N}\}=\sum \operatorname{KL}(\mathcal{N},\operatorname{StopGrad}(\mathcal{P}))+\operatorname{KL}(\operatorname{StopGrad}(\mathcal{P}),\mathcal{N}),\\
        \mathcal{L}_{\mathcal{P}}\{\mathcal{P}, \mathcal{N}\}=\sum \operatorname{KL}(\mathcal{P}, \operatorname{StopGrad}(\mathcal{N}))+\operatorname{KL}(\operatorname{StopGrad}(\mathcal{N}), \mathcal{P}),
\label{eq:inter_intra_NP_loss}
\end{aligned}
\end{equation}
where $\operatorname{KL}(\cdot||\cdot)$ represents the KL divergence distance, and $\operatorname{StopGrad}$ denotes gradient stoppage, employed for asynchronously optimizing the two branches. The combined loss is defined as:
\begin{equation}
    \mathcal{L}_{cont}=\frac{\mathcal{L}_\mathcal{N}-\mathcal{L}_ \mathcal{P}}{len( \mathcal{N})}.
\end{equation}
Due to the mismatch in dimensions between $\mathcal{N}$ and $\mathcal{P}$, an initial upsampling of both is required for comparability. For the inter-patch view, where only differences between patches exist, we replicate the final $\mathcal{N}$ within patches. Conversely, for the intra-patch view, exhibiting anomalies between points within patches, we replicate it multiple times to obtain the final $\mathcal{P}$. This also addresses the necessity for introducing multi-scale, as it compensates for the information loss during upsampling.

To prevent model degradation, we augment the original loss function with an additional constraint from the projection head, defined as:
\begin{equation}
\mathcal{L}_{proj}=\frac{\mathcal{L}_{\mathcal{N}^\prime}-\mathcal{L}_ \mathcal{P}}{len( \mathcal{N})} + \frac{\mathcal{L}_\mathcal{N}-\mathcal{L}_ {\mathcal{P}^\prime}}{len( \mathcal{N})}.
\end{equation}
The final PatchAD loss comprises the two aforementioned components and the reconstruction term (\(\mathcal{L}_{rec} = \operatorname{MSE}(\hat{\mathcal{X}},\mathcal{X})\)), which is accompanied by a constraint coefficient to regulate the strength of the constraint.
\begin{equation}\label{eq:loss_fin}
    \mathcal{L} = (1-c) \cdot \mathcal{L}_{cont} + c \cdot \mathcal{L}_{proj} + \mathcal{L}_{rec}.
\end{equation}
\subsection{Anomaly Score}
Based on the assumption that normal points across different views share hidden patterns, while anomalies exhibit larger discrepancies, the difference between the two views can be employed as an anomaly score. The final anomaly score can be expressed as:
\begin{equation}
    \operatorname{AnomalyScore}(\mathcal{X})=\sum \operatorname{KL}(\mathcal{N},\mathcal{P})+\operatorname{KL}(\mathcal{P},\mathcal{N}).
\end{equation}
Based on the aforementioned anomaly score, we set a hyperparameter $\sigma$ to determine whether a point is considered an anomaly. If it surpasses the threshold, it is identified as an outlier.

\subsection{Multi-Scale of PatchAD}\label{sec:multi_scale}
For single-scale PatchAD, it features a comprehensive Patching, Patch Mixer Encoder, and Dual Project. This means that, assuming the patch size for single-scale PatchAD is \(P_i\), we obtain the final output of the Patch Mixer Encoder, \(\mathcal{\hat{X}}_i\), as well as the final outputs of the Dual Project, \(\mathcal{N}_i\) and \(\mathcal{P}_i\). These outputs are used to update the loss function \(\mathcal{L}_i\) (as shown in Eq. \eqref{eq:loss_fin}).

Here, \(i\) denotes the \(i\)-th single scale. For a multi-scale PatchAD, there are typically \(M\) branches. We optimize the loss functions \(\{\mathcal{L}_1, \mathcal{L}_2, \ldots, \mathcal{L}_M\}\) in an integrated manner, allowing different branches to learn features at various scales. During the testing phase, each single-scale branch produces its own anomaly score \(AS_i\), and the final anomaly score is the average of the scores from all branches, given by \(AS = \frac{1}{M} \sum_{i=1}^M AS_i\).


\section{Experiments}
\subsection{Setups and Evaluations}
\subsubsection{Datasets}
We evaluated different algorithms using a total of 8 datasets, with details referred to in the appendix.

\subsubsection{Metrics}

Considering comprehensively, we adopt two groups of metrics to conduct extensive experiments.
Detailed explanations for why these metrics are used can be found in Appendix \ref{app:metrics}.
\begin{enumerate}
    \item Metrics using point adjustment (PA): We prioritize these metrics because they are now the most commonly utilized group. To compare with advanced algorithms (including but not limited to \cite{dc_detector,anomaly_trans,OmniAnomaly,InterFusion,v1_USAD,timesnet}) that used point-adjusted metrics for evaluation, we employ Acc, Pre, Rec, and PA-F1 as the main metrics.
    \item Metrics without point adjustment: Meanwhile, as many current studies \cite{patch_bls,TSB-AD} are gradually using classic metrics (AUC, F1-cls) and interval evaluation metrics (Aff-F1, VUS), in order to compare with these algorithms (including but not limited to \cite{v1_couta,d3r,gpt2,NPSR,deep_if,trans_ad,dc_detector}), we organize additional experiments for further analysis in Sec. \ref{sec:comp_wo_pa}.
\end{enumerate}


\subsubsection{Baselines}\label{sec:baseline}
We compared our model against 30+ benchmark models, including:
(1) Reconstruction-based;
(2) Autoregression-based;
(3) Density-based;
(4) Clustering-based;
(5) The classic methods; 
(6) Change point detection and time series segmentation methods;
(7) Contrastive learning;
(8) Large Language Model;
(9) Diffusion model.

The details regarding the selection of the algorithms are further discussed in Appendix \ref{app:baseline}.

\subsection{Comparison Results}

\subsubsection{Comparisons via Metrics with PA}\label{sec:comp_w_pa}
From Table \ref{tab:comparison}, PatchAD achieves the best average-level PA-F1 performance. 
PatchAD attains superior results across multiple datasets and multiple metrics. Despite PatchAD's relatively simple network structure, its performance remains competitive. 
In the evaluation of the PA-F1 metric, DCdetector and AnomTrans emerge as the state-of-the-art(SOTA) models, with their robust capability to capture local anomalies granting them a competitive edge over other models in PA-F1. In contrast, PatchAD enhances its applicability to local anomalies through adaptive learning of local feature representations, achieving SOTA performance. 
Compared with AnomTrans's limited temporal granularity, PatchAD's multi-scale approach can better capture anomalies of different lengths, thus achieving superior PA-F1 performance.

Simultaneously, we observe that the additional models do not perform exceptionally well on the PA-F1 metric, primarily due to two reasons: firstly, they are mainly designed for the detection of interval anomalies, a fact that becomes evident in the subsequent analysis. Secondly, the PA-F1 metric employs point adjustment techniques, which can result in some undetected anomalies being classified as true positives, thereby introducing a potential inflation effect in the metric \cite{tkde_Flawed}. While PA-F1 is suitable for comparing the performance of different models in research contexts, it is less appropriate for assessing model performance in real-world scenarios. Consequently, this also accounts for the challenge faced by PatchAD in distinguishing itself from previous models. Therefore, in the following section, a new set of metrics will be employed to analyze the models from different perspectives.



\begin{table*}[!tbp]
  \centering  
  \caption{Comprehensive comparative results on four datasets. The P, R and PA-F1 are the precision, recall and F1 with PA. All results are in \%, the best in \textbf{Bold}, and the second in \uline{underlined}.}
  \resizebox{0.9\linewidth}{!}{
\begin{tabular}{c|ccc|ccc|ccc|ccc|c}
    \toprule
    \multicolumn{1}{c|}{\textbf{Dataset}} & \multicolumn{3}{c|}{\textbf{MSL}} & \multicolumn{3}{c|}{\textbf{SMAP}} & \multicolumn{3}{c|}{\textbf{PSM}} & \multicolumn{3}{c|}{\textbf{SWAT}} & \multicolumn{1}{c}{\multirow{2}[3]{*}{\textbf{\makecell*[c]{Average\\ PA-F1}}}} \\
\cmidrule{1-13}    \multicolumn{1}{c|}{\textbf{Metric}} & \textbf{P} & \textbf{R} & \textbf{PA-F1} & \textbf{P} & \textbf{R} & \textbf{PA-F1} & \textbf{P} & \textbf{R} & \textbf{PA-F1} & \textbf{P} & \textbf{R} & \textbf{PA-F1} &  \\
\midrule
LOF  & 47.72  & 85.25  & 61.18  & 58.93  & 56.33  & 57.60  & 57.89  & 90.49  & 70.61  & 72.15  & 65.43  & 68.62  & 64.50  \\
OCSVM  & 59.78  & 86.87  & 70.82  & 53.85  & 59.07  & 56.34  & 62.75  & 80.89  & 70.67  & 45.39  & 49.22  & 47.23  & 61.27  \\
U-Time  & 57.20  & 71.66  & 63.62  & 49.71  & 56.18  & 52.75  & 82.85  & 79.34  & 81.06  & 46.20  & 87.94  & 60.58  & 64.50  \\
IForest  & 53.94  & 86.54  & 66.45  & 52.39  & 59.07  & 55.53  & 76.09  & 92.45  & 83.48  & 49.29  & 44.95  & 47.02  & 63.12  \\
DAGMM  & 89.60  & 63.93  & 74.62  & 86.45  & 56.73  & 68.51  & 93.49  & 70.03  & 80.08  & 89.92  & 57.84  & 70.40  & 73.40  \\
ITAD  & 69.44  & 84.09  & 76.07  & 82.42  & 66.89  & 73.85  & 72.80  & 64.02  & 68.13  & 63.13  & 52.08  & 57.08  & 68.78  \\
VAR  & 74.68  & 81.42  & 77.90  & 81.38  & 53.88  & 64.83  & 90.71  & 83.82  & 87.13  & 81.59  & 60.29  & 69.34  & 74.80  \\
MMPCACD  & 81.42  & 61.31  & 69.95  & 88.61  & 75.84  & 81.73  & 76.26  & 78.35  & 77.29  & 82.52  & 68.29  & 74.73  & 75.93  \\
CL-MPPCA  & 73.71  & 88.54  & 80.44  & 86.13  & 63.16  & 72.88  & 56.02  & 99.93  & 71.80  & 76.78  & 81.50  & 79.07  & 76.05  \\
TS-CP2  & 86.45  & 68.48  & 76.42  & 87.65  & 83.18  & 85.36  & 82.67  & 78.16  & 80.35  & 81.23  & 74.10  & 77.50  & 79.91  \\
Deep-SVDD  & 91.92  & 76.63  & 83.58  & 89.93  & 56.02  & 69.04  & 95.41  & 86.49  & 90.73  & 80.42  & 84.45  & 82.39  & 81.44  \\
BOCPD  & 80.32  & 87.20  & 83.62  & 84.65  & 85.85  & 85.24  & 80.22  & 75.33  & 77.70  & 89.46  & 70.75  & 79.01  & 81.39  \\
LSTM-VAE  & 85.49  & 79.94  & 82.62  & 92.20  & 67.75  & 78.10  & 73.62  & 89.92  & 80.96  & 76.00  & 89.50  & 82.20  & 80.97  \\
BeatGAN  & 89.75  & 85.42  & 87.53  & 92.38  & 55.85  & 69.61  & 90.30  & 93.84  & 92.04  & 64.01  & 87.46  & 73.92  & 80.78  \\
LSTM  & 85.45  & 82.50  & 83.95  & 89.41  & 78.13  & 83.39  & 76.93  & 89.64  & 82.80  & 86.15  & 83.27  & 84.69  & 83.71  \\
OmniAnomaly  & 89.02  & 86.37  & 87.67  & 92.49  & 81.99  & 86.92  & 88.39  & 74.46  & 80.83  & 81.42  & 84.30  & 82.83  & 84.56  \\
InterFusion  & 81.28  & 92.70  & 86.62  & 89.77  & 88.52  & 89.14  & 83.61  & 83.45  & 83.52  & 80.59  & 85.58  & 83.01  & 85.57  \\
THOC  & 88.45  & 90.97  & 89.69  & 92.06  & 89.34  & 90.68  & 88.14  & 90.99  & 89.54  & 83.94  & 86.36  & 85.13  & 88.76  \\
AnomalyTrans  & 91.92  & 96.03  & 93.93  & 93.59  & \uline{99.41 } & \uline{96.41 } & 96.94  & 97.81  & 97.37  & 89.79  & \textbf{100.0 } & 94.62  & 95.58  \\
DCdetetor  & \textbf{92.22 } & 97.48  & \uline{94.77 } & \uline{94.43 } & 98.41  & 96.38  & 97.19  & 98.08  & \uline{97.63 } & \uline{93.25 } & \textbf{100.0 } & \uline{96.51 } & \uline{96.32 } \\
TimesNet   & 89.54  & 75.36  & 81.84  & 90.14  & 56.40  & 69.39  & 98.51  & 96.20  & 97.34  & 90.75  & 95.40  & 93.02  & 85.40  \\
PatchTST  & 88.34  & 70.96  & 78.70  & 90.64  & 55.46  & 68.82  & \textbf{98.84 } & 93.47  & 96.08  & 91.10  & 80.94  & 85.72  & 82.33  \\
GPT2-Adapter   & 82.00  & 82.91  & 82.45  & 90.60  & 60.95  & 72.88  & \uline{98.62 } & 95.68  & 97.13  & 92.20  & 96.34  & 94.23  & 86.67  \\
NPSR-pt \cite{NPSR} & 10.60  & \textbf{100.0 } & 19.16  & 24.99  & 96.37  & 39.68  & 27.74  & \textbf{100.0 } & 43.43  & 12.29  & \textbf{100.0 } & 21.89  & 31.04  \\
NPSR-seq \cite{NPSR} & 17.66  & \textbf{100.0 } & 30.02  & 12.79  & \textbf{100.0 } & 22.68  & 37.87  & \uline{99.95 } & 54.81  & 86.76  & 88.58  & 87.66  & 48.80  \\
M2N2 \cite{M2N2}  & 89.38  & 92.48  & 90.90  & 87.37  & 72.15  & 79.04  & 70.51  & 89.99  & 79.07  & 99.42  & 70.96  & 82.82  & 82.96 \\
AdaMemBLS \cite{adamembls} & 88.01  & 93.20  & 90.53  & 88.01  & 93.20  & 90.53  & 89.59  & 95.45  & 92.43  & 93.19  & 82.78  & 87.68  & 90.29 \\
\midrule
PatchAD & \uline{92.05 } & \uline{98.20 } & \textbf{95.02 } & \textbf{94.49 } & 99.13  & \textbf{96.75 } & 97.72  & 98.52  & \textbf{98.11 } & \textbf{93.28 } & \textbf{100.0 } & \textbf{96.52 } & \textbf{96.60 } \\
\bottomrule
\end{tabular}

}
  \label{tab:comparison}%
\end{table*}%

\begin{figure*}[!t]
  \centering
  \resizebox{0.9\linewidth}{!}{
  \begin{subfigure}[b]{0.2\textwidth}
    \includegraphics[width=\linewidth]{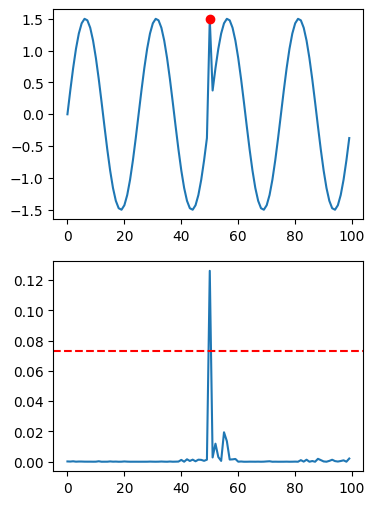}
    \caption{Global point}
  \end{subfigure}%
  \begin{subfigure}[b]{0.2\textwidth}
    \includegraphics[width=\linewidth]{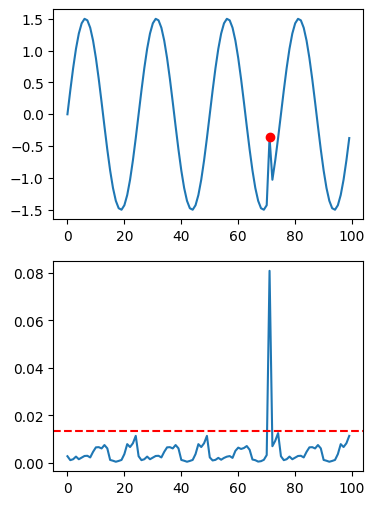}
    \caption{Contextual point}
  \end{subfigure}%
  \begin{subfigure}[b]{0.2\textwidth}
    \includegraphics[width=\linewidth]{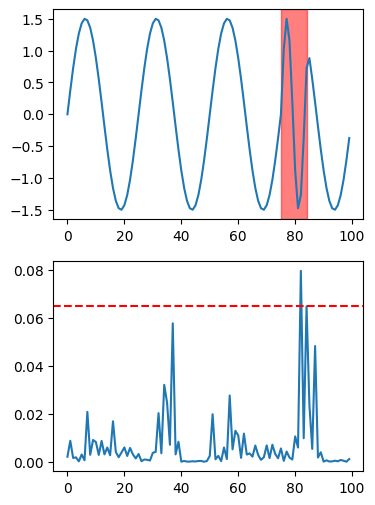}
    \caption{Seasonal}
  \end{subfigure}%
  \begin{subfigure}[b]{0.2\textwidth}
    \includegraphics[width=\linewidth]{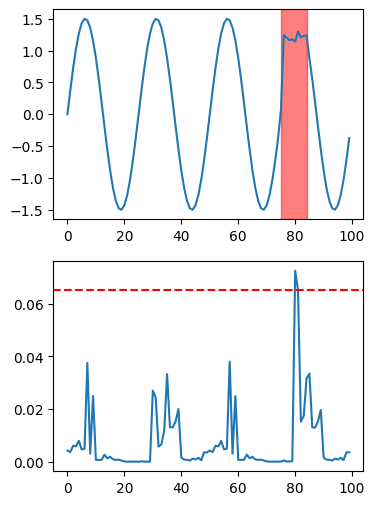}
    \caption{Group point}
  \end{subfigure}%
  \begin{subfigure}[b]{0.2\textwidth}
    \includegraphics[width=\linewidth]{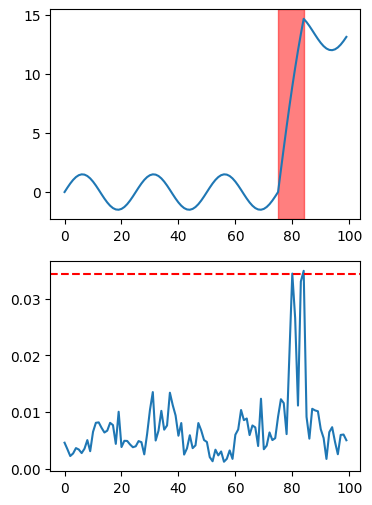}
    \caption{Trend}
  \end{subfigure}%
  }
  \caption{Visualization of generated synthetic data, PatchAD anomaly scores and ground truth.}
  \label{fig:anom_types}
    \vspace{-0.2cm}
\end{figure*}


\subsubsection{Comparisons via Metrics without PA}\label{sec:comp_wo_pa}

Figure \ref{sfig:overall_metric} depicts the average performance of PatchAD compared to various baselines across multiple datasets (MSL, SMAP, SWAT, WADI, PSM, and SWAN). 
Here, Aff-F1 is the affiliation metric \cite{AFFI}. V-ROC and V-RR are volumes under the surfaces created based on ROC curve and PR curve \cite{VUS}, respectively. 
In terms of the F1-cls, Aff-F1, and V-ROC metrics, PatchAD demonstrates relative performance improvements of 6.84\%, 4.27\%, and 2.49\% compared to the second-place model, respectively. This indicates a significant enhancement in PatchAD's capacity for detecting interval anomalies compared to previous models. Notably, we observe that AnomTrans and DCdetector, which perform well on the PA-F1 metric, exhibit poor performance on interval anomaly metrics. This is primarily because these models are more focused on local or point anomalies, resulting in better performance on the PA-F1 metric. PatchAD, on the other hand, benefits from additional learning of generalized global features, enhancing its performance on interval anomalies, thus improving both PA-F1 and Aff-F1 metrics.

Furthermore, we note that although TimesNet was proposed earlier, it currently exhibits relatively strong performance when considering the two sets of metrics. This can be attributed to the diversity of contemporary anomalies, which vary in type, periodicity, and length. TimesNet's capability to learn temporal information across different frequencies enables it to handle various anomalies more effectively. PatchAD similarly adopts this approach, with multi-scale temporal feature extraction ensuring the learning of information across different temporal frequencies. 

NPSR also takes into account the utilization of two distinct feature types in its design and uniquely processes anomaly scores, thereby ensuring its excellent theoretical lower bound, which guarantees good performance in the face of real interval anomalies. Likewise, PatchAD benefits from the assurance provided by the lower bound of information entropy and the diverse features extracted through different Mixers, resulting in similarly outstanding performance. In contrast, COUTA's training relies on anomaly injection, leading to varying adaptability across different real datasets; for instance, it performs well on the PSM and SWAN datasets but shows average performance on the WADI and SWAT datasets.

\begin{table*}[htbp]
  \centering
  \caption{Comprehensive comparative results on six datasets. All results are in \%, the best in \textbf{Bold}, and the second in \uline{underlined}.}
  \resizebox{0.9\linewidth}{!}{
\begin{tabular}{l|ccccc|ccccc|ccccc}
\toprule
\textbf{Datasets} & \multicolumn{5}{c|}{\textbf{MSL}}     & \multicolumn{5}{c|}{\textbf{SMAP}}    & \multicolumn{5}{c}{\textbf{SWAT}} \\
\midrule
\textbf{Methods} & \textbf{AUC} & \textbf{F1-cls} & \textbf{Aff-F1} & \textbf{V-ROC} & \textbf{V-PR} & \textbf{AUC} & \textbf{F1-cls} & \textbf{Aff-F1} & \textbf{V-ROC} & \textbf{V-PR} & \textbf{AUC} & \textbf{F1-cls} & \textbf{Aff-F1} & \textbf{V-ROC} & \textbf{V-PR} \\
\midrule
LOF [2000] & 55.75  & 19.36  & 66.86  & 60.85  & 17.19  & \uline{62.39 } & 23.64  & 58.74  & 57.56  & \textbf{18.59 } & 84.59  & 58.22  & 5.25  & 66.67  & 41.88  \\
IForest [2008] & 59.09  & 10.90  & {70.44 } & 64.74  & 18.29  & 61.15  & 23.78  & 50.52  & \uline{59.31 } & 16.45  & 83.80  & 38.52  & 5.27  & 72.29  & 52.07  \\
PCA [2003] & 53.25  & 10.33  & \uline{70.72 } & 59.42  & 18.08  & 58.70  & 22.96  & 50.72  & 43.90  & 11.69  & 81.85  & 26.05  & 5.25  & 61.87  & 44.75  \\
Deep SVDD [2018] & \uline{61.80 } & \uline{24.30 } & 68.98  & 59.54  & 19.36  & 61.14  & 21.28  & 59.64  & 41.95  & 11.56  & 86.81  & 22.57  & 69.77  & 50.96  & 56.48  \\
USAD [2020] & 57.12  & 22.50  & 67.69  & 58.86  & 19.77  & \textbf{62.79 } & 16.57  & 50.41  & 33.78  & 11.28  & 88.66  & 21.70  & 69.28  & 84.38  & 62.74  \\
TCN-ED [2021] & 51.14  & 19.66  & 67.93  & 58.62  & 18.13  & 58.91  & 22.90  & 67.89  & 43.79  & 11.73  & 89.23  & 21.65  & 69.26  & 61.53  & 57.94  \\
TranAD [2022] & 50.03  & 22.81  & 68.97  & 57.25  & 17.55  & 59.74  & 22.74  & 50.51  & 45.15  & 11.98  & 88.90  & 25.50  & 71.42  & \uline{86.25 } & 64.19  \\
NCAD [2022] & 60.20  & 22.05  & 67.25  & \uline{67.29 } & \uline{21.32 } & 53.45  & 23.09  & 68.32  & 48.37  & 13.22  & 82.92  & 68.72  & 73.89  & 76.76  & 44.91  \\
Deep IF [2023] & 55.94  & 19.11  & 67.94  & 52.33  & 17.04  & 60.09  & \textbf{29.14 } & 69.59  & 56.89  & 14.77  & 89.52  & 21.65  & 69.26  & 50.19  & 56.39  \\
AnomTrans [2021] & 52.61  & 18.39  & 66.81  & 53.03  & 13.21  & 52.18  & 16.06  & \textbf{71.44 } & 52.61  & 16.55  & 80.80  & 23.44  & 66.77  & 21.49  & 8.29  \\
TimesNet [2022] & 57.18  & 21.24  & 67.64  & 64.63  & 20.67  & 53.73  & 24.37  & 66.19  & 48.83  & 12.91  & 73.24  & 21.66  & 71.55  & 31.10  & 10.29  \\
GPT2-Ada. [2023] & 52.03  & 13.72  & 68.81  & 53.57  & 17.38  & 55.48  & 24.12  & 69.48  & 57.31  & 15.78  & 52.30  & 22.30  & 68.41  & 50.01  & 10.29  \\
DCdetector [2023] & 50.31  & 11.62  & 67.85  & 52.00  & 14.68  & 58.50  & {26.56 } & \uline{71.37 } & \textbf{60.38 } & \uline{16.87 } & 52.78  & 23.24  & 68.56  & 50.84  & 14.68  \\
D3R [2024] & \textbf{63.00 } & 23.98  & 69.79  & \textbf{69.02 } & {20.93 } & 54.56  & 22.71  & 67.92  & 48.67  & 13.23  & 79.95  & 45.89  & 68.96  & 84.38  & 57.27  \\
COUTA [2024] & 55.59  & 20.88  & 67.68  & 61.49  & 18.68  & 58.74  & 22.69  & 67.97  & 44.72  & 11.91  & 75.54  & 47.65  & 46.92  & 72.57  & 25.06  \\
NPSR [2023] & 61.16  & 23.72  & 68.42  & 66.44  & 19.72  & 38.74  & 22.68  & 67.95  & 41.91  & 11.42  & \uline{90.18 } & \uline{76.88 } & \uline{75.86 } & 84.93  & {65.42 } \\
M2N2 [2024]  & 59.15  & 21.76  & 67.88  & 65.40  & \textbf{21.55 } & 54.05  & 22.92  & 67.95  & 48.74  & 14.36  & 67.79  & 38.50  & 73.44  & 61.14  & 26.16  \\
AdaMemBLS [2025] & 51.78  & 19.11  & \textbf{\uline{71.41 }} & 57.46  & 16.06  & 53.66  & \uline{26.93 } & 69.49  & 55.57  & 14.26  & 81.61  & 74.10  & 69.76  & 82.00  & \textbf{66.12 } \\
\midrule
PatchAD & 60.94  & \textbf{24.92 } & 70.10  & 66.43  & 19.65  & 60.27  & 24.74  & 70.95  & 48.30  & 13.71  & \textbf{91.19 } & \textbf{77.69 } & \textbf{78.76 } & \textbf{89.91 } & \uline{65.74 } \\
\midrule
\textbf{Datasets} & \multicolumn{5}{c|}{\textbf{WADI}}    & \multicolumn{5}{c|}{\textbf{PSM}}     & \multicolumn{5}{c}{\textbf{NIPS-TS-Swan}} \\
\midrule
\textbf{Methods} & \textbf{AUC} & \textbf{F1-cls} & \textbf{Aff-F1} & \textbf{V-ROC} & \textbf{V-PR} & \textbf{AUC} & \textbf{F1-cls} & \textbf{Aff-F1} & \textbf{V-ROC} & \textbf{V-PR} & \textbf{AUC} & \textbf{F1-cls} & \textbf{Aff-F1} & \textbf{V-ROC} & \textbf{V-PR} \\
\midrule
LOF [2000] & 65.44  & 10.58  & 34.36  & 66.67  & 41.88  & \textbf{81.28 } & 23.91  & 62.32  & \textbf{80.42 } & 64.55  & 77.89  & 21.37  & 19.62  & 89.42  & 88.56  \\
IForest [2008] & 74.84  & 30.88  & 67.90  & 72.29  & 52.07  & 75.12  & 38.52  & 56.77  & \uline{75.19 } & 55.50  & \textbf{85.12 } & 46.18  & 27.81  & 91.27  & 90.98  \\
PCA [2003] & 49.47  & 34.30  & 50.01  & 61.87  & 44.75  & 72.06  & 43.20  & \textbf{78.74 } & 73.97  & 54.55  & 72.88  & 49.18  & 27.35  & 90.42  & 90.23  \\
Deep SVDD [2018] & 59.47  & 7.34  & 52.77  & 50.96  & 56.48  & \uline{78.09 } & 44.46  & 70.06  & 53.97  & \uline{66.92 } & 75.32  & \uline{55.99 } & 69.66  & 80.89  & 91.74  \\
USAD [2020] & 53.61  & 10.86  & 70.90  & 84.38  & 62.74  & 64.53  & 47.90  & 70.88  & 64.02  & \textbf{69.18 } & 66.76  & 52.63  & 61.22  & 91.40  & \textbf{94.50 } \\
TCN-ED [2021] & 53.01  & 10.86  & 70.90  & 61.53  & 57.94  & 63.97  & 43.46  & 69.42  & 57.48  & 42.64  & 71.37  & 49.22  & 71.03  & 80.88  & 91.73  \\
TranAD [2022] & 52.60  & 11.78  & 69.54  & \uline{86.25 } & \uline{64.19 } & 65.22  & 43.20  & 73.30  & 72.42  & 56.02  & 70.00  & 51.93  & 70.49  & 88.01  & 92.59  \\
NCAD [2022] & 62.84  & 10.87  & 70.96  & 76.76  & 44.91  & 61.76  & 46.61  & 63.10  & 67.09  & 48.26  & 52.24  & 37.57  & 68.87  & 85.31  & 93.44  \\
Deep IF [2023] & 51.43  & 10.86  & 70.90  & 50.19  & 56.39  & 69.06  & 43.47  & 69.41  & 67.71  & 45.98  & 73.30  & 49.19  & 71.05  & 80.89  & 91.74  \\
AnomTrans [2021] & 52.98  & 11.24  & 66.77  & 21.49  & 8.29  & 52.18  & 39.81  & 68.13  & 49.12  & 31.56  & 54.62  & 44.86  & 69.11  & 83.56  & 81.64  \\
TimesNet [2022] & 65.08  & 17.65  & 49.05  & 31.10  & 10.29  & 58.74  & 43.60  & 72.15  & 66.37  & 48.05  & 53.62  & 43.16  & 69.53  & \uline{91.58 } & \uline{94.30 } \\
GPT2-Ada. [2023] & 51.21  & 10.68  & 71.04  & 50.01  & 10.29  & 51.23  & 35.24  & 69.17  & 52.45  & 32.79  & 58.77  & 45.23  & 65.09  & 84.31  & 84.23  \\
DCdetector [2023] & 50.12  & 11.33  & 73.23  & 50.84  & 14.68  & 50.38  & 22.72  & 68.08  & 51.73  & 32.44  & 50.35  & 48.83  & 70.99  & 85.84  & 83.11  \\
D3R [2024] & 51.39  & 12.86  & 72.17  & 84.38  & 57.27  & 60.64  & 44.24  & 69.99  & 66.92  & 49.38  & 62.54  & 49.45  & \uline{72.10 } & 89.27  & 88.29  \\
COUTA [2024] & 50.52  & 26.92  & 37.44  & 72.57  & 25.06  & 69.36  & 48.21  & 73.21  & 64.50  & 43.15  & 71.38  & 47.19  & 66.54  & 82.15  & 92.26  \\
NPSR [2023] & \uline{80.19 } & \uline{50.67 } & \uline{75.74 } & 84.93  & \textbf{65.42 } & 70.67  & \textbf{51.02 } & 70.56  & 70.55  & 52.12  & 62.61  & 49.30  & 71.07  & 90.90  & 90.35  \\
M2N2 [2024]  & 55.68  & 13.55  & 70.90  & 43.99  & 7.98  & 54.91  & 47.71  & 69.84  & 40.06  & 30.61  & 70.27  & \uline{57.47 } & 71.05  & 89.77  & 90.18  \\
AdaMemBLS [2025] & 53.85  & 29.09  & 70.93  & 63.47  & 21.40  & 63.26  & 46.69  & 71.51  & 65.07  & 49.24  & 51.27  & 49.35  & 71.09  & 87.12  & 85.92  \\
\midrule
PatchAD & \textbf{85.27 } & \textbf{52.82 } & \textbf{79.04 } & \textbf{88.29 } & 49.60  & 64.74  & \uline{49.33 } & \uline{76.73 } & 65.11  & 40.72  & \uline{79.53 } & \textbf{64.52 } & \textbf{72.36 } & \textbf{95.62 } & 91.52  \\
\bottomrule
\end{tabular}%
    }
  \label{tab:comparison_wo_pa}%
  
\end{table*}%

\subsection{Model Analysis}
\subsubsection{Ablation Studies}\label{sec:ablation}

\begin{figure}
    \centering
    \includegraphics[width=0.6\linewidth]{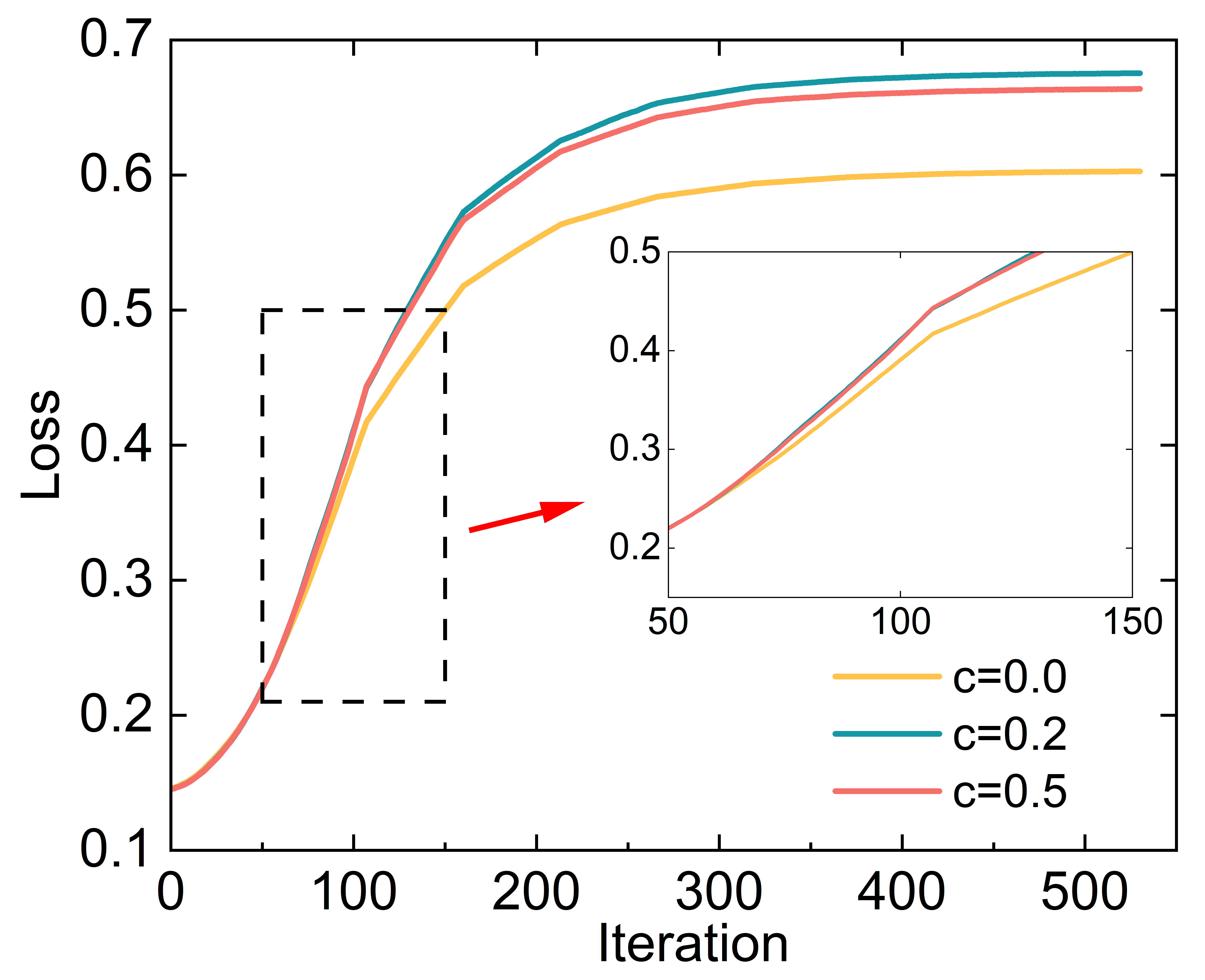}
    \caption{Loss analysis of PatchAD.}
    \vspace{-3mm}
    \label{fig:contraint_loss}
\end{figure}

\textbf{Ablation on Components}:
To study the impact of different components in our model, we conducted ablation experiments across four datasets, as presented in Table \ref{tab:ablation}.
In a manner analogous to the comparative experimental setup, we selected two sets of metrics to analyze the impact of components on the model from different perspectives. The first set comprises point-adjusted metrics, including Acc, Pre, Rec, and PA-F1. These metrics exhibit superior adaptability to noisy labels and demonstrate greater robustness in the presence of mislabeled data \cite{patch_bls,TSB-AD}. 
The second set consists of metrics that do not employ point adjustments, including AUC, F1-cls, V-ROC, and V-PR. While AUC and F1-cls exhibit poorer robustness to label noise and lower flexibility, their stricter evaluation criteria ensure a lower bound on the accuracy of the assessment \cite{TSB-AD}. Conversely, V-ROC and V-PR are two interval-based evaluation metrics that provide more accurate assessments for longer intervals of anomalies, thereby enhancing comprehensiveness.
The ``wo\_ch\_sharing'' method employs two independent Channel Mixer modules to model channel features, while ``wo\_ch\_mixer'' denotes removing the Channel Mixer module. These two variants exhibit significant effects, particularly on the MSL and SMAP datasets, indicating that modelling inter-channel information and sharing channel information benefit model learning. Besides, ``wo\_pos\_emb'' represents a variant without positional encoding, and removing positional encoding significantly affects PatchAD, highlighting the crucial role of positional encoding in its design. Furthermore, ``wo\_constraint'' showcases the impact of removing the constraint on the projection head. In conjunction with Figure \ref{fig:contraint_loss}, we observed that a certain level of constraint facilitates easier model optimization, leading to improved results. This constraint prevents the model from converging to trivial solutions. In subsequent experiments, the default setting for the projection head constraint is 0.2.

Considering the limitation of insufficient real labels for threshold selection in practical scenarios, we validated the practicality of PatchAD using the dynamic threshold algorithm based on Extreme Value Theory (SPOT) \cite{spot}, as shown in the second-to-last row of Table \ref{tab:ablation}. Results indicate that its performance gap with the optimal model is minimal, demonstrating that dynamic threshold settings can replace manual \(\sigma\) tuning to achieve favorable outcomes in real-world applications. It is important to note that AUC, V-ROC, and V-PR are threshold-agnostic metrics, so using SPOT does not affect these results.

\begin{table*}[htbp]
  \centering
  \caption{Ablation studies on MSL, SMAP, WADI, and PSM datasets.}
  \resizebox{0.9\linewidth}{!}{
\begin{tabular}{l|cccc|cccc|cccc|cccc}
\toprule
\textbf{Dataset} & \multicolumn{4}{c|}{\textbf{MSL}} & \multicolumn{4}{c|}{\textbf{SMAP}} & \multicolumn{4}{c|}{\textbf{WADI}} & \multicolumn{4}{c}{\textbf{PSM}} \\
\midrule
\textbf{With PA} & \textbf{Acc} & \textbf{Pre} & \textbf{Rec} & \textbf{PA-F1} & \textbf{Acc} & \textbf{Pre} & \textbf{Rec} & \textbf{PA-F1} & \textbf{Acc} & \textbf{Pre} & \textbf{Rec} & \textbf{PA-F1} & \textbf{Acc} & \textbf{Pre} & \textbf{Rec} & \textbf{PA-F1} \\
\midrule
wo\_ch\_sharing & 98.78 & 91.93 & 96.9  & 94.35 & 98.84 & 93.5  & 97.75 & 95.58 & 98.71 & 84.99 & 94.25 & 89.38 & 97.69  & 97.62  & 93.96  & 95.75  \\
wo\_ch\_mixer & 98.78 & 92.03 & 96.82 & 94.36 & 98.77 & 93.5  & 97.1  & 95.27 & 98.92 & 85.4  & 97.93 & 91.24 & 96.77  & 98.23  & 89.98  & 93.92  \\
wo\_mixrep\_mixer & 96.41 & 89.54 & 74.66 & 81.43 & 95.25 & 91.52 & 68.55 & 78.7  & 95.76 & 67.55 & 51.99 & 58.76 & 97.35  & 97.87  & 92.44  & 95.08  \\
wo\_pos\_emb & 97.85 & 91.51 & 87.73 & 89.58 & 96.47 & 92.18 & 79.13 & 85.16 & 98.23 & 83.96 & 85.53 & 84.74 & 96.89  & 97.84  & 90.77  & 94.17  \\
wo\_constraint & 98.51 & 91.67 & 94.49 & 93.06 & 99.01 & 93.55 & 99.05 & 96.22 & 98.73 & 85.26 & 94.25 & 89.53 & 93.95  & 97.22  & 80.45  & 88.04  \\
\midrule
PatchAD with SPOT   & 98.49 & 89.94 & 96.46 & 93.08 & 98.93 & 94.2  & 97.65 & 95.89 & 98.53 & 84.18 & 91.53 & 87.7  & 97.75 & 97.79 & 93.99 & 95.85 \\
PatchAD (Ours) & 98.92 & 92.05 & 98.2  & 95.02 & 99.1  & 94.49 & 99.13 & 96.75 & 99.06 & 85.92 & 100   & 92.43 & 98.89  & 97.72  & 98.52  & 98.11  \\
\midrule
\textbf{Dataset} & \multicolumn{4}{c|}{\textbf{MSL}} & \multicolumn{4}{c|}{\textbf{SMAP}} & \multicolumn{4}{c|}{\textbf{WADI}} & \multicolumn{4}{c}{\textbf{PSM}} \\
\midrule
\textbf{Without PA} & \textbf{AUC} & \textbf{F1} & \textbf{V-ROC} & \textbf{V-PR} & \textbf{AUC} & \textbf{F1} & \textbf{V-ROC} & \textbf{V-PR} & \textbf{AUC} & \textbf{F1} & \textbf{V-ROC} & \textbf{V-PR} & \textbf{AUC} & \textbf{F1} & \textbf{V-ROC} & \textbf{V-PR} \\
\midrule
wo\_ch\_sharing & 53.54  & 20.57  & 60.07  & 18.14  & 59.46  & 22.71  & 43.36  & 11.60  & 85.24  & 50.33  & 88.13  & 49.22  & 64.65  & 48.00  & 65.93  & 40.13  \\
wo\_ch\_mixer & 53.09  & 20.16  & 59.69  & 17.99  & 59.47  & 22.68  & 43.06  & 11.59  & 82.41  & 46.50  & 86.88  & 48.00  & 65.02  & 48.08  & 66.59  & 40.65  \\
wo\_mixrep\_mixer & 53.50  & 20.70  & 60.02  & 18.13  & 59.35  & 22.69  & 43.41  & 11.61  & 81.14  & 50.61  & 87.96  & 47.09  & 64.47  & 48.11  & 65.91  & 40.09  \\
wo\_pos\_emb & 52.91  & 19.55  & 59.68  & 17.95  & 55.95  & 22.68  & 46.87  & 12.15  & 84.63  & 51.02  & 87.36  & 49.96  & 64.76  & 48.04  & 66.17  & 40.25  \\
wo\_constraint & 53.53  & 20.79  & 59.97  & 18.11  & 59.93  & 22.71  & 42.85  & 11.52  & 83.56  & 51.29  & 86.78  & 46.09  & 64.93  & 47.95  & 66.54  & 40.53  \\
\midrule
PatchAD with SPOT   & 60.94  & 23.45 & 66.43 & 19.65 & 60.27  & 23.77 & 48.30  & 13.71  & 85.27  & 52.75 & 88.29  & 49.60  & 64.74  & 48.58  & 65.11  & 40.72  \\
PatchAD (Ours) & 60.94  & 23.92  & 66.43  & 19.65  & 60.27  & 24.74  & 48.30  & 13.71  & 85.27  & 52.82  & 88.29  & 49.60  & 64.74  & 49.33  & 65.11  & 40.72  \\
\bottomrule
\end{tabular}%
    }
  \label{tab:ablation}%
\end{table*}%



\textbf{Ablation on Loss Functions}:
To facilitate the spatial separation of the two distinct views, the design in Eq. (\ref{eq:inter_intra_NP_loss}) employs the Kullback-Leibler (\(\operatorname{KL}\)) divergence as the foundational loss. To further analyze the impact of various types of loss functions on the final model performance, we selected the following three common and widely utilized distance or distributional metric functions as alternatives, enabling an examination of their effects on the ultimate outcomes.
\begin{enumerate}
    \item L2 Distance (L2).
    \item Wasserstein Distance (Wasserstein).
    \item Jensen–Shannon Divergence (JSD).
\end{enumerate}
Table \ref{tab:abl_loss} illustrates the performance of the PatchAD model after substituting the three loss functions in comparison to the model utilizing KL divergence (Ours). 

In the comparison of the first set of metrics employing PA, the models based on L2 and JSD exhibit average performance that falls short of the final model; however, the overall difference from Ours is minimal. Conversely, the Wasserstein distance yields a PA-F1 score of only 83.93 on the WADI dataset, indicating a performance that is significantly inferior to that of the other distance functions. This may be attributed to the fact that the Wasserstein distance contributes to anomaly scores in a more smooth manner, which, in turn, results in a reduced recall rate for the model and consequently leads to suboptimal performance.

In the comparison of the second set of metrics, PatchAD consistently ranks as the best across nearly all datasets and metrics. Additionally, models trained with the other three foundational distances exhibit markedly lower performance in AUC, V-ROC, and V-PR compared to PatchAD. This suggests that these models do not possess sufficiently clear decision boundaries for anomalies, thereby complicating the model's ability to accurately differentiate between anomalies and normal instances.

Overall, we posit that JSD represents the best alternative to KL divergence, primarily because the construction of JSD is based on KL divergence, thereby retaining KL divergence's property of optimizing over probability distributions. However, JSD is a symmetric loss function, while KL divergence is asymmetric; this discrepancy limits the ability to achieve asynchronous optimization of the parameters in different modules (the Inter Mixer and Intra Mixer). Ultimately, this results in JSD performing poorly in enhancing the distinction between normal and abnormal instances, a phenomenon that aligns with previous work \cite{anomaly_trans}.

\begin{table*}[htbp]
  \centering
  \caption{The Influence of different loss functions as fundamental distances on PatchAD.} 
  \resizebox{0.9\linewidth}{!}{
\begin{tabular}{l|cccc|cccc|cccc|rrrr}
\toprule
\textbf{Dataset} & \multicolumn{4}{c|}{\textbf{MSL}} & \multicolumn{4}{c|}{\textbf{SMAP}} & \multicolumn{4}{c|}{\textbf{WADI}} & \multicolumn{4}{c}{\textbf{PSM}} \\
\midrule
\textbf{With PA} & \textbf{Acc} & \textbf{Pre} & \textbf{Rec} & \textbf{PA-F1} & \textbf{Acc} & \textbf{Pre} & \textbf{Rec} & \textbf{PA-F1} & \textbf{Acc} & \textbf{Pre} & \textbf{Rec} & \textbf{PA-F1} & \multicolumn{1}{c}{\textbf{Acc}} & \multicolumn{1}{c}{\textbf{Pre}} & \multicolumn{1}{c}{\textbf{Rec}} & \multicolumn{1}{c}{\textbf{PA-F1}} \\
\midrule
L2    & 98.12  & 88.92  & 93.82  & 91.30  & 98.90  & 93.75  & 97.94  & 95.80  & \uline{99.03 } & \uline{86.89 } & 97.93  & \uline{92.08 } & 97.12  & 97.65  & 91.84  & 94.65  \\
Wasserstein & 98.16  & 88.85  & \uline{94.41 } & 91.55  & \textbf{99.20 } & \uline{95.33 } & \uline{98.57 } & \textbf{96.92 } & 98.07  & 80.35  & 87.85  & 83.93  & 97.59  & 97.45  & 93.75  & 95.56  \\
JSD   & \uline{98.32 } & \uline{90.79 } & 93.51  & \uline{92.13 } & \uline{99.16 } & \textbf{97.81 } & 95.55  & 96.66  & 99.01  & 85.37  & \uline{98.57 } & 91.50  & \uline{98.41 } & \textbf{97.88 } & \uline{96.33 } & \uline{97.10 } \\
\midrule
PatchAD (Ours) & \textbf{98.92 } & \textbf{92.05 } & \textbf{98.20 } & \textbf{95.02 } & 99.10  & 94.49  & \textbf{99.13 } & \uline{96.75 } & \textbf{99.06 } & \textbf{85.92 } & \textbf{100.00 } & \textbf{92.43 } & \textbf{98.89 } & \uline{97.72 } & \textbf{98.52 } & \textbf{98.11 } \\
\midrule
\textbf{Dataset} & \multicolumn{4}{c|}{\textbf{MSL}} & \multicolumn{4}{c|}{\textbf{SMAP}} & \multicolumn{4}{c|}{\textbf{WADI}} & \multicolumn{4}{c}{\textbf{PSM}} \\
\midrule
\textbf{Without PA} & \textbf{AUC} & \textbf{F1-cls} & \textbf{V-ROC} & \textbf{V-PR} & \textbf{AUC} & \textbf{F1-cls} & \textbf{V-ROC} & \textbf{V-PR} & \textbf{AUC} & \textbf{F1-cls} & \textbf{V-ROC} & \textbf{V-PR} & \multicolumn{1}{c}{\textbf{AUC}} & \multicolumn{1}{c}{\textbf{F1}} & \multicolumn{1}{c}{\textbf{V-ROC}} & \multicolumn{1}{c}{\textbf{V-PR}} \\
\midrule
L2    & 53.52  & \uline{20.10 } & 60.24  & \uline{18.15 } & 58.69  & 22.68  & 44.01  & \uline{11.67 } & \textbf{85.54 } & 50.22  & 87.44  & 47.71  & 64.64  & 48.17  & 65.17  & 40.12  \\
Wasserstein & 53.39  & 19.62  & 60.13  & 18.08  & \uline{59.56 } & \uline{22.69 } & 43.24  & 11.57  & 84.15  & 51.61  & 86.58  & 46.13  & 64.66  & 48.20  & \textbf{65.71 } & \uline{40.56 } \\
JSD   & \uline{53.64 } & 19.57  & \uline{60.34 } & 18.13  & 58.78  & 22.68  & \uline{44.02 } & \uline{11.67 } & 85.08  & \uline{52.62 } & \uline{87.92 } & \uline{49.24 } & \textbf{65.66 } & \uline{49.11 } & \uline{65.69 } & 40.36  \\
\midrule
PatchAD (Ours) & \textbf{60.94 } & \textbf{23.92 } & \textbf{66.43 } & \textbf{19.65 } & \textbf{60.27 } & \textbf{24.74 } & \textbf{48.30 } & \textbf{13.71 } & \uline{85.27 } & \textbf{52.82 } & \textbf{88.29 } & \textbf{49.60 } & \uline{64.74 } & \textbf{49.33 } & 65.11  & \textbf{{40.72 }} \\
\bottomrule
\end{tabular}%
    }
  \label{tab:abl_loss}%
\end{table*}%

\subsubsection{Visual Analysis}
\begin{figure}[!t]
    \centering
    \subfloat[Case study on SWAT dataset]{\includegraphics[width=0.5\linewidth]{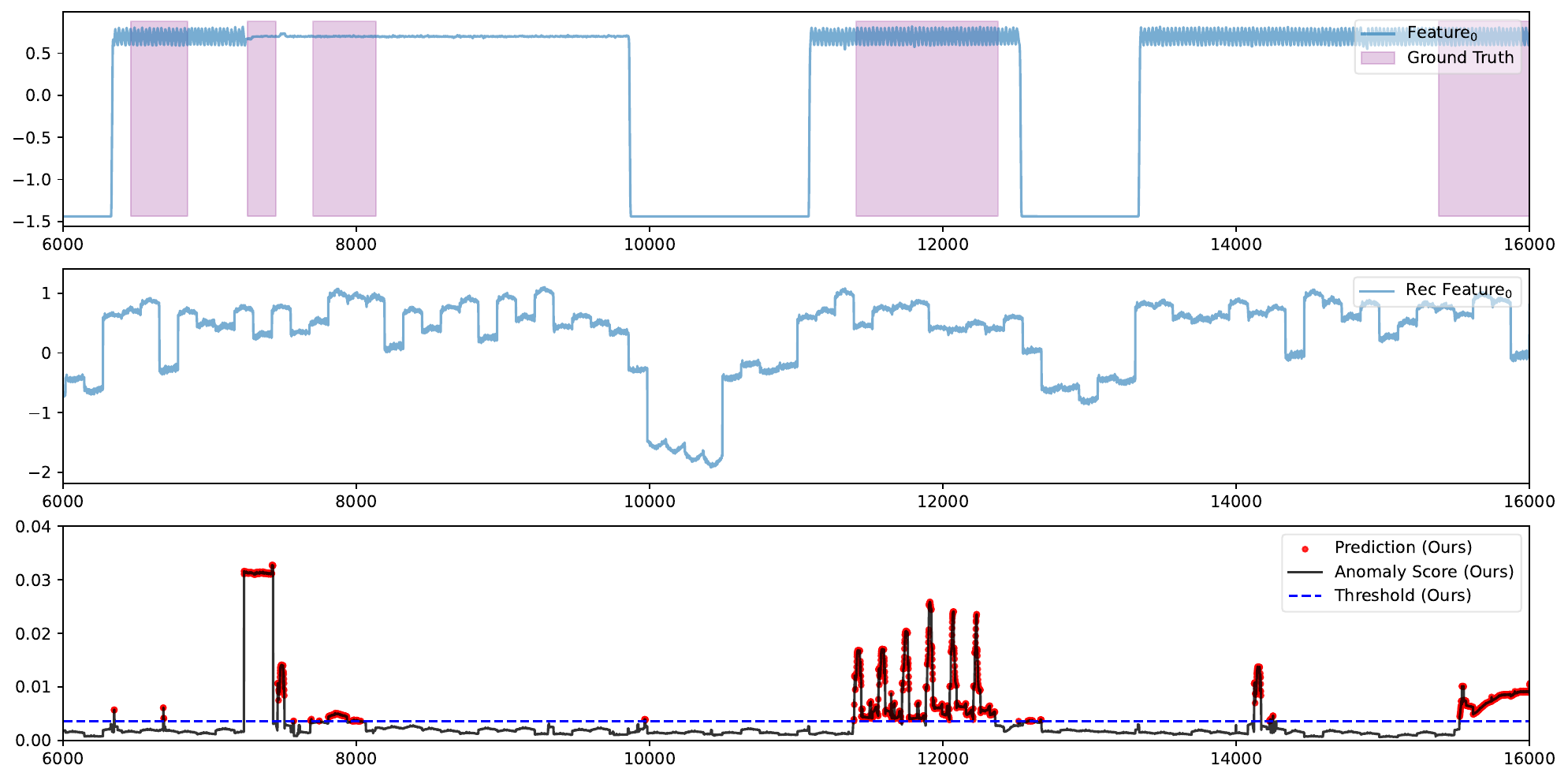}} 
    \subfloat[Case study on WADI dataset]{\includegraphics[width=0.5\linewidth]{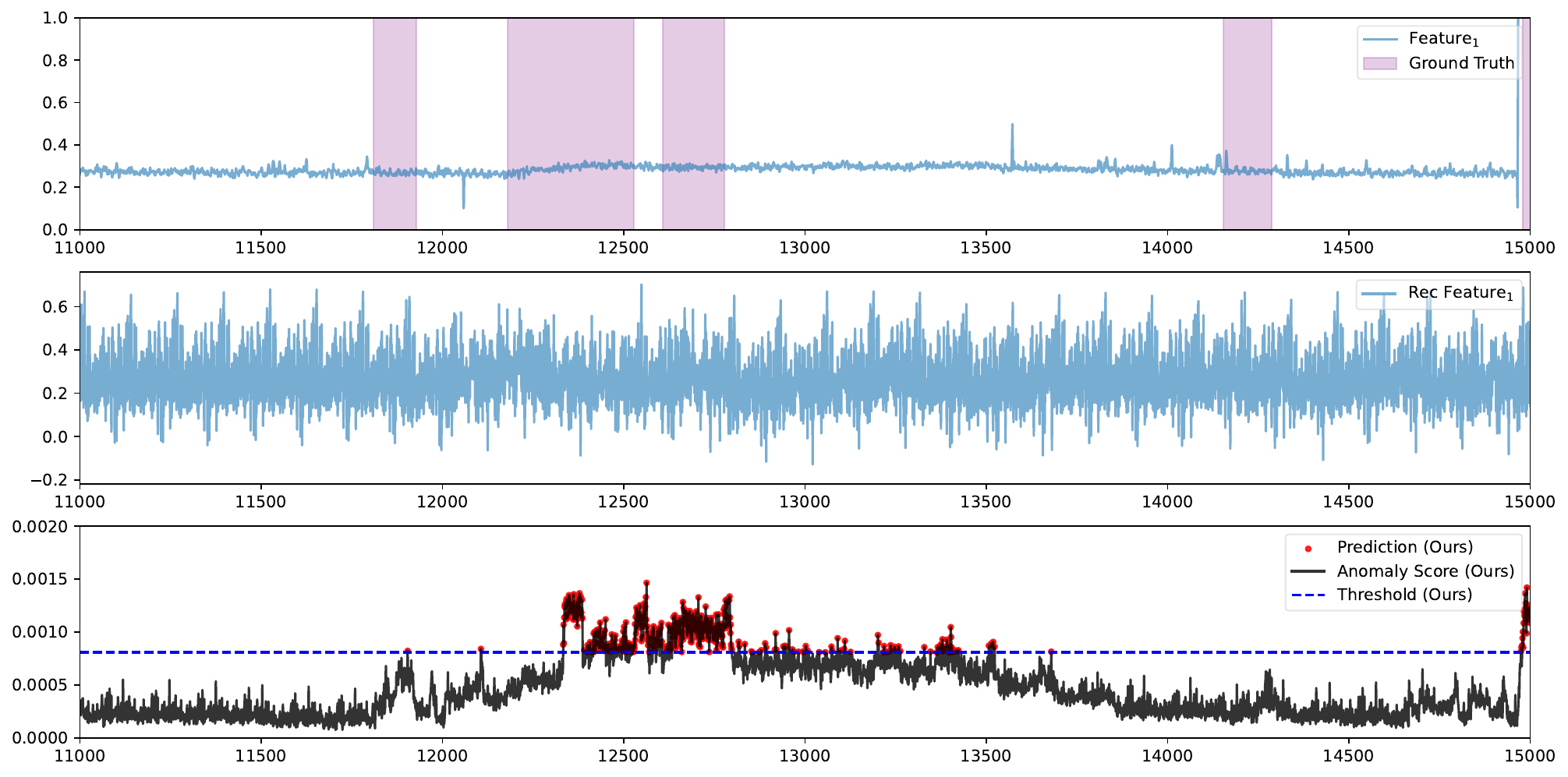}}
    \caption{The case studies on SWAT and WADI datasets.}
    \label{fig:case_study}
    \vspace{-0.2cm}
\end{figure}

We visualized PatchAD's response to various anomalies in Figure \ref{fig:anom_types}. Synthetic univariate time series were generated by \cite{v1_rfgs}, incorporating global point anomalies, contextual point anomalies, and pattern anomalies such as seasonal, group, and trend-based anomalies. The first row depicts the original data, while the second row displays PatchAD's output of anomaly scores against the anomaly threshold. In particular, PatchAD has the ability to detect various types of anomalies.
Figure \ref{fig:case_study} presents the practical performance of PatchAD on two real-world datasets. PatchAD exhibits the capability to detect the majority of anomalies, albeit with two anomalies remaining undetected. Furthermore, a noticeable disparity is observed between the reconstructed features by PatchAD and the original features, indicating its proficiency in learning the underlying patterns within the data, rather than simply replicating the original features.

\subsubsection{Scalability Analysis}
Figure \ref{sfig:model_size} shows the parameters, FLOPs, and latency of top-performing models.
The primary network structures employed by these models include MLP, TCN, and attention models. All of these models were trained on the SWAT dataset, and in this analysis, our model demonstrates the least parameters, \textbf{0.403M}, achieving model compactness and maintaining a certain advantage over the TCN-based TimesNet.
Meanwhile, we measured the FLOPs and latency of these models within a time window of length 105. Theoretically, our model has only 0.52G FLOPs, the lowest among all models. In actual latency tests, it achieved \textbf{\SI{7.67}{ms}}, surpassing only two pure MLP models—USAD and M2N2. These two represent cutting-edge lightweight technology but fall short of PatchAD in detection capability. Notably, TimesNet, lacking optimization for its Fourier transform operation, has the highest actual inference latency at \SI{54.44}{ms}, despite having fewer FLOPs and parameters. Conversely, GPT2-Ada, though with higher FLOPs and parameter counts, has relatively lower latency due to extensive inference optimization. In summary, our model excels in being lightweight and in theoretical and actual inference speed. It can process a 105-length time window in just \SI{7.67}{ms}, meeting the real-time requirements for real-world stream data.
\rvtwo{Based on the surveyed data\footnote{\url{https://web.eece.maine.edu/~vweaver/group/green_machines.html}}, our model can be deployed and inference on most edge devices.}

\section{Mechanism and Theory}
This section endeavors to analyze the mechanisms of PatchAD from both experimental and theoretical perspectives, further elucidating the principles underlying its effectiveness. It aims to clarify the design motivations behind PatchAD.

\subsection{Unveiling the Mechanism}\label{sec:mech}

\begin{figure*}
    \centering
    \subfloat[Case 1]{\includegraphics[width=0.2\linewidth]{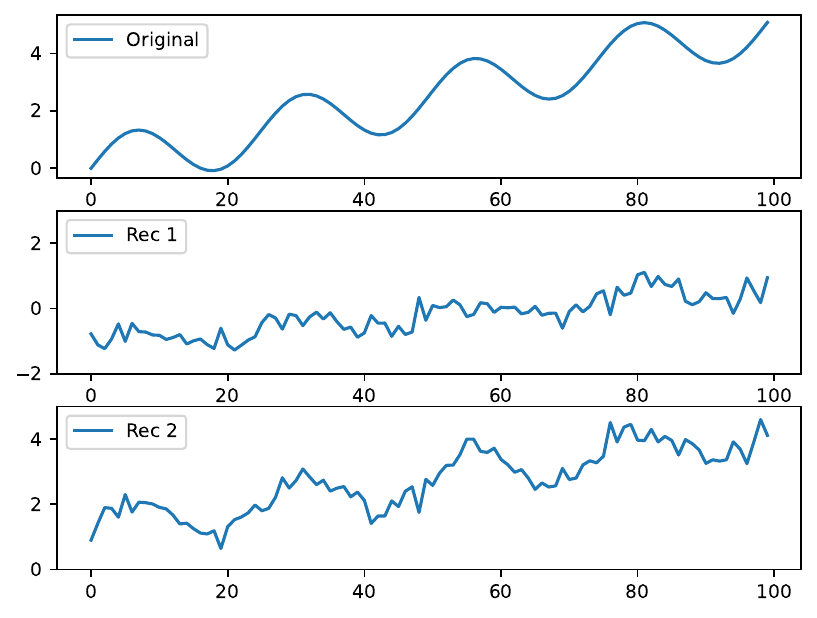}\label{sfig:case_125}}
    \subfloat[Case 2]{\includegraphics[width=0.195\linewidth]{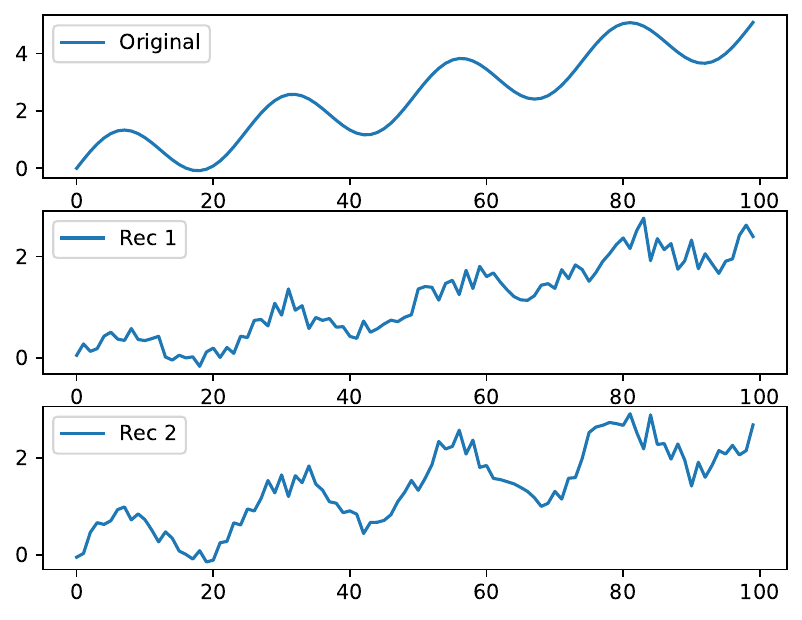}\label{sfig:case_1}}
    \subfloat[Case 3]{\includegraphics[width=0.2\linewidth]{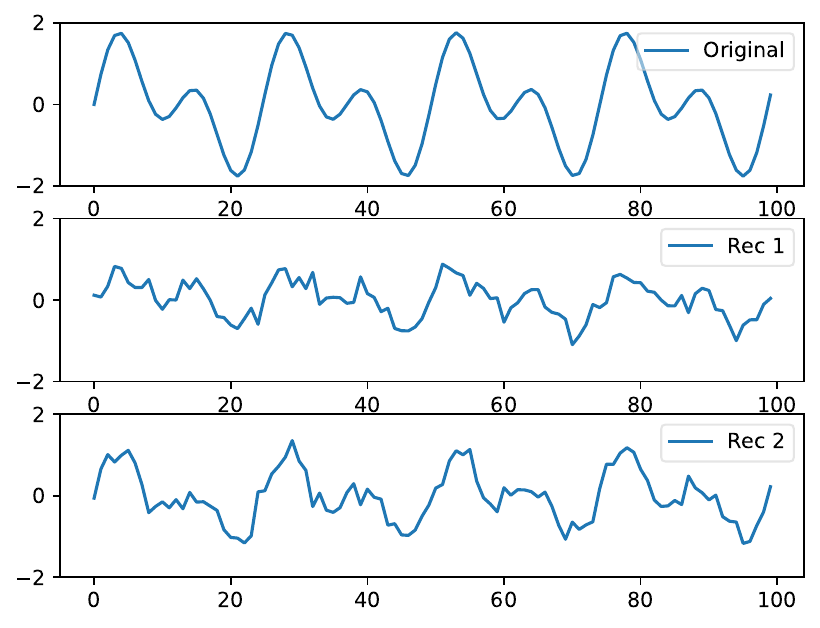}}
    \subfloat[Case 4]{\includegraphics[width=0.2\linewidth]{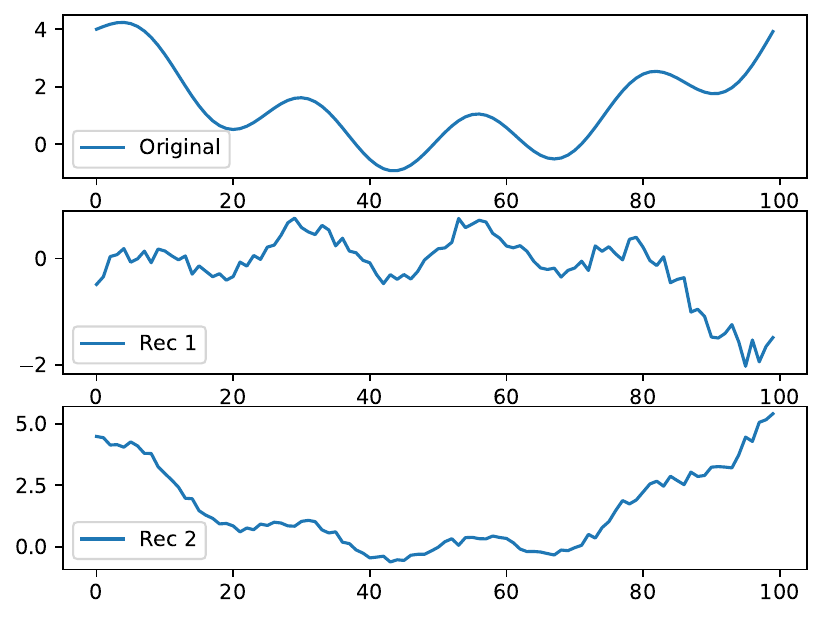}}
    \subfloat[Case 5]{\includegraphics[width=0.195\linewidth]{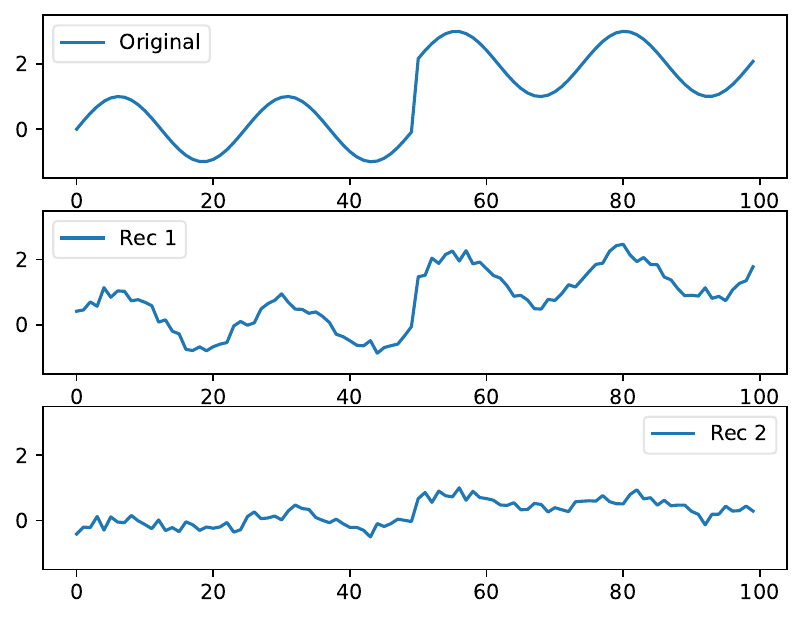}}
    
    \caption{Visualization of the features learnt by Inter and Intra Mixer.}
    \label{fig:case_trend}
\end{figure*}

Figure \ref{fig:case_trend} illustrates the knowledge acquired by PatchAD in four distinct cases. Cases 2, 3, 4 and 5 have a patch size of 1, whereas case 1 has a patch size of \([1,2,5]\). The \textbf{Original} represents the original data, while \textbf{Rec 1} and \textbf{Rec 2} correspond to the features learned by the Inter Mixer and the Intra Mixer, respectively. It is observed that the Intra Mixer captures the generalized knowledge in the data, which encompasses the overall trend information. Conversely, the Inter Mixer learnsthe detailed features of the data, compensating for any aspects that may not be captured by the Intra Mixer. A comparison between case 1 and case 2 reveals that using three distinct patch scales facilitates the acquisition of more detailed representations and enhances the generalization of the acquired features. Note that the y-axis ranges for Figure \ref{sfig:case_125} and Figure \ref{sfig:case_1} are different. Moreover, these two mixers collaborate to segregate and amplify the distinctive features effectively. The decoupling of features assists PatchAD in amplifying the anomaly-related characteristics, as indicated in Figure \ref{fig:overall}(c), thereby enhancing the detectability of anomalies. The superior performance of PatchAD can also be attributed to the varying sizes of the Inter and Intra Mixer branches, which prevent overfitting.
This is the reason why PatchAD does not require the construction of negative samples in contrastive learning, as the two different networks construct implicit positive and negative samples.

\begin{figure}[!t]
    \centering
    \includegraphics[width=0.95\linewidth]{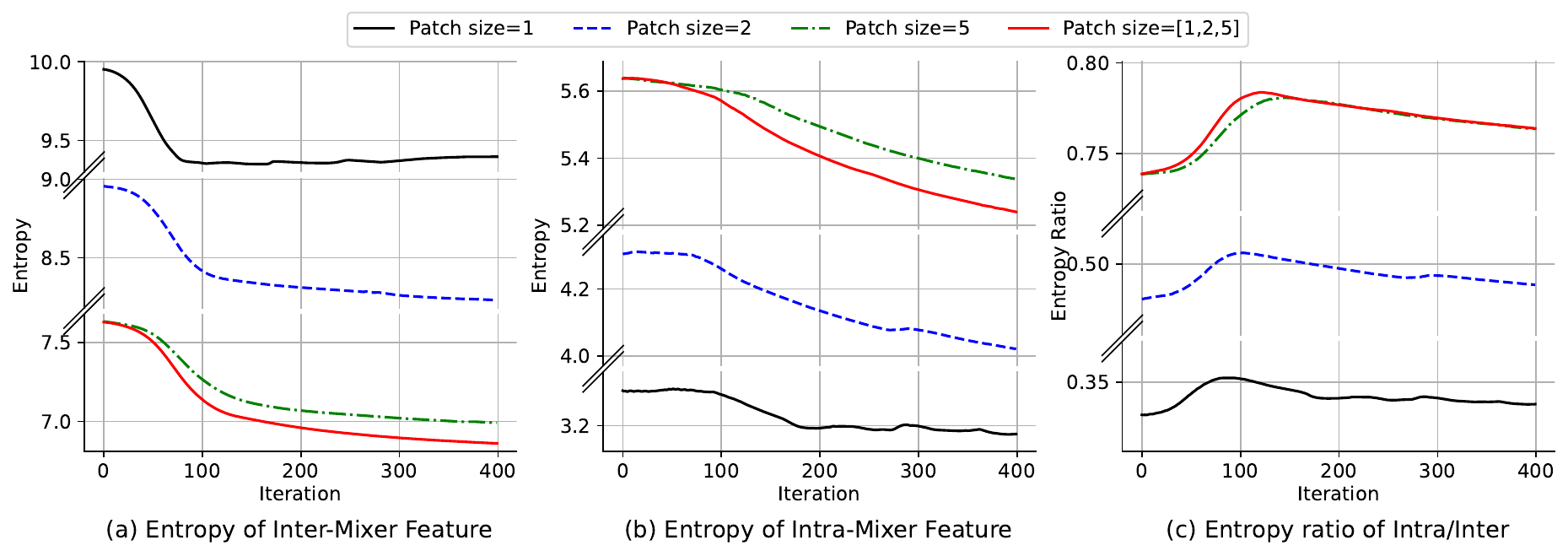}
    \caption{Entropy analysis.}
    \label{fig:entropy}
\end{figure}

Moreover, we empirically validate this conclusion through experiments. Case 1 was chosen as the baseline, and we investigated the entropy of the features acquired by the Inter and Intra Mixers under varying patch size conditions. Figures \ref{fig:entropy}(a) and \ref{fig:entropy}(b) illustrate that both the Inter Mixer and Intra Mixer display a consistent trend of decreasing entropy throughout the learning process. This observation suggests that the models effectively acquire knowledge and reduce uncertainty. Furthermore, we noticed that the Inter Mixer demonstrates the highest learning speed and exhibits the most substantial decrease in entropy when the patch size is configured as [1, 2, 5]. Additionally, as the patch size decreases, the entropy of the Inter Mixer increases, and it actually rises when the patch size is set to 1. This implies that the network is required to learn an excessive number of features. Figure \ref{fig:entropy}(c) computes the ratio of Intra to Inter entropy, and our findings indicate that as the patch size increases, the features acquired by the Intra Mixer gain greater significance, thereby accelerating model learning through the utilization of multi-scale patch sizes.

\begin{figure}[!t]
    \centering
    \includegraphics[width=0.2\linewidth]{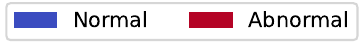}
    \includegraphics[width=0.95\linewidth]{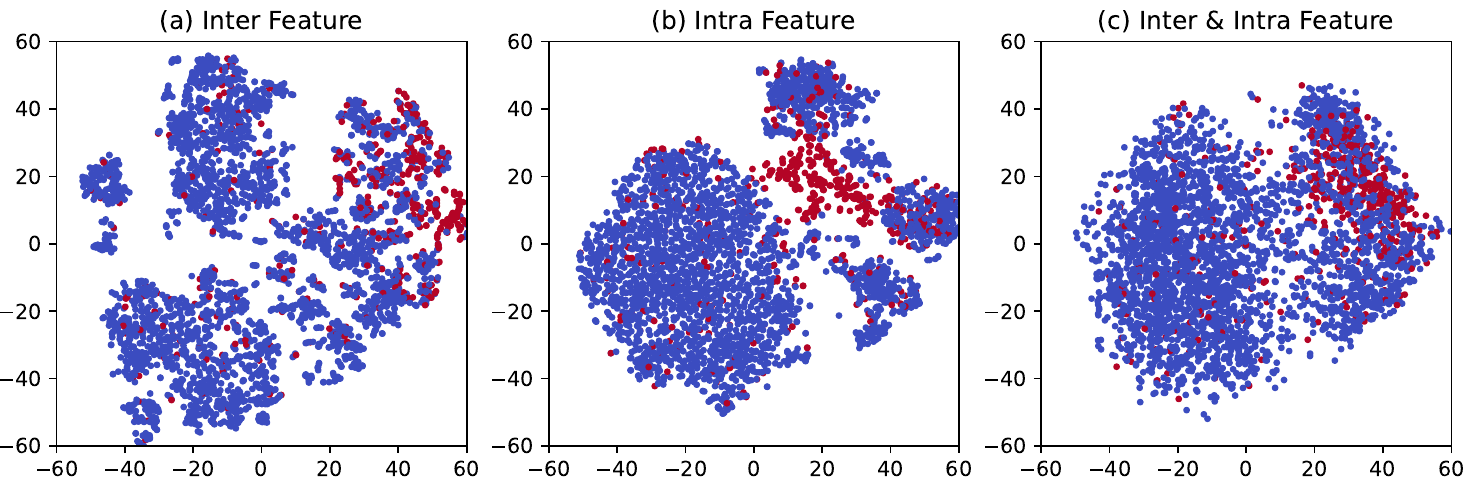}
    \caption{Latent distribution's visualization by T-SNE.}
    \label{fig:tsne}
\end{figure}

Figure \ref{fig:tsne} illustrates the visualization of feature distributions in the SWAT dataset under three conditions. It is observed that anomalies are more easily detected in the Intra Mixer view (See Figure \ref{fig:tsne}(b)), as its distinctiveness is more pronounced compared to the other two views (Figure \ref{fig:tsne} (a) \& (c)).

According to our analysis, maximizing the disparities between the Inter Mixer and Intra Mixer can enhance the model's detection capabilities. During network optimization, the Inter-Intra loss compels the Intra Mixer to acquire more generalized representations, effectively mitigating model overfitting.

\subsection{Theory Analysis}\label{sec:theory}

Considering spatial constraints and readability, further details are elaborated in Appendix \ref{app:theory}.

\section{Conclusion}
We presents PatchAD, an innovative algorithm tailored for time series anomaly detection. PatchAD developed a lightweight multi-scale, patch-based MLP Mixer architecture grounded in contrastive learning. This architecture is unique in using a purely MLP-based approach to discern inter-patch and intra-patch relationships in time series sequences. By amplifying the differences between two data views, PatchAD enhance its capability to differentiate between normal and anomalous patterns. Furthermore, PatchAD adeptly learns the relationships between different data channels and incorporates a Dual Project Constraint to avert model degradation. Moreover, the efficacy of PatchAD, including its various components, is validated through extensive experimentation. The results demonstrate that proposed lightweight PatchAD outperforms over 30 methods across eight benchmark datasets from different application scenarios.
Additionally, we have conducted a comprehensive and extensive analysis of PatchAD from both theoretical and experimental perspectives, confirming the alignment between the model's structure and its design motivations.





\bibliographystyle{IEEEtranN}
\bibliography{main.bib}


\begin{IEEEbiography}[{\includegraphics[width=1in,height=1.25in,clip,keepaspectratio]{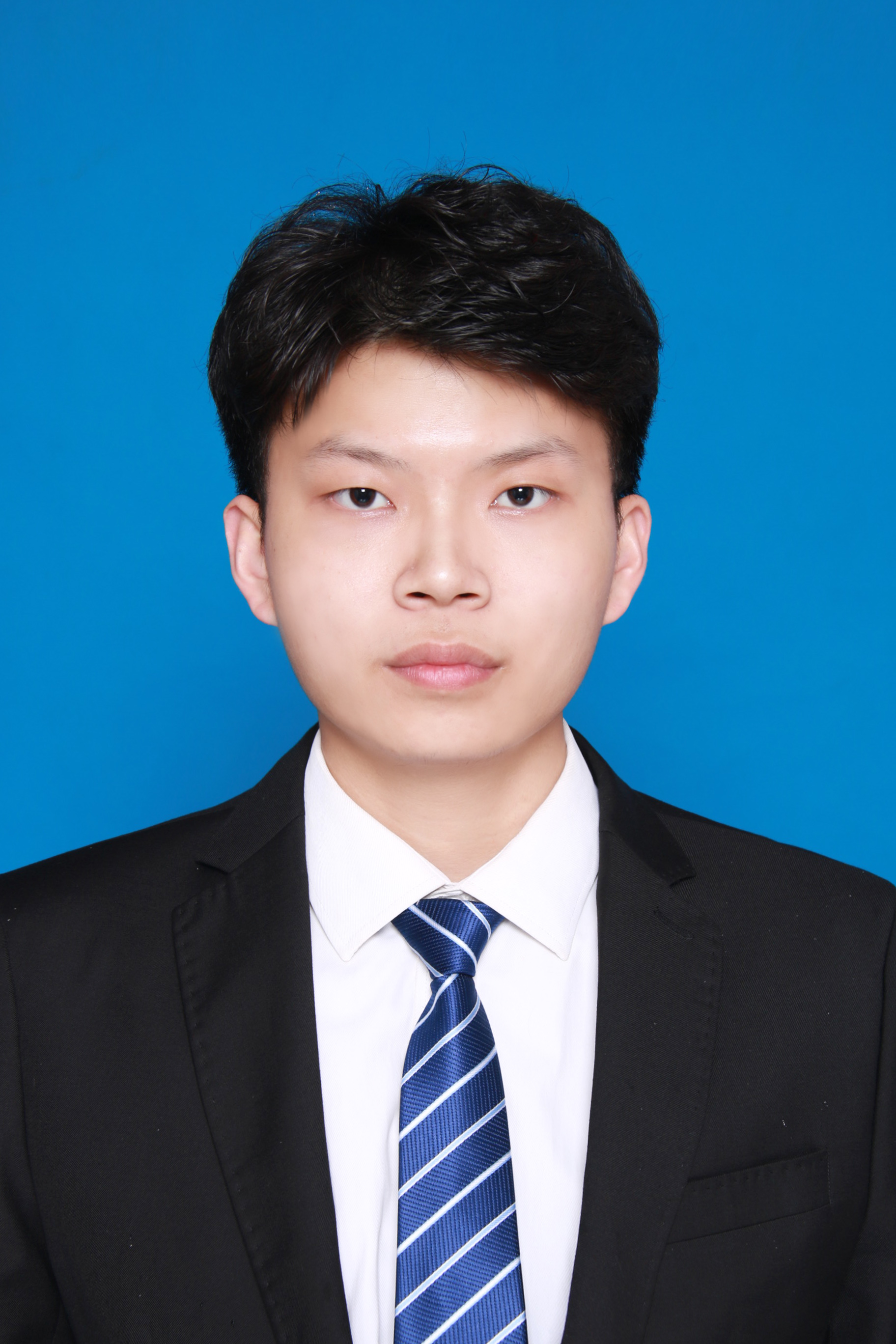}}]{Zhijie Zhong} received the B.S. degree in 2022 from the Harbin Engineering University, Harbin, China and he is currently pursuing the Ph.D. degree in the School of Future Technology, South China University of Technology, Guangzhou, China. His research interests include data mining, machine learning, time series analysis, anomaly detection, and large language model (LLM).
\end{IEEEbiography}
\begin{IEEEbiography} [{\includegraphics[width=1in,height=1.25in,clip,keepaspectratio]{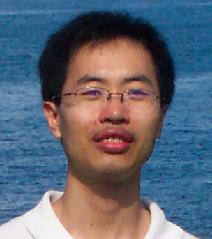}}]{Zhiwen Yu (S'06-M'08-SM'14)} is a Professor in School of Computer Science and Engineering, South China University of Technology, China. He received the Ph.D. degree from the City University of Hong Kong, Hong Kong, in 2008. Dr. Yu has authored or coauthored more than 200 refereed journal articles and international conference papers, including more than 70 articles in the journals of IEEE Transactions. His google citation is more than 10000, and h-index is 44. He is an Associate Editor of the IEEE Transactions on systems, man, and cybernetics: systems. He is a senior member of IEEE and ACM, a Member of the Council of China Computer Federation (CCF).
\end{IEEEbiography}
\begin{IEEEbiography} [{\includegraphics[width=1in,height=1.25in,clip,keepaspectratio]{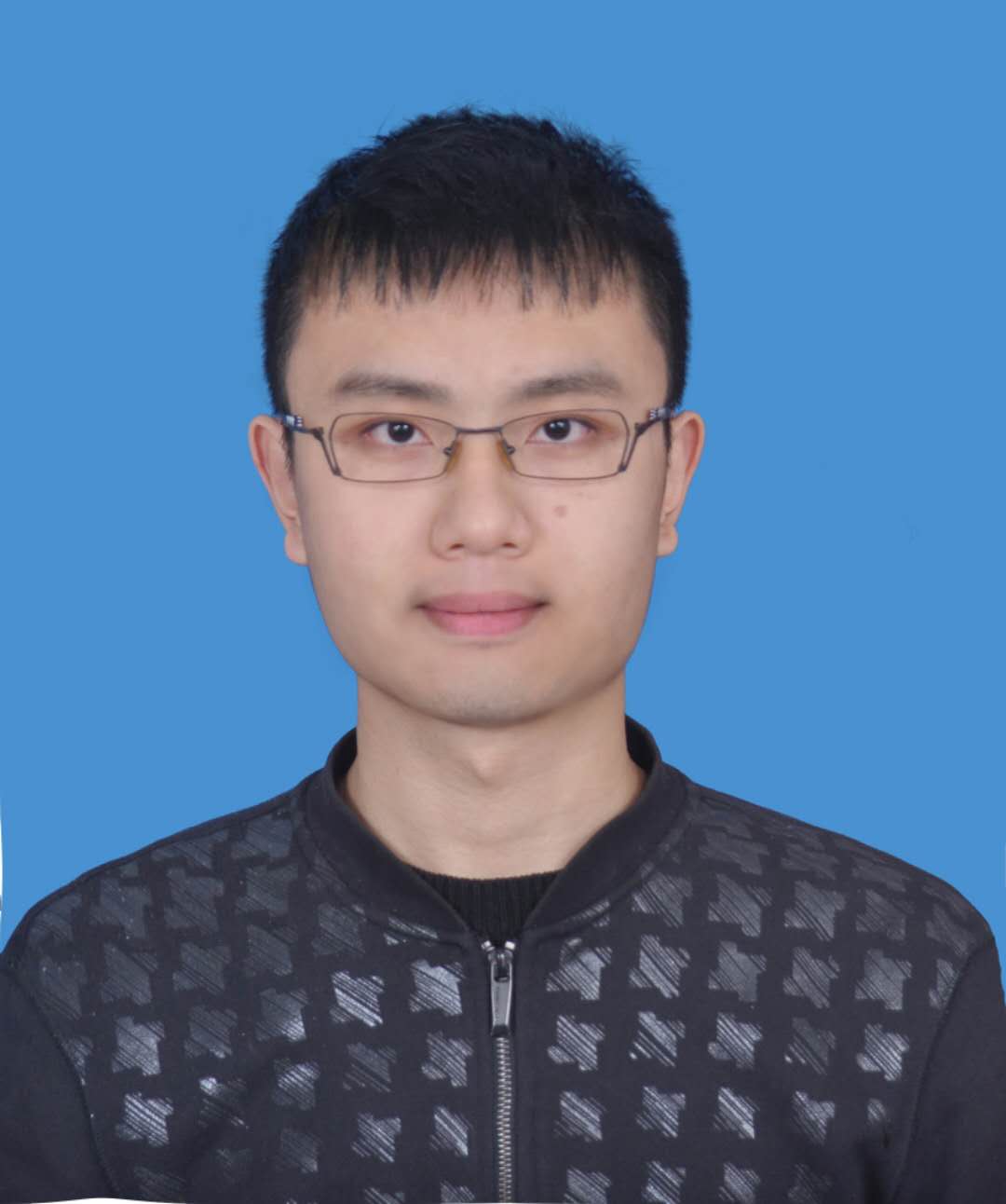}}] {Kaixiang Yang (M'21)} received the B.S. degree and M.S. degree from the University of Electronic Science and Technology of China and Harbin Institute of Technology, China, in 2012 and 2015, respectively, and the Ph.D. degree from the School of Computer Science and Engineering, South China University of Technology, China, in 2020.
He has been a Research Engineer with the 7th Research Institute, China Electronics Technology Group Corporation, Guangzhou, China, from 2015 to 2017, and has been a Postdoctoral Researcher with Zhejiang University from 2020 to 2021. He is now with the School of Computer Science and Engineering, South China University of Technology. His research interests include pattern recognition, machine learning, and industrial data intelligence.
\end{IEEEbiography}
\begin{IEEEbiography}
[{\includegraphics[width=1in,height=1.25in,clip,keepaspectratio]
{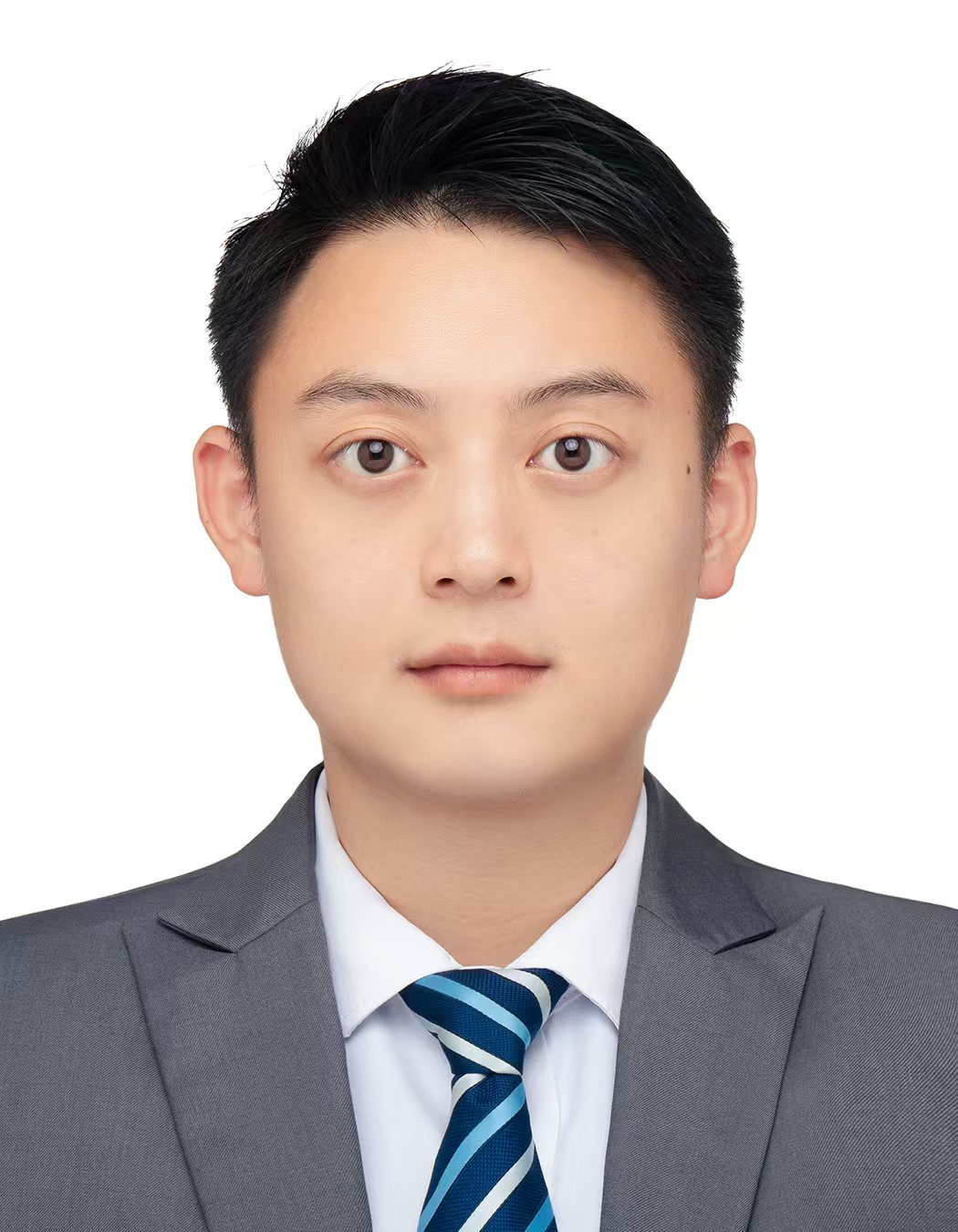}}]{Weizheng Wang} received his Ph.D. degree in Computer Science from the City University of Hong Kong (CityU) in 2025. He is currently a Postdoctoral Research Fellow in the Department of Electrical and Electronic Engineering at the Hong Kong Polytechnic University, Hong Kong SAR, China.
\end{IEEEbiography}
\begin{IEEEbiography}[{\includegraphics[width=1in,height=1.25in,clip,keepaspectratio]{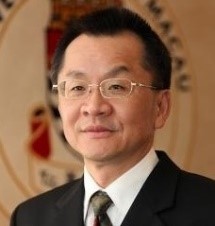}}] {C. L. Philip Chen (S'88-M'88-SM'94-F'07)}
received the M.S. degree in electrical engineering from the University of Michigan, Ann Arbor, MI, USA, in 1985, and the Ph.D. degree in electrical engineering from Purdue University, West Lafayette, IN, USA, in 1988.
He is the Chair Professor and the Dean of the School of Computer Science and Engineering, South China University of Technology, Guangzhou, China. Being a Program Evaluator of the Accreditation Board of Engineering and Technology Education (ABET) in USA, for computer engineering, electrical engineering, and software engineering programs, he successfully architects the University of Macau’s Engineering and Computer Science programs receiving accreditations from Washington/Seoul Accord through Hong Kong Institute of Engineers (HKIE), which is considered as his utmost contribution in engineering/computer science education for Macau as the former Dean of the Faculty of Science and Technology. His current research interests include cybernetics, systems, and computational intelligence.
Dr. Chen is a fellow of AAAS, IAPR, CAA, and HKIE, and a member of the Academia Europaea (AE), the European Academy of Sciences and Arts (EASA), and the International Academy of Systems and Cybernetics Science (IASCYS). He received IEEE Norbert Wiener Award in 2018 for his contribution in systems and cybernetics, and machine learnings. He is also a highly cited researcher by Clarivate Analytics in 2018 and 2019. He was a recipient of the 2016 Outstanding Electrical and Computer Engineers Award from his alma mater, Purdue University, in 1988. He was the Chair of TC 9.1 Economic and Business Systems of International Federation of Automatic Control from 2015 to 2017 and currently is a Vice President of Chinese Association of Automation (CAA). He was the IEEE Systems, Man, and Cybernetics Society President from 2012 to 2013, the Editor-in-Chief for the IEEE TRANSACTIONS ON SYSTEMS, MAN, AND CYBERNETICS: SYSTEMS from 2014 to 2019, and currently, he is the Editor-in-Chief for the IEEE TRANSACTIONS ON CYBERNETICS, and an Associate Editor of IEEE TRANSACTIONS ON ARTIFICIAL INTELLIGENCE and IEEE TRANSACTIONS ON FUZZY SYSTEMS.
\end{IEEEbiography}

\iftrue

\newpage
\appendix
This is the appendix of PatchAD: A Lightweight Patch-Based MLP-Mixer for Time Series Anomaly Detection.

\subsection{Preliminaries}

\subsubsection{MLP Mixer}\label{app:mixer}
Consider a time series data sequence \(\mathbf{X} = [\mathbf{x}_1, \mathbf{x}_2, \cdots, \mathbf{x}_T]\), where \(\mathbf{x}_t \in \mathbb{R}^D\) represents the input feature vector at time step \(t\), \(T\) is the length of the sequence, and \(D\) is the feature dimension. The objective is to learn a model \(f\) that maps the input sequence \(\mathbf{X}\) to an output sequence \(\mathbf{Y} = [\mathbf{y}_1, \mathbf{y}_2, \cdots, \mathbf{y}_T]\), where \(\mathbf{y}_t \in \mathbb{R}^{D'}\) and \(D'\) is the output feature dimension.
Subsequently, we first introduce the core part of \cite{v1_mlpmix}, and then proceed with the necessary supplementary introduction of our paper.  

\textbf{Patch Embedding}
First, the input time series sequence is partitioned into non-overlapping patches. Suppose the sequence of length \(T\) is divided into \(N\) patches, each of length \(L\), such that \(T = N\times L\). The \(i\) - th patch is denoted as \(\mathbf{p}_i = [\mathbf{x}_{(i - 1)L+1}, \mathbf{x}_{(i - 1)L + 2}, \cdots, \mathbf{x}_{iL}]\in\mathbb{R}^{L\times D}\).

Each patch is mapped to a low-dimensional embedding space through a linear projection layer. For the \(i\) - th patch \(\mathbf{p}_i\), its embedding vector \(\mathbf{e}_i\) is calculated as follows:
\begin{equation}
\mathbf{e}_i=\mathbf{W}_e\mathrm{flatten}(\mathbf{p}_i)+\mathbf{b}_e
\end{equation}
where \(\mathbf{W}_e\in\mathbb{R}^{C\times(LD)}\) is the projection matrix, \(\mathbf{b}_e\in\mathbb{R}^C\) is the bias vector, \(C\) is the embedding dimension, and \(\mathrm{flatten}(\cdot)\) is the operation that flattens a two-dimensional matrix into a one-dimensional vector. The resulting patch embedding sequence is \(\mathbf{E}=[\mathbf{e}_1,\mathbf{e}_2,\cdots,\mathbf{e}_N]\in\mathbb{R}^{N\times C}\).

\textbf{Mixer Layer}
The MLP Mixer model is composed of multiple Mixer layers stacked together. Each Mixer layer consists of two MLP sub-layers: the Token mixing MLP and the Channel mixing MLP.

\textbf{Token mixing MLP}
The Token mixing MLP is used to interact information among different patches. Before entering the Token - mixing MLP, layer normalization (Layer Normalization) is applied to the input embedding sequence \(\mathbf{E}\). Let \(\mathbf{E}'=\mathrm{LN}(\mathbf{E})\), where \(\mathrm{LN}(\cdot)\) represents the layer normalization function.

The Token mixing MLP operates on the transposed \(\mathbf{E}'^T\in\mathbb{R}^{C\times N}\). Let \(\mathbf{Z}_1^T=\mathrm{MLP}_{token}(\mathbf{E}'^T)\), where \(\mathrm{MLP}_{token}\) is a MLP containing two fully connected layers. Specifically,
\begin{equation}
\mathbf{Z}_1^T=\mathbf{W}_{token2}\sigma(\mathbf{W}_{token1}\mathbf{E}'^T+\mathbf{b}_{token1})+\mathbf{b}_{token2}
\label{eq:token_mixer}
\end{equation}
where \(\mathbf{W}_{token1}\in\mathbb{R}^{H_{token}\times N}\), \(\mathbf{b}_{token1}\in\mathbb{R}^{H_{token}}\), \(\mathbf{W}_{token2}\in\mathbb{R}^{N\times H_{token}}\), \(\mathbf{b}_{token2}\in\mathbb{R}^{N}\), \(H_{token}\) is the hidden layer dimension of the Token mixing MLP, and \(\sigma(\cdot)\) is the activation function (e.g., GELU).

Then, the intermediate output \(\mathbf{Z}_1\) is obtained through a residual connection: \(\mathbf{Z}_1=\mathbf{E}+\mathbf{Z}_1^T\).

\textbf{Channel mixing MLP}
The Channel mixing MLP is used to interact information among different channels of each patch. Similarly, layer normalization is first applied to \(\mathbf{Z}_1\) to obtain \(\mathbf{Z}_1'=\mathrm{LN}(\mathbf{Z}_1)\).

The Channel mixing MLP operates directly on \(\mathbf{Z}_1'\). Let \(\mathbf{Z}_2=\mathrm{MLP}_{channel}(\mathbf{Z}_1')\), where \(\mathrm{MLP}_{channel}\) is also a MLP containing two fully connected layers. Specifically,
\begin{equation}
\mathbf{Z}_2=\mathbf{W}_{channel2}\sigma(\mathbf{W}_{channel1}\mathbf{Z}_1'+\mathbf{b}_{channel1})+\mathbf{b}_{channel2}
\label{eq:channel_mixer}
\end{equation}
where \(\mathbf{W}_{channel1}\in\mathbb{R}^{H_{channel}\times C}\), \(\mathbf{b}_{channel1}\in\mathbb{R}^{H_{channel}}\), \(\mathbf{W}_{channel2}\in\mathbb{R}^{C\times H_{channel}}\), \(\mathbf{b}_{channel2}\in\mathbb{R}^{C}\), and \(H_{channel}\) is the hidden layer dimension of the Channel mixing MLP.

Finally, the output of the Mixer layer \(\mathbf{Z}\) is obtained through a residual connection: \(\mathbf{Z}=\mathbf{Z}_1+\mathbf{Z}_2\).

In fact, the essence of the MLP Mixer lies in using multiple MLPs to operate on different dimensions of the original data, so as to learn different types of information. In the original MLP Mixer structure, the Token mixing MLP is employed to learn information about different image patches, and the Channel mixing MLP is utilized to learn information about the RGB channels of image. Therefore, different mixers can also be abstracted as a new type of MLP that operates on different dimensions, which can be defined as:
\begin{equation}
\begin{aligned}
    &\operatorname{Mixer}(X,\text{dim = i})\\
    &=\operatorname{Transpose}(\operatorname{MLP}(\operatorname{Transpose}(X,dim = i)),\text{dim = i}), 
\end{aligned}
\end{equation}
where \(\operatorname{MLP}\) represents the Multi-Layer Perceptron, and its processing procedure is basically consistent with Eqs. \ref{eq:token_mixer} and \ref{eq:channel_mixer}. In our implementation, we adopt the residual connection and LayerNorm techniques. Finally, it can be expressed as:
\begin{equation}
     \operatorname{MLP}(\mathcal{X})=\operatorname{FC}(\operatorname{GELU}(\operatorname{FC}(\operatorname{Norm}(\mathcal{X}))))+\mathcal{X}.
\end{equation}
In addition, \(\operatorname{Transpose}(\cdot,\text{dim = i})\) denotes the change of the feature shape, which exchanges the \(i\) - th dimension with the last dimension. Through appropriate feature transposition, the MLP can process the data. Thus, \(\text{dim = i}\) implicitly indicates on which dimension the MLP acts.
There are four different Mixers in PatchAD, namely the Channel Mixer, the Inter Mixer, the Intra Mixer, and the MixRep Mixer.
The input of the Channel Mixer is \(\mathcal{N}\in \mathbb{R}^{C\times N \times D}\) or \(\mathcal{P} \in \mathbb{R}^{C \times P \times D}\). Since the channel mixer is designed to capture the channel information within different time series, the \(\operatorname{Transpose}\) operation needs to be performed on the first dimension (dimension \(C\)), swapping the dimensions of \(C\) and \(D\), so that the MLP can process the features. That is, there are transformations \(\mathcal{N}\in \mathbb{R}^{C\times N \times D}\to\mathcal{N}\in \mathbb{R}^{D\times N \times C}\) and \(\mathcal{P} \in \mathbb{R}^{C \times P \times D}\to\mathcal{P} \in \mathbb{R}^{D \times P \times C}\). After that, the newly obtained \(\mathcal{N}\) and \(\mathcal{P}\) are fed into the MLP, and the feature transformed \(\mathcal{N}\in \mathbb{R}^{D \times N \times C}\) and \(\mathcal{P} \in \mathbb{R}^{D \times P \times C}\) can be obtained. Finally, the second \(\operatorname{Transpose}\) is used to convert the dimensional shape of the features back to the original dimensional shape, \textit{i.e.}, \(\mathcal{N}\in \mathbb{R}^{D\times N \times C}\to\mathcal{N}\in \mathbb{R}^{C\times N \times D}\) and \(\mathcal{P} \in \mathbb{R}^{D \times P \times C}\to\mathcal{P} \in \mathbb{R}^{C \times P \times D}\).

The fundamental processes of other Mixers follow a similar pattern. Initially, the tensor's shape is transformed, enabling the MLP to operate on it. Subsequently, the tensor is reverted to its original shape for subsequent processing steps.

For instance, the Inter Mixer and Intra Mixer are designed to process \(\mathcal{N}\) and \(\mathcal{P}\) respectively. In a like manner, the shapes of \(\mathcal{N}\) and \(\mathcal{P}\) are first modified. Specifically, \(\mathcal{N}\) undergoes a transformation from \(\mathbb{R}^{C\times N \times D}\) to \(\mathbb{R}^{C\times D \times N}\), and \(\mathcal{P}\) is changed from \(\mathbb{R}^{C \times P \times D}\) to \(\mathbb{R}^{C \times D \times P}\). Subsequently, the MLP is tasked with learning the inter patch and intra patch knowledge. Once this learning process is complete, the tensors are transformed back to their original configurations.

The MixRep Mixer serves a function analogous to that presented in \cite{v1_simclr}. This mixer allows the model to discover a greater range of invariant information by extracting unified information inside a low-dimensional space.  Such information includes recurring cycles and trend-related patterns in time series data. In a similar vein, the MixRep Mixer focuses on learning unified information along the \(D\) dimension. This is achieved by first reshaping the tensor, subjecting it to processing by the MLP, and ultimately restoring it to its original form. (Notably, direct processing by the MLP is feasible as it inherently operates on the last dimension by default.) 

\subsubsection{Contrastive Learning for Time Series}\label{app:contrastive}

This section introduces the theoretical foundations of contrastive learning (CL) in time series representation learning and its application to anomaly detection. 

Contrastive learning aims to learn discriminative representations by maximizing similarity between positive pairs (samples from the same time series) and minimizing similarity between negative pairs (samples from different time series). For a time series dataset \(\mathcal{D} = \{ \mathbf{x}_1, \mathbf{x}_2, \dots, \mathbf{x}_N \}\), where each \(\mathbf{x}_i \in \mathbb{R}^{T \times D}\) with length \(T\) and \(D\) features, CL constructs pairs through data augmentation (e.g., masking, time warping).  

The contrastive loss function, such as the InfoNCE loss \cite{TS_CP2}, is defined as:  
\begin{equation}
    \mathcal{L}_{\text{CL}} = -\frac{1}{N} \sum_{i=1}^N \log \frac{\exp \left( \text{sim}(f(\mathbf{x}_i^+), f(\mathbf{x}_i^-)) / \tau \right)}{\sum_{j=1}^{N} \exp \left( \text{sim}(f(\mathbf{x}_i^+), f(\mathbf{x}_j^-)) / \tau \right)},
    \label{eq:cont}
\end{equation}
where \(\mathbf{x}_i^+\) and \(\mathbf{x}_i^-\) denote augmented positive and negative samples of \(\mathbf{x}_i\), \(f(\cdot)\) is the encoder network, \(\text{sim}(\cdot, \cdot)\) is a similarity metric (\textit{e.g.}, cosine similarity), and \(\tau\) is the temperature hyperparameter.

Additionally, a more straightforward implementation approach lies in minimizing the feature distance between positive and negative samples. To achieve this goal effectively, we can employ distance measures such as the L2 distance or the KL divergence. By using these measures, we can optimize the similarity between \(f(\mathbf{x}_i^+)\) and \(f(\mathbf{x}_i^-)\) so that they are more closely aligned \cite{simCSE}. This optimization process can be directly carried out as follows: 
\begin{equation}
    \mathcal{L}= \text{sim}(f(\mathbf{x}_i^+), f(\mathbf{x}_i^-)).
\end{equation}

In fact, most models based on contrastive learning operate under the same assumption that different data augmentations or transformations do not alter the invariant information \cite{v1_simclr,v1_simsiam,dc_detector,anomaly_trans,v1_byol}. The samples generated through data augmentation can be referred to as negative samples, represented as \(\mathbf{x}_i^-\) in Eq. \eqref{eq:cont}; while positive samples may consist of the original data or other augmented data. For instance, in images, applying rotational transformations and horizontal flips to a picture of a cat, or even adding noise, does not change the semantic information that there is a cat in the image. Similarly, in time series, introducing slight noise does not alter the overall trend information (although detailed information may be compromised). This assumption is applicable to time series prediction and representation tasks \cite{TS2Vec}; however, it may not always hold true for time series anomaly detection. In extreme cases, altering the signal of a particular channel could be interpreted as a peak anomaly \cite{v1_couta}.

Due to the insufficient diversity of manually designed negative samples, which may contain harmful samples for model training, an alternative approach in contrastive learning is to utilize uniquely designed feature variations to achieve a similar effect to data augmentation, thereby preserving invariant information. For example, in \cite{v1_byol}, a contrastive learning approach is proposed that does not require negative samples by constructing two distinct network architectures. Similarly, \cite{v1_simsiam} suggests employing gradient stopping as a substitute for the network model in contrastive learning, further validating its effectiveness through \cite{MoCo}.

In the domain of time series anomaly detection, \cite{anomaly_trans} hypothesizes that anomalies exhibit lower similarity to neighboring time points, thereby constructing a prior association as a data variation method to eliminate the need for negative samples. Furthermore, the followed work of \cite{dc_detector} also continues the design philosophy of not constructing negative samples, as it employs two distinct attention models to learn the correlation of time series data across different representation spaces.

Within contrastive learning, the model degradation problem (also known as feature collapse or trivial solutions) may arise during the training process. This phenomenon occurs when the feature vectors of all samples converge to the same point or direction, resulting in an inability to distinguish between different categories or samples \cite{v1_byol,v1_simsiam}. Such degradation diminishes the model's capacity to learn effective feature representations, ultimately leading to a significant decline in performance on downstream tasks. This is often caused by defects in negative sample design, deviations in loss function optimization, and instability in training dynamics.

PatchAD draws upon prior works \cite{v1_byol,v1_simsiam,anomaly_trans,dc_detector} in its overall design philosophy for contrastive learning. It employs the Inter Patch Mixer and Intra Patch Mixer to facilitate a contrastive learning approach that does not require negative samples. Additionally, the design of the MixRep Mixer and optimization constraints helps prevent deviations during the optimization process, enhancing training stability and thereby reducing the likelihood of model degradation in contrastive learning.





\subsection{Theory Analysis}\label{app:theory}
To perform a theoretical analysis of PatchAD's effectiveness, we want to prove that normal time series data points exhibit reduced information entropy. We have made the following assumptions based on the theoretical analysis presented in previous studies.
Subsequently, we further theoretically prove why the Inter Patch and Intra Patch Encoders can focus on different types of information and explain how PatchAD can utilize different mixers to enhance its ability to uncover anomalies. 

\textbf{Assumption 1}: 
Anomalies represent deviations from the normal pattern. These deviations may arise from unexpected events (\textit{e.g.}, sensor faults, cyberattacks), noise increase (\textit{e.g.}, hardware degradation), or rare patterns (\textit{e.g.}, extreme values outside the normal range). And variance (\(\sigma^2\)) quantifies the spread of data around the mean. From the perspective of statistics, larger variance implies wider dispersion of data points, higher uncertainty in predictions, or greater deviation from the expected behavior \cite{genIAS}.

\textbf{Assumption 2}:  
Normal time series data points are independently and identically distributed (i.i.d.) according to a Gaussian distribution with mean \(\mu\) and variance \(\sigma_1^2\), \textit{i.e.}, \(X \sim \mathcal{N}(\mu, \sigma_1^2)\). Abnormal time series data points are i.i.d. with the same mean \(\mu\) but larger variance \(\sigma_2^2 > \sigma_1^2\), \textit{i.e.}, \(Y \sim \mathcal{N}(\mu, \sigma_2^2)\) \cite{genIAS}.

\textbf{Proof 1:}
The differential entropy \(H\) of a Gaussian distribution \(\mathcal{N}(\mu, \sigma^2)\) is given by:
\(
H(X) = \frac{1}{2} \ln(2\pi e \sigma^2).
\)
For the normal time series \(X\), the entropy is:  
\(
H(X) = \frac{1}{2} \ln(2\pi e \sigma_1^2).
\)
For the abnormal time series \(Y\), the entropy is:  
\(
H(Y) = \frac{1}{2} \ln(2\pi e \sigma_2^2).
\)
Since \(\sigma_2^2 > \sigma_1^2\), we have:  
\(
\ln(\sigma_2^2) > \ln(\sigma_1^2) \implies \ln(2\pi e \sigma_2^2) > \ln(2\pi e \sigma_1^2).
\)
Thus, \(H(Y) > H(X)\).  
Abnormal time series exhibit higher entropy because their larger variance reflects greater uncertainty and disorder compared to normal time series. This aligns with the intuition that anomalies introduce irregularity, widening the distribution and increasing entropy. Similar reasoning applies to other distributions (\textit{e.g.}, uniform distributions over wider intervals).

\textbf{Prerequisite 1}:
\begin{enumerate}
    \item Time series window: It has a length of \(T\), which is divided into \(N\) non-overlapping patches, and each patch has a length \(P\) (i.e., \(T = N\times P\)).
    \item The Inter Patch Encoder applies a MLP in the \(N\) dimension (across patches). The shape of its input is \([B, C, N, P]\). While the Intra Patch Encoder applies an MLP in the \(P\) dimension (within a single patch). The shape of its input is \([B, C, P, N]\), and the output shape remains unchanged.
    \item \(X_{\text{inter}}\): The output features of the Inter Patch Encoder. \(X_{\text{intra}}\): The output features of the Intra Patch Encoder.
\(H(X)\): The information entropy of feature \(X\). \(\sigma_{\text{inter}}^2\): The variance of the output of the Inter Patch Encoder. \(\sigma_{\text{intra}}^2\): The variance of the output of the Intra Patch Encoder.
\end{enumerate}

\textbf{Proof 2}:
The Inter Patch Encoder and the Intra Patch Encoder learn detailed features and general knowledge respectively. Meanwhile, different types of anomalies have different impacts on the output features. 

To simplify the proof, the feature processing of PatchAD can be simplified to the feature processing by the Inter Mixer and the Intra Mixer, because what other mixers learn are other types of information. For example, the Channel Mixer is only related to channel information.

Next, assume that both mixers are composed of MLP with linear transformations, with the weight matrix \(W\) and the bias \(b\).

For the Inter Mixer, which acts on the \(N\) dimension and performs feature transformation on the channel \(c\) of each patch simultaneously:
\(
    X_{\text{inter}}^{c} = W_{\text{inter}} \cdot X^{c}_{1:N} + b_{\text{inter}},
\)
where \(X^{c}_{1:N} \in \mathbb{R}^N\) represents the features of all patches in channel \(c\).

Similarly, the Intra Mixer acts on the \(P\) dimension, thus performing feature transformation on the channel \(c\) of a single patch simultaneously: 
\(
      X_{\text{intra}}^{c} = W_{\text{intra}} \cdot X^{c}_{1:P} + b_{\text{intra}},
\)
where \(X^{c}_{1:P} \in \mathbb{R}^P\) represents the feature composed of all time points of a single patch in channel \(c\).


According to Assumption 2, different types of features can be regarded as Gaussian distributions. At this time, the variance of the distribution is independent of the mean. Therefore, without affecting the final conclusion, it can be assumed that the mean of each row of the weight matrix \(W\) is 0 and the operations are simplified independently. Then the variance after the linear transformation is:  
\begin{equation}
    \text{Var}(WX + b) = \|W\|_F^2 \cdot \text{Var}(X),
\end{equation}
where \(\|W\|_F^2\) is the square of the Frobenius norm of the weight matrix, reflecting the amplification/attenuation effect of the weights on the input.


Subsequently, the variance of the features of the Inter Mixer can be calculated.
\begin{equation}
      \sigma_{\text{inter}}^2 = \|W_{\text{inter}}\|_F^2 \cdot \text{Var}(X_{\text{inter}}^{c})
      \label{eq:inter_sigam}
\end{equation}
Since the input \(X_{\text{inter}}^{c}\) is the feature of multiple patches, if there are abnormal patches with variance \(\sigma_2^2\) and the variance of the rest is \(\sigma_1^2\), then:  
\begin{equation}
    \text{Var}(X_{\text{inter}}^{c}) = 
     \frac{1}{N} \left( k \sigma_2^2 + (N - k) \sigma_1^2 \right),
\end{equation}
where \(k\) is the number of abnormal patches.

Similarly, the variance of the features of the Intra Mixer can be obtained:  
\begin{equation}
    \sigma_{\text{intra}}^2 = 
    \|W_{\text{intra}}\|_F^2 \cdot \text{Var}(X_{\text{intra}}^{c}).
    \label{eq:intra_simga}
\end{equation}
At this time, the input \(X_{\text{intra}}^{c}\) consists of time points within a single patch. If this patch contains an anomaly, its variance is \(\sigma_2^2\); otherwise, it is \(\sigma_1^2\). Eventually, the variance of the features of the Intra Mixer can be derived as follows:
\begin{equation}
    \text{Var}(X_{\text{intra}}^{c}) = 
    \frac{1}{P} \left( m \sigma_2^2 + (P - m) \sigma_1^2 \right),
\end{equation}
where \(m\) is the number of abnormal time points within a single patch.


Subsequently, a comparison is made between the variances of the two. Since the parameters of the fully connected network are initialized by the He initialization \cite{he_init}, it is assumed that the amplification effects of the weight matrices are basically the same, that is, \(\|W_{\text{inter}}\|_F^2=\|W_{\text{intra}}\|_F^2 = \lambda\).

It is reasonable to suppose that the anomalies are concentrated in a small number of patches when the ratio of anomalies falls under normal conditions.  Currently, the Inter Mixer's output variance is:   
\begin{equation}
    \sigma_{\text{inter}}^2 \approx 
    \lambda \left( \sigma_1^2 + \frac{k}{N} (\sigma_2^2 - \sigma_1^2) \right).
    \label{eq:inter_sigma}
\end{equation}  
While within the abnormal patch, the output variance of the Intra Mixer is
\begin{equation}
    \sigma_{\text{intra}}^2 \approx 
    \lambda \left( \sigma_1^2 + \frac{m}{P} (\sigma_2^2 - \sigma_1^2) \right).
    \label{eq:intra_sigma}
\end{equation}


Since the Inter Mixer aggregates the impacts of anomalies from all patches, while the Intra Mixer only processes a single patch, and at this moment, \(k < N\) and \(m \ll P\) are satisfied. For instance, if the length of the time window is 128, \(N = 4\), and there is an anomaly in only one patch, then \(k = 1\). In this case, the value of \((N - k)\sigma_1^2\) in Eq.(\ref{eq:inter_sigma}) has a greater influence than the value of \((P - m)\sigma_1^2\) in Eq.(\ref{eq:intra_sigma}). Similarly, the value of \(k\sigma_2^2\) is close to that of \(m\sigma_2^2\). Consequently, 
\begin{equation}
    \sigma_{\text{inter}}^2 > \sigma_{\text{intra}}^2 \implies H(X_{\text{inter}}) > H(X_{\text{intra}}),
\end{equation} 
In other words, the output features of the Intra Mixer typically have a lower information entropy, whereas the output features of the Inter Mixer typically have a greater information entropy when the number of anomalies is modest.  This suggests that while the Intra Mixer learns a more comprehensive knowledge representation, the Inter Mixer learns specific features.  Actually, this makes sense as well because the Inter Mixer must learn from more patches, which increases its uncertainty and information entropy. 
In the meantime, we examine a less probable situation in which abnormalities are evenly dispersed throughout the entire.  Stated differently, there are a few aberrant places in every patch.  In such cases, the criterion \(m/P\approx k/N\) is satisfied and it can be assumed that there are anomalies throughout the time series window: 
\begin{equation}
  \sigma_{\text{inter}}^2 \approx \sigma_{\text{intra}}^2 \implies H(X_{\text{inter}}) \approx H(X_{\text{intra}}).
\end{equation}  

Consequently, by integrating these two different cases, we can arrive at the final conclusion that \(H(X_{\text{inter}}) \leq H(X_{\text{intra}})\). Thus, we have obtained the result of \textbf{Proof 2}.


\textbf{Proof 3}: The salience of anomalies differs in the Inter Mixer and the Intra Mixer.

The extent of change in the feature distribution, or the information gain caused by anomalies, is the first way to describe anomaly salience.
\begin{equation}
\Delta H = H(X_{\text{abnormal}}) - H(X_{\text{normal}})
\end{equation}  
Subsequently, calculate \(\Delta H_{\text{inter}}\) and \(\Delta H_{\text{intra}}\) for the Inter Mixer and the Intra Mixer respectively.
The information gain of the Inter Mixer is: 
\(
  \Delta H_{\text{inter}} = \frac{1}{2} \ln\left(\frac{\sigma_{\text{inter,abnormal}}^2}{\sigma_{\text{inter,normal}}^2}\right),
\)
where \textbf{normal} and \textbf{abnormal}, respectively, indicate the presence and absence of abnormalities; this will be the case going forward. Additionally, it is simple to acquire based on the earlier proof method (\textbf{Proof 2}):
\begin{equation}
  \sigma_{\text{inter,normal}}^2 = \lambda \sigma_1^2, \quad \sigma_{\text{inter,abnormal}}^2 = \lambda \left( \sigma_1^2 + \frac{k}{N} (\sigma_2^2 - \sigma_1^2) \right).
\end{equation}  

Similarly, the information gain of the Intra Mixer can be calculated by the following:
\(
  \Delta H_{\text{intra}} = \frac{1}{2} \ln\left(\frac{\sigma_{\text{intra,abnormal}}^2}{\sigma_{\text{intra,normal}}^2}\right),
\)
where,
\begin{equation}
  \sigma_{\text{intra,normal}}^2 = \lambda \sigma_1^2, \quad \sigma_{\text{intra,abnormal}}^2 = 
  \lambda \left( \sigma_1^2 + \frac{m}{P} (\sigma_2^2 - \sigma_1^2) \right).
\end{equation}

Similarly, assume the most common scenario where anomalies are concentrated in a few patches (\(k = 1, m = 1\)) and \(1\ll P\ll N\). Then,
\begin{align}
\Delta H_{\text{inter}} \approx \frac{1}{2} \ln\left(1 + \frac{\sigma_2^2 - \sigma_1^2}{N \sigma_1^2}\right), \\ \Delta H_{\text{intra}} \approx \frac{1}{2} \ln\left(1 + \frac{\sigma_2^2 - \sigma_1^2}{P \sigma_1^2}\right).
\end{align} 
Given that \(N > P\), or the number of patches \(N\), is typically substantially greater than the length of a single patch \(P\), this means that: 
\begin{equation}
\Delta H_{\text{inter}} < \Delta H_{\text{intra}}.
\end{equation}  
It should be noted that if the length of a single patch is greater than the number of patches, \textit{i.e.}, \(P > N\), then the opposite conclusion is obtained, namely:
\(
    \Delta H_{\text{inter}} > \Delta H_{\text{intra}}.
\)

As a result, abnormalities are typically more noticeable in the Intra Mixer than the Inter Mixer at this time due to their relative scarcity.  This aligns with the behavior described in Sec. \ref{sec:mech} as well.  This suggests that the capabilities of the Intra Mixer and the Inter Mixer can be improved by improving the Inter-Intra Discrepancy, which would facilitate the detection of anomalies.  It makes sense since anomalies tend to appear in a small number of patches, and since they deviate more from the overall features at this time, they are simpler to identify.

\textbf{Analysis of Model Collapse.}
Although both Mixers receive the same input, PatchAD does not suffer from trivial solutions (model collapse). SimSiam \cite{v1_simsiam} and DCdetector \cite{dc_detector} have demonstrated that using a gradient stop to optimize two separate branches can prevent model collapse. However, our experiments show that while this method does not lead to trivial solutions, the model does not converge to a better result. In Sec. \ref{sec:ablation}, we analyze the introduced Dual Project Constraint. From Fig. \ref{fig:contraint_loss}, we found that the final loss is approximately 0.6 without the constraint, while it converges to about 0.68 with the Dual Project Constraint. This not only proves that our model does not collapse during training but also indicates that the added constraint enables the model's parameters to converge to a more optimal state.

Theoretically, our two branches have different structures, and thus they extract different features. According to the unified perspective proposed in \cite{why_simsiam}, a branch's output can be represented as $\mathbf{Z}$, which can be decomposed into two parts: $\mathbf{Z}=c+r$, where $c=\mathbb{E}[\mathbf{Z}]$ is the central vector defined as the center of the entire representation space, and $r$ can be seen as a radius vector. When model collapse occurs, all features $\mathbf{Z}$ tend toward the central vector $c$, meaning $c$ dominates $r$. Our two branches can be represented as $\mathcal{N}=c_{\mathcal{N}}+r_{\mathcal{N}}$ and $\mathcal{P}=c_{\mathcal{P}}+r_{\mathcal{P}}$. If the two branches have a symmetric structure, i.e., $c_{\mathcal{N}}=c_{\mathcal{P}}$, the distance between the two spaces is $\mathcal{N}-\mathcal{P}=r_\mathcal{N}-r_\mathcal{P}$. Since $r_{\mathcal{N}}$ and $r_{\mathcal{P}}$ are from the same input sample, the model tends to collapse. However, due to the different structures of the two branches in PatchAD, $r_\mathcal{P}$ and $r_{\mathcal{N}}$ are similar, and the centers of the two branches are less likely to be the same.

Furthermore, in a previous proof, we showed that the feature variances of the two branches are defined by Eq. (\ref{eq:inter_sigma}) and Eq. (\ref{eq:intra_sigma}), respectively. At the end of \textbf{Proof 2}, we proved that these two variances are typically not identical. This demonstrates that even if specific model parameters cause the feature means of the two branches to be close, their differing variances will ultimately lead to different central vectors for the two branches. Therefore, due to our asymmetric design, PatchAD is highly unlikely to fall into model collapse.

\subsection{Algorithm's Pseudo-code}
The pseudocode of PatchAD refers to Algorithm. \ref{alg:patch_ad}.
\definecolor{codeblue}{rgb}{0.25,0.5,0.5}
\lstset{
    language=Python,
    basicstyle=\tiny\ttfamily,
    keywordstyle=\bfseries,
    commentstyle=\itshape,
    showstringspaces=false,
    columns=flexible,
}
\begin{algorithm}[h]
\caption{PatchAD's Python-like code}
    
\begin{lstlisting}[language=python]
PE = PositionEmbedding(d_model) # Postional Embedding
VE_n, VE_p = Linear(dim1, d_model), Linear(dim2, d_model) 
# Value Embedding of N and P views

# Patching and Embedding
x = PE(x)
x_n = rearange(x, 'b (n p) c -> b c n p', p=patch_size)
x_p = rearange(x, 'b (n p) c -> b c p n', p=patch_size)
x_n = VE_n(x_n)
x_p = VE_p(x_p)
# x_n's shape (B, c_dim, n_dim, d_model)
# x_p's shape (B, c_dim, p_dim, d_model)

Channel_Mixer = MLP(c_dim, transpose_dim=1)
Inter_Mixer = MLP(n_dim, transpose_dim=2)
Intra_Mixer = MLP(p_dim, transpose_dim=2)
MixRep_Mixer = MLP(d_model, transpose_dim=3)

# Patch Mixer Encoding
x_inter = Channel_Mixer(x_n)
x_intra = Channel_Mixer(x_p)
x_inter = Inter_Mixer(x_inter)
x_intra = Intra_Mixer(x_intra)
x_inter = MixRep_Mixer(x_inter)
x_intra = MixRep_Mixer(x_intra)

# Dual Project
x_inter = x_inter.mean(1)
x_intra = x_intra.mean(1)
Inter_Proj = MLP(d_model)
Intra_Proj = MLP(d_model)
x_inter2 = Inter_Proj(x_inter)
x_intra2 = Intra_Proj(x_intra)
# Ouput shape (B, n_dim, d_model) and (B, p_dim, d_model)
# Upsample to (B, L, d_model)

rec1 = Rec1(x_inter)
rec2 = Rec2(x_intra)
rec_x = rec1 + rec2

# Update
def loss_fn(a, b):
    return KL(a, b.detach()) + KL(b.detach(), a)

def inter_intra_loss_fn(n, p):
    return loss_fn(n, p) - loss_fn(p, n)
    
loss1 = inter_intra_loss_fn(x_inter, x_intra)

loss_n_proj = inter_intra_loss_fn(x_inter, x_intra2)
loss_p_proj = inter_intra_loss_fn(x_inter2, x_intra)
loss_proj = loss_n_proj + loss_p_proj
loss_rec = MSE(rec_x, x)

final_loss = (1-c) * loss1 + c * loss_proj + loss_rec
# backward and optimize
\end{lstlisting}
\label{alg:patch_ad}
\end{algorithm}

\subsection{Experiments}

\subsubsection{Datasets}\label{app:datasets}
To extensively validate the generalizability of the algorithm, we selected time series datasets from diverse domains, encompassing web monitoring, server performance monitoring, IoT, and health monitoring, among others.
We applied seven multivariable and one univariable datasets for evaluation: (1) MSL (2) SMAP \footnote{MSL \& SMAP: \url{https://github.com/ML4ITS/mtad-gat-pytorch}}(3) PSM (4) SWAT \footnote{\url{https://itrust.sutd.edu.sg/itrust-labs\_datasets}} (5) WADI \footnote{\url{https://itrust.sutd.edu.sg/testbeds/water-distribution-wadi}} (6) NIPS-TS-SWAN (SWAN) 
(7) NIPS-TS-GECCO (GECCO)
\footnote{Others:\url{http://github.com/DAMO-DI-ML/KDD2023-DCdetector}}. 
(8) UCR \footnote{\url{https://www.cs.ucr.edu/~eamonn/time_series_data_2018/UCR_TimeSeriesAnomalyDatasets2021.zip}}
Table \ref{tab:data_set} shows the details. 
Here, \#Training and \#Test represent the lengths of the time series used for training and testing, respectively. Dimension denotes the number of channels in the time series, such as signals from different sensors. Anomaly ratio indicates the average proportion of anomalies within the dataset.

\begin{table}[htb]
	\centering
	\caption{Details of benchmark datasets. }
    \resizebox{0.95\linewidth}{!}{
	\begin{tabular}{c|r|r|r|r}
		\toprule
		Dataset  & \#Training & \#Test (Labeled) & Dimension & Anomaly ratio (\%) \\
		\midrule
    MSL   & 58317 & 73729 & 55    & 10.5  \\
    SMAP  & 135183 & 427617 & 25    & 12.8  \\
    PSM & 132481 & 87841 & 25 & 27.8 \\
    SWAT & 99000 & 89984 & 26 & 12.2 \\
    WADI & 241921 & 34561 & 123 & 5.74\\
    NIPS-TS-SWAN & 60000 & 60000 & 38 & 32.6 \\
    NIPS-TS-GECCO & 69260 & 69261 & 9 & 1.1 \\
		\bottomrule
	\end{tabular}
}
	\label{tab:data_set}
\end{table}

\subsubsection{Settings}
\textbf{Data Source Clarification}: Due to substantial device variations, to ensure a fair comparison with anomaly transformer (AnomalyTrans) and DCdetector, all results mentioned in the paper concerning them are obtained through our reimplementation of their codes. The results of AnomalyTrans and DCdetector in Table \ref{tab:comparison} and Table \ref{tab:comparison_nips_ts} are our reproductions.
The results presented in Table \ref{tab:comparison_nips_ts} primarily derive from \cite{dc_detector}, with the exception of the results from \cite{dc_detector} and \cite{anomaly_trans}. The decision not to utilize the original reported outcomes stems from considerations that include, but are not limited to, the inherent instability of these methods, which necessitates a fair comparison of different algorithms' performances within the same experimental environment. Additionally, the setup of the experimental environment significantly impacts the performance of these two methods. Given the challenges associated with reproducibility, many algorithms lack open-source code and specific data handling protocols. Furthermore, the current community's varying attitudes towards point-adjusted F1 scores \cite{AFFI,TSB-AD,patch_bls} render the sole consideration of PA-F1 insufficient, prompting us to primarily employ classical metrics and interval anomaly indicators \cite{AFFI,VUS} for comparative analysis of different approaches.

Moreover, since \cite{dc_detector} provides a substantial number of reliable performance results for various methods, most of the data in Table \ref{sec:comp_wo_pa} originates from the reports of \cite{dc_detector}. However, in light of the rapid advancements in TSAD, we have additionally replicated and incorporated TimesNet, PatchTST, GPT2-Adapter, and NPSR as comparative methods to maintain the relevance of our comparisons. Consequently, based on the aforementioned reasons, the majority of the reports in Table \ref{tab:comparison_nips_ts} are also sourced from \cite{dc_detector}. Thus, the comparative experiments in this part do not constitute a primary contribution of our paper, but rather serve to provide foundational data for subsequent research.

\subsubsection{Metircs}\label{app:metrics}
Additionally, we employed various evaluation metrics for assessing, encompassing commonly used metrics in time series anomaly detection such as AUC, accuracy (Acc), precision (Pre), recall (Rec), F1-score with point adjustment (PA-F1), and classic F1-score (F1-cls) and AUC \cite{anomaly_trans}. Moreover, we adopted state-of-the-art evaluation techniques like Affiliation F1 (Aff-F1) \cite{AFFI}, along with Volume under the surface (VUS) \cite{VUS}. 
Although the PA-F1 is a widely used metric, it does not effectively evaluate anomalous events. Therefore, we introduce the following two new metrics.
Affiliation F1 is an evaluation metric based on the proximity of predicted events to the true labels.
VUS considers including anomalous events in its evaluation by computing the receiver operator characteristic (ROC) curve. 
Employing various methodologies allows for a more comprehensive evaluation of our model. To compare with other methods \cite{dc_detector,anomaly_trans,OmniAnomaly,LSTM_VAE,VAR} fairly, we utilized the point adjustment technique, identifying an anomalous segment if the model detects any anomaly point within the segment.

It should be particularly noted that, due to the large number of current metrics, different metrics are suitable for different occasions \cite{TSB-AD}. For example, AUC and F1-cls are relatively strict and lack flexibility, yet their more stringent evaluation also ensures the lower limit of evaluation accuracy \cite{TSB-AD,patch_bls,tkde_Flawed}. Of the sequence/event anomaly evaluation metrics, Aff-F1 and VUS offer the most thorough assessment; nevertheless, they are unable to account for the effects of incorrectly labeled data due to annotation errors. On the other hand, PA-F1 has been extensively employed in previous studies \cite{dc_detector,anomaly_trans,timesnet,gpt2} because of its strong resistance to incorrectly labeled data; however, it may also carry the risk of exaggeration \cite{patch_bls,TSB-AD,tkde_Flawed,PATE}.
For example, when faced with datasets where different anomalies have long-lasting abnormal intervals or contain significant label noise (such as manually mislabeled data), this could lead to inflated model scores. Additionally, it lacks clear discriminability for stochastic algorithms.

\subsubsection{Baselines}\label{app:baseline}
We compared our model against 30+ benchmark models, including:
(1) Reconstruction-based: AutoEncoder \cite{AutoEncoder}, LSTM-VAE \cite{LSTM_VAE}, OmniAnomaly \cite{OmniAnomaly}, BeatGAN \cite{BeatGAN}, InterFusion \cite{InterFusion}, USAD \cite{v1_USAD}, COUTA \cite{v1_couta}, M2N2 \cite{M2N2}, AdaMemBLS \cite{adamembls};
(2) Autoregression-based: VAR \cite{VAR}, Autoregression \cite{Autoregression}, LSTM-RNN \cite{LSTM_RNN}, LSTM \cite{LSTM}, CL-MPPCA \cite{CL_MPPCA}, TimesNet \cite{timesnet}, NPSR-pt and NPSR-seq \cite{NPSR};
(3) Density-based: LOF \cite{LOF}, MMPCACD \cite{MMPCACD}, DAGMM \cite{DAGMM};
(4) Clustering-based: DeepSVDD \cite{DeepSVDD}, THOC \cite{THOC}, ITAD \cite{ITAD}; Matrix Profile \cite{MatrixProfile};
(5) The classic methods: OCSVM \cite{OCSVM}, OCSVM-based subsequence clustering (OCSVM*), IForest \cite{IForest}, IForest-based subsequence clustering (IForest*), Gradient boosting regression (GBRT) \cite{GBRT}; 
(6) Change point detection and time series segmentation methods: BOCPD \cite{BOCPD}, U-Time \cite{U_Time}, TS-CP2 \cite{TS_CP2};
(7) Contrastive learning: Anomaly Transformer (AnomTrans) \footnote{\url{https://github.com/thuml/Anomaly-Transformer}} \cite{anomaly_trans}, DCdetector \footnote{\url{https://github.com/DAMO-DI-ML/KDD2023-DCdetector}} \cite{dc_detector};
(8) Large Language Model: GPT2-Adapter\cite{gpt2};
(9) Diffusion model: D3R \cite{d3r};

Given the rapid development of time series anomaly detection algorithms, it is challenging to conduct a comprehensive analysis of all the algorithms. Additionally, due to the research and advancement of evaluation metrics, the community generally acknowledges that different algorithms and metrics have their own suitable applications \cite{TSB-AD,patch_bls,tkde_Flawed}. Consequently, there are numerous challenges in setting up comparative experiments. To evaluate various algorithms as comprehensively as possible, we have conducted two sets of comparative experiments based on the previously established metrics.

In this setting, AnomTrans, DCdetector, and GPT2-Adapter are considered the state-of-the-art methods under the evalution of PA metrics (Acc, Pre, Rec, and PA-F1). On the other hand, TimesNet, D3R, and NPSR should be recognized as the state-of-the-art algorithms under other evaluation metrics (AUC, Aff-F1, VUS). The purpose of this approach is to mitigate the inaccuracies in the evaluation that may arise from differences in datasets, metrics, and algorithms, and further demonstrate the versatility of PatchAD across diverse datasets and evaluation scenarios.
In this setting, AnomTrans, DCdetector, and GPT2-Adapter are considered the state-of-the-art methods when evaluated using the PA metrics (Acc, Pre, Rec, and PA-F1). Conversely, TimesNet, D3R, COUTA and NPSR are regarded as the state-of-the-art algorithms under other evaluation metrics, such as AUC, F1-cls, Aff-F1, and VUS. The purpose of this approach is to address the potential inaccuracies in the evaluation that may arise from disparities in datasets, metrics, and algorithms. Additionally, it aims to further illustrate the versatility of PatchAD across a wide range of datasets and evaluation scenarios.

\subsubsection{Parameter Sensitivity}

\begin{figure*}[t]
\centering
\includegraphics[width=.8\linewidth]{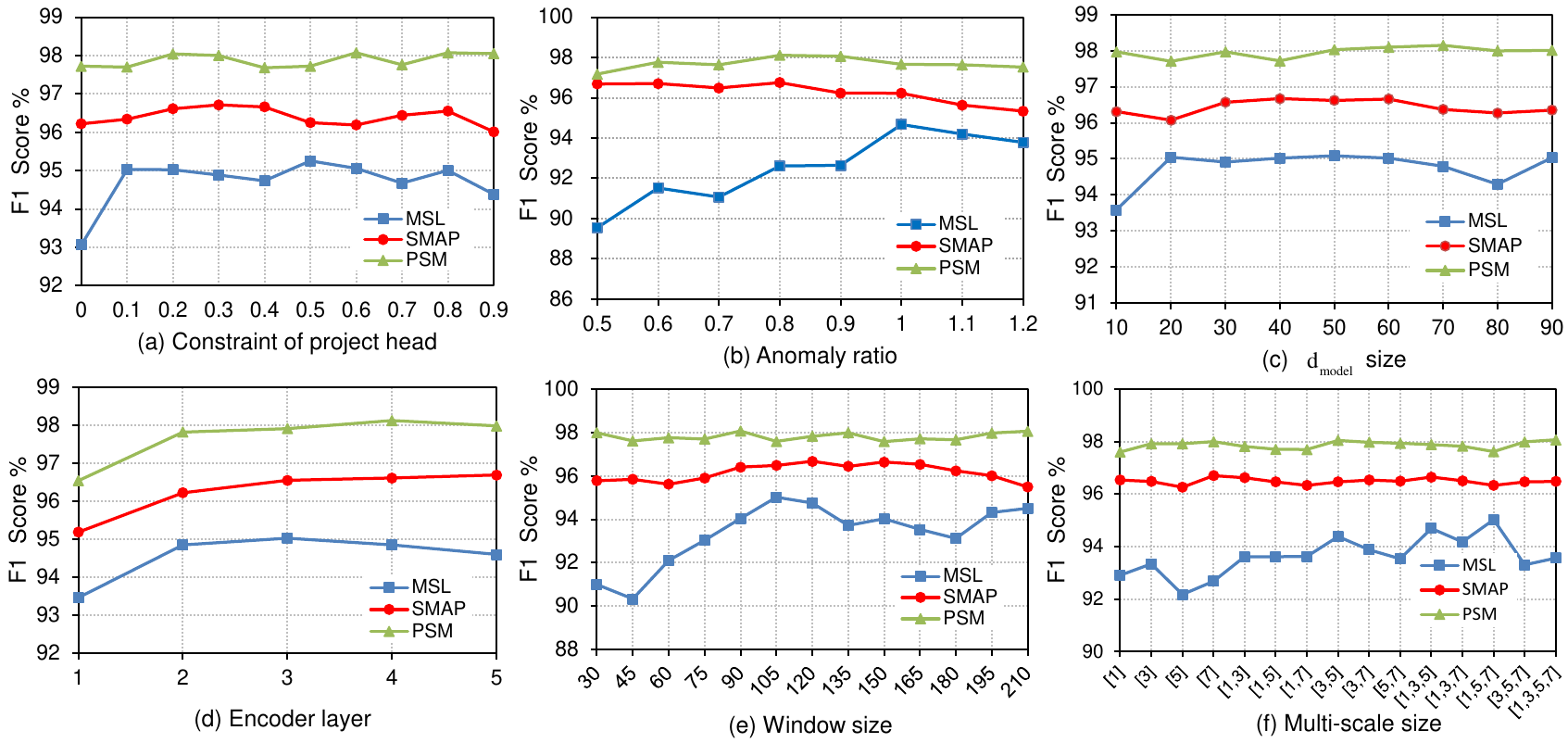}
\caption{Parameter sensitivity studies of main hyper-parameters in PatchAD.}
  \label{fig:para_sen}
  \vspace{-0.3cm}
\end{figure*}

We conducted a sensitivity analysis on PatchAD's parameters. The default parameter settings remain consistent with the basic configuration. Figure \ref{fig:para_sen}(a) demonstrates the impact of the projection head constraint on PatchAD. Overall, it holds more significance for smaller-scale datasets like MSL. However, excessively strong constraints hinder the model's learning process. Therefore, setting this constraint between 0.1 to 0.4 is more advisable.
Figure \ref{fig:para_sen}(b) analyzes PatchAD's final model performance when the anomaly threshold $\sigma$ ranges from 0.5 to 1.2. The selection of the anomaly threshold appears more robust on the PSM and SMAP datasets than the MSL dataset. Setting this value between 0.8 and 1 for most datasets yields optimal results.
It should be noted that the selection of anomaly thresholds \cite{thres1,top_k_threshold,tail-p_threshold} is currently widely studied. Generally, in current research \cite{dc_detector,anomaly_trans,timesnet,patch_bls}, the optimal threshold is searched to calculate the final performance. In practical scenarios, better threshold settings can be selected by combining other prior knowledge with threshold-selection algorithms \cite{thres1,top_k_threshold,tail-p_threshold}.
Figure \ref{fig:para_sen}(c) illustrates how the model's performance is influenced by the network dimension $d_{model}$. PatchAD demonstrates better results with smaller $d_{model}$ sizes. Therefore, to balance model performance and complexity, we set $d_{model}=40$ to ensure PatchAD operates effectively. 
On the MSL dataset, $d_{\text{model}}=10$ performs poorly. This is because MSL combines multiple datasets, increasing its complexity.
Figure \ref{fig:para_sen}(d) indicates that PatchAD's performance is influenced by the number of encoder layers, and given a compromise, we set it to 3 or 4.
When the number of encoder layers increases, performance on SMAP and PSM either improves or stabilizes. However, on the more complex and noisy MSL dataset, performance declines due to a higher risk of overfitting. This phenomenon also occurs when the model dimension increases.
Figure \ref{fig:para_sen}(e) demonstrates PatchAD's robustness concerning the window size, which ranges from 60 to 210. The time window is a critical parameter, typically set to an initial value of 105 across most datasets.
Lastly, Figure \ref{fig:para_sen}(f) displays PatchAD's performance under different multi-scale patch size combinations. (\textit{e.g.}, a patch size of [3,5] represents using two scales in PatchAD.) The outcomes suggest that combining various patch sizes can enhance PatchAD's performance.

\subsection{Implementation Details}
All experiments used PyTorch implementation on a single NVIDIA GTX 1080TI GPU. We summarize the default hyperparameters of PatchAD as follows: PatchAD consists of 3 encoder layers ($L$=3). The hidden layer dimension $d_{{model}}$ is set to 40. The default projector constraint coefficient is $0.2$. The Adam optimizer's default learning rate is $10^{-4}$, with a batch size of 128, trained for 3 epochs across all datasets. Different patch sizes and window sizes were chosen for each dataset, defaulting to [3, 5] and 105, respectively. If an anomaly score of any given timestamp exceeds a threshold $\sigma$, the model identifies it as an anomaly. The default value for $\sigma$ is 1.
Additionally, in terms of the learning rate strategy, we do not use extra learning rate adjustment strategies and only rely on the learning rate adjustment of the Adam optimizer algorithm, with no weight decay set in Adam. To ensure consistent experimental environments, all models undergo standardization processing, namely Z-score standardization, during the preprocessing phase for all data.

\subsection{Comparison Results}

\subsubsection{Additional Analysis}
In Table \ref{sec:comp_wo_pa}, we have also evaluated the newly added methods, M2N2 and AdaMemBLS. The results show that they can achieve the best performance on individual metrics for specific datasets. For example, M2N2 is the best model in terms of the V-PR metric on the MSL dataset, while AdaMemBLS achieves the best performance in the Aff-F1 metric on MSL and the V-PR metric on SWAT. However, despite their strong performance on individual datasets, these methods perform poorly on more complex datasets. For instance, on the SWAT, WADI, and Swan datasets, their performance is relatively weak. Therefore, we believe that these two lightweight models are more suitable for simple datasets or scenarios with relatively straightforward temporal features. They are not well-suited for complex, high-dimensional, and long-length time-series data.
In contrast, PatchAD demonstrates superior performance on complex datasets such as SWAT, WADI, and Swan, and it also performs well on simpler datasets. This indicates that the lightweight design of PatchAD can adapt to datasets of varying complexities without compromising the model's expressive power and performance due to its lightweight structure.

\subsubsection{Comparisons via Metrics with PA}
In Table \ref{tab:comparison_nips_ts} in Appendix, the NIPS-TS-SWAN and NIPS-TS-GECCO datasets represent more challenging datasets, encompassing a wider array of anomaly types. Compared to other baseline methods, PatchAD consistently achieves SOTA results on these two datasets. Particularly noteworthy is the 9.3\% increase in PA-F1 over the DCdetector and a 24.9\% increase over the Anomaly Transformer on the NIPS-TS-GECCO dataset. 
Table 
\ref{tab:comparison_ucr} in Appendix demonstrates that our approach outperforms previous methods on univariate dataset UCR. This shows that our method can also fetch state-of-the-art performance on univariate datasets.

\begin{table}[!t]
  \centering
  \setlength\tabcolsep{3pt}
  \caption{Overall results on NIPS-TS datasets. All results are in \%, the best in \textbf{Bold}, and the second in \uline{underlined}.}
      \resizebox{0.8\linewidth}{!}{
    \begin{tabular}{c|ccc|ccc|c}
    \toprule
    \textbf{Dataset} & \multicolumn{3}{c|}{\textbf{NIPS-TS-SWAN}} & \multicolumn{3}{c|}{\textbf{NIPS-TS-GECCO}} & \multicolumn{1}{c}{\multirow{2}[4]{*}{\textbf{ \makecell*[c]{Average\\ F1} }}} \\
\cmidrule{1-7}\textbf{Metric} & \textbf{P} & \textbf{R} & \textbf{PA-F1} & \textbf{P} & \textbf{R} & \textbf{PA-F1} &  \\
    \midrule
    MatrixProfile \scriptsize{\cite{MatrixProfile}} & 16.70  & 17.50  & 17.10  & 4.60  & 18.50  & 7.40  & 12.25  \\
    GBRT \scriptsize{\cite{GBRT}} & 44.70  & 37.50  & 40.80  & 17.50  & 14.00  & 15.60  & 28.20  \\
    LSTM-RNN \scriptsize{\cite{LSTM_RNN}} & 52.70  & 22.10  & 31.20  & 34.30  & 27.50  & 30.50  & 30.85  \\
    Autoregression \scriptsize{\cite{Autoregression}} & 42.10  & 35.40  & 38.50  & 39.20  & 31.40  & 34.90  & 36.70  \\
    OCSVM \scriptsize{\cite{OCSVM}} & 47.40  & 49.80  & 48.50  & 18.50  & 74.30  & 29.60  & 39.05  \\
    IForest* \scriptsize{\cite{IForest}} & 40.60  & 42.50  & 41.60  & 39.20  & 31.50  & 39.00  & 40.30  \\
    AutoEncoder \scriptsize{\cite{AutoEncoder}} & 49.70  & 52.20  & 50.90  & 42.40  & 34.00  & 37.70  & 44.30  \\
    AnomalyTrans \scriptsize{\cite{anomaly_trans}} & 90.70  & 47.40  & 62.30  & 25.70  & 28.50  & 27.00  & 44.65  \\
    IForest \scriptsize{\cite{IForest}} & 56.90  & 59.80  & 58.30  & 43.90  & 35.30  & 39.10  & 48.70  \\
    AnomalyTrans \scriptsize{\cite{anomaly_trans}} & 90.71  & 47.43  & 62.29  & 29.96  & 48.63  & 37.08  & 49.69  \\
    DCdetector \scriptsize{\cite{dc_detector}} & \uline{96.57 } & \uline{59.08 } & \uline{73.31 } & \uline{38.41 } & \uline{59.73 } & \uline{46.76 } & \uline{60.04 } \\
    \midrule
    \textbf{PatchAD (Ours)} & \textbf{96.75 } & \textbf{59.15 } & \textbf{73.41 } & \textbf{40.62 } & \textbf{62.88 } & \textbf{49.35 } & \textbf{61.38 } \\
    \bottomrule
    \end{tabular}
        }
  \label{tab:comparison_nips_ts}%
\end{table}%

\begin{table}[]
\centering
\caption{Results on univariable UCR dataset.}
\resizebox{0.7\linewidth}{!}{
\begin{tabular}{c|ccccc}
\toprule
\textbf{Metric} & \textbf{Acc} & \textbf{P} & \textbf{R} & \textbf{F1} \\
\midrule
AnomalyTrans & 99.49 & 60.41 & \textbf{100} & 73.08 \\
DCdetector & \uline{99.51} & \uline{61.62} & \textbf{100} & \uline{74.05}  \\ \midrule
\textbf{PatchAD (Ours)} & \textbf{99.52 } & \textbf{72.72 } & \textbf{100} & \textbf{84.00 } \\
\bottomrule
\end{tabular}%
}
\label{tab:comparison_ucr}
\end{table}

\subsection{Visualization comparison}\label{app:vis}
Figure \ref{fig:anom_types2} shows the performance of WindowAD-based algorithms (TCN-ED and TimesNet) in the face of different types of anomalies. 
The results for the first and second cases are derived from the TCN-ED algorithm, while the result for the final case is obtained from the TimesNet algorithm.
For such straightforward synthetic time series data, in fact, most algorithms can handle them effectively \cite{TSB-AD,catch}. Therefore, we gathered instances of their suboptimal performance for case analysis.
These methods have difficulty in detecting contextual and group point anomalies and are able to detect seasonal anomalies, but there are some false positives, which indicates that it has a low precision.
\begin{figure}[!h]
  \centering
  \resizebox{\linewidth}{!}{
  \begin{minipage}[b]{0.33\textwidth}
    \includegraphics[width=1\linewidth]{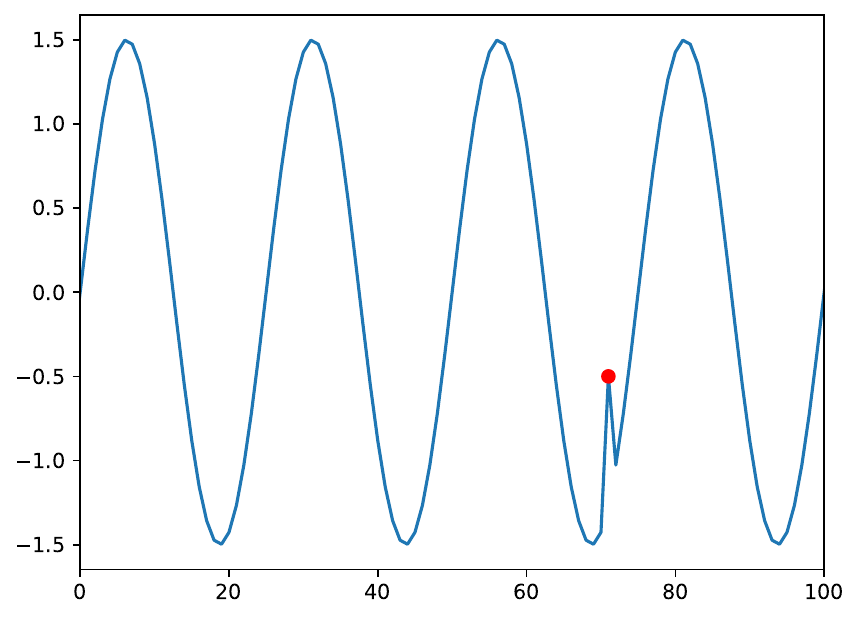}
  \end{minipage}%
  \begin{minipage}[b]{0.33\textwidth}
    \includegraphics[width=1\linewidth]{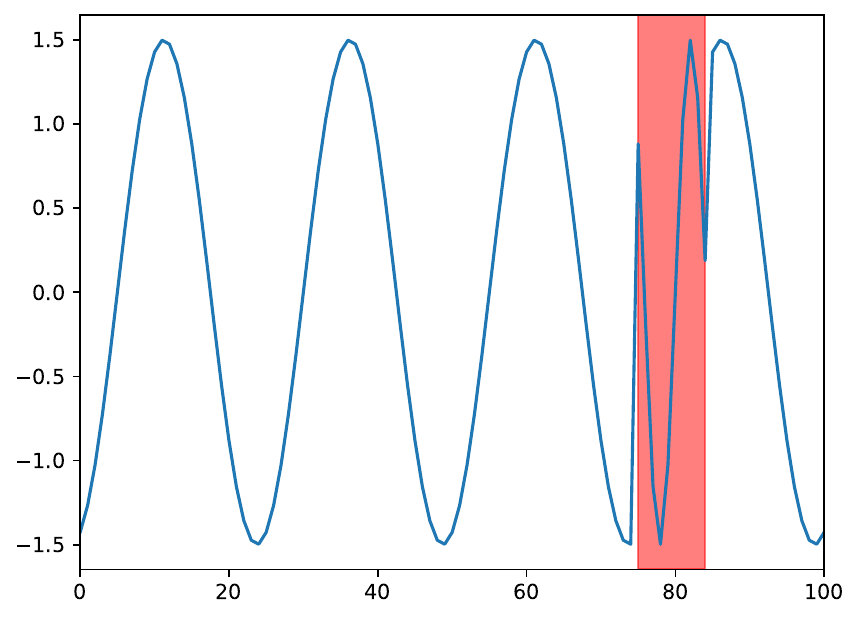}
  \end{minipage}%
  \begin{minipage}[b]{0.33\textwidth}
    \includegraphics[width=1\linewidth]{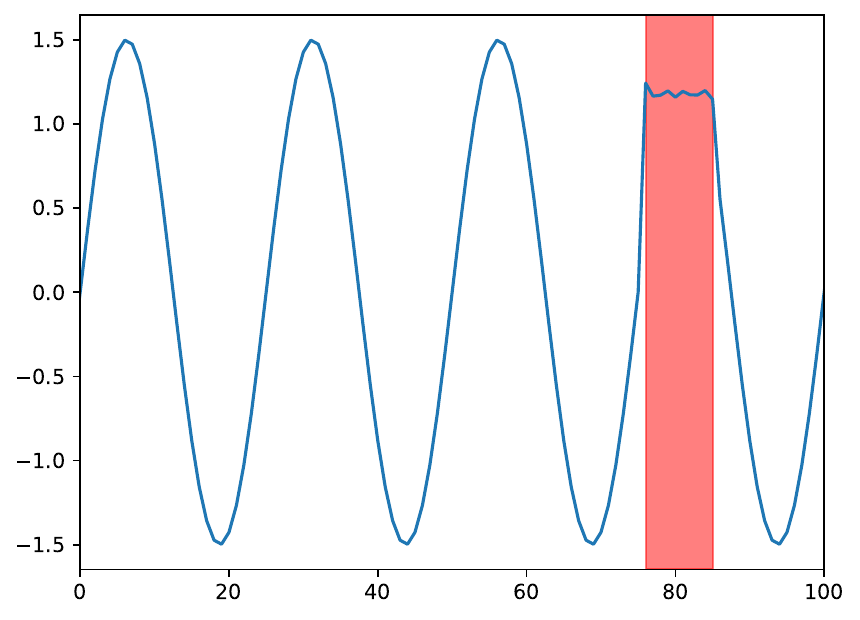}
  \end{minipage}%
  }

  \resizebox{\linewidth}{!}{
  \begin{minipage}[b]{0.33\textwidth}
    \includegraphics[width=1\linewidth]{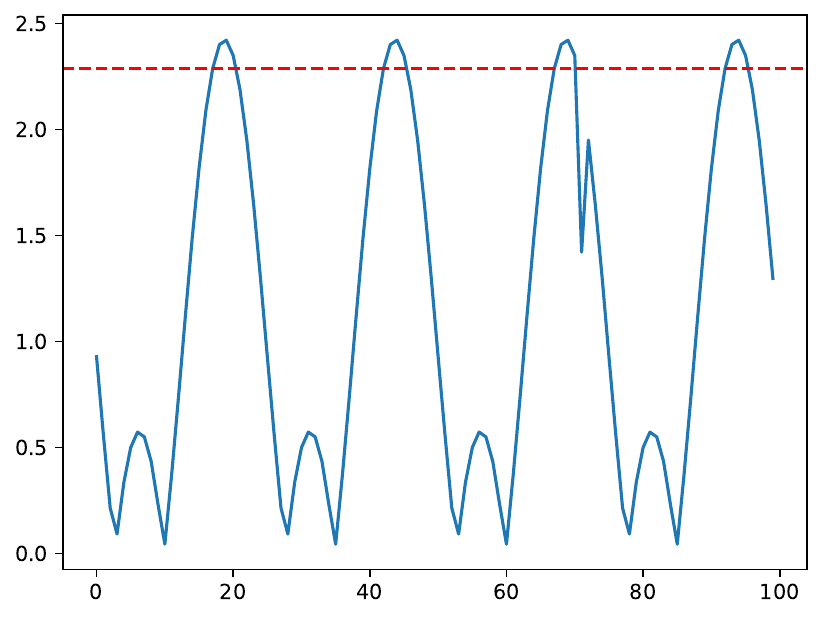}
    \caption{\small{Contextual}}
  \end{minipage}%
  \begin{minipage}[b]{0.33\textwidth}
    \includegraphics[width=1\linewidth]{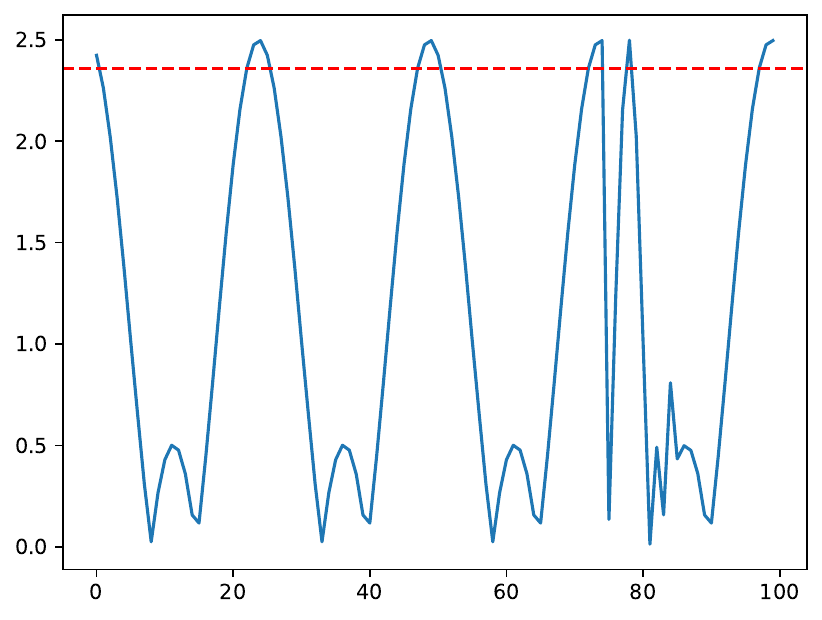}
    \caption{Seasonal}
  \end{minipage}%
  \begin{minipage}[b]{0.33\textwidth}
    \includegraphics[width=1\linewidth]{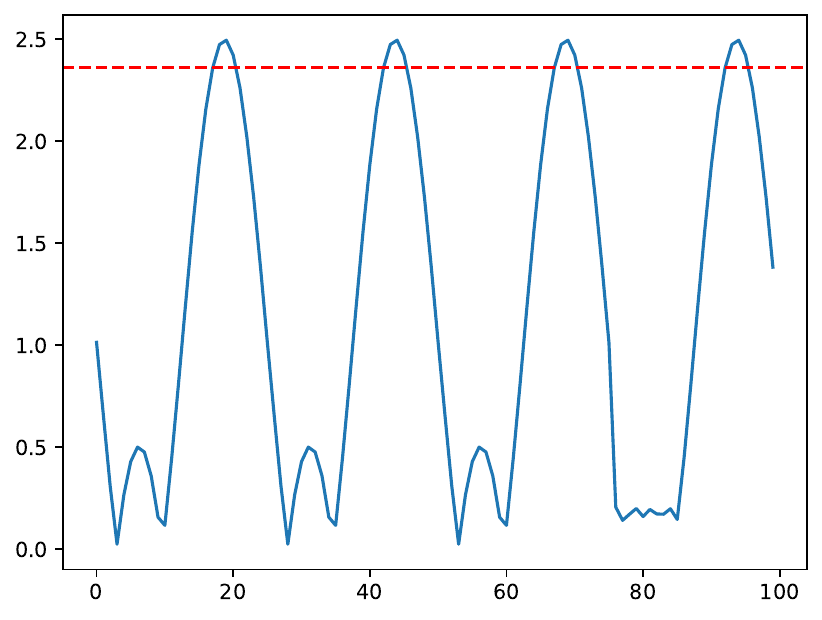}
    \caption{Group}
  \end{minipage}%
  }
  \caption{Visualization of WindowAD-based algorithm's score.}
  \label{fig:anom_types2}
\end{figure}

\subsection{Scalability Analysis}\label{app:scalability}
\subsubsection{Time Cost}
We conducted a time consumption analysis on the SWAT dataset under identical experimental conditions. 
Figure \ref{fig:patch_ad_cost} demonstrates that under different sliding step conditions, our method outperforms DCdetector. Specifically, when the step is set to 1, PatchAD's speed is 1.45x faster than DCdetector.
Additionally, Figure \ref{fig:constant_win} illustrates the time required for model training with a fixed window size (window size=105). This indicates that PatchAD exhibits higher efficiency in training compared to DCdetector.
Figure \ref{fig:constant_step} shows the time required for model training with a fixed sliding step (step=35) while varying the window size from 35 to 175 for different $d_{model}$ sizes.
Both graphs show that the time consumption of PatchAD exhibits a linear growth with respect to the independent variables. Thus, increasing the window size or enlarging $d_{model}$ has a relatively minor impact on PatchAD. However, the time step noticeably affects the duration as this parameter increases data volume.

\begin{figure}
    \includegraphics[width=0.8\linewidth]{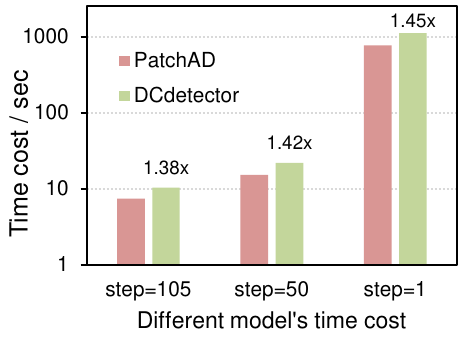}
    \caption{Time cost comparison of PatchAD and DCdetector.}
    \label{fig:patch_ad_cost}
\end{figure}%

\begin{figure*}
    \centering
  \begin{subfigure}[b]{0.4\textwidth}
    \includegraphics[width=\linewidth]{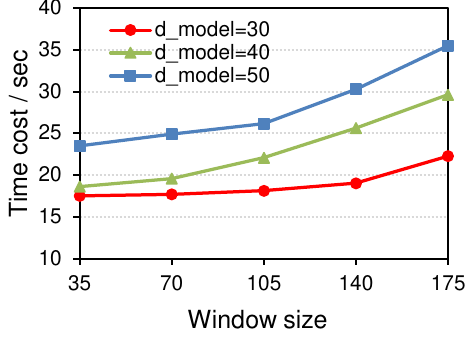}
    \caption{Different $d_{model}$ }
    \label{fig:constant_step}
  \end{subfigure}%
  \begin{subfigure}[b]{0.4\textwidth}
    \includegraphics[width=\linewidth]{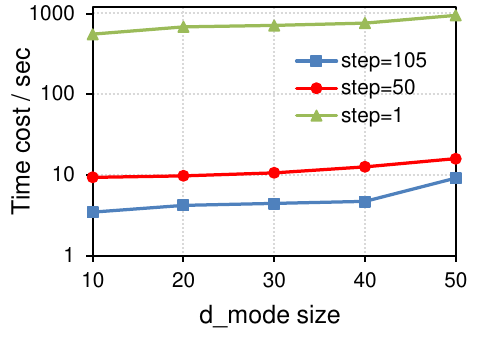}
    \caption{Different window}
    \label{fig:constant_win}
  \end{subfigure}%
  \caption{The scalability analysis of PatchAD.}
  \label{fig:time_cost}
  \vspace{-0.3cm}
\end{figure*}

\subsubsection{Model Params, FLOPs, and letency}
Figure \ref{sfig:model_size} compares the parameters, FLOPs, and latency of different models on SWAT dataset. Our model has only \textbf{0.403M} parameters, which is fewer than those of AnomTrans (AT) and DCdetector, and significantly fewer than the LLM model (GPT2-Ada.) and the diffusion model (D3R). 

These observations indicate that our model is more lightweight and resource-efficient.

\subsubsection{Scaling of Model Parameters}
Next, we theoretically analyze the scaling of model parameters, with further scaling analysis provided in Appendix \ref{app:scalability}.

The parameters primarily arise from the MLP Mixers. Each MLP comprises two fully connected (FC) layers. The parameter counts per layer are as follows:  
\begin{enumerate}
    \item Channel Mixer: \( O(C^2) \).  
    \item Inter Mixer: \( O(N^2) \).  
    \item Intra Mixer: \( O(P^2) \).  
    \item MixRep Mixer: \( O(D^2) \).  
\end{enumerate}
Thus, the total parameter count for PatchAD can be expressed as:  
\(
O\left(C^2 + N^2 + P^2 + D^2\right)=\left(\min(C,N,P,D)\right)^2.
\)
It is important to note that when comparing models with different layer configurations, the layer count can be treated as a constant, allowing for the omission of the layer coefficient.

For models that rely entirely on MLP, such as USAD, the parameter count can be expressed as:
\(
    O_{mlp}= O((CT)^2).
\)
Here, \( T = N \cdot P \) represents the time series window length. Since USAD compresses the time window and channels entirely, its MLP parameter count is predominantly determined by the temporal window length and channel dimensions.

In the case of TCN networks, the parameter count primarily derives from the convolutional neural network. If we consider each channel of the temporal sequence being processed independently, the parameter count can be expressed as:
\(
O_{tcn} = O(C \cdot D^2 \cdot K).
\)
If a channel-mixing approach is adopted, this can be reduced to \( O_{tcn} = O(D^2 \cdot K) \), where \( K \) denotes the size of the convolution kernel.

Similarly, for common attention models, the parameter count depends on both the attention network and the MLP component. When considering channels independently, the total can be expressed as:
\( O_{attn} = O(C \cdot D^2) \). If channel mixing is taken into account, the parameter count can be reduced to \( O(D^2) \). It is worth noting that, in practice, the dimension \( D \) of attention models is typically set to exceed the temporal window length to achieve better performance. In contrast, PatchAD distributes its parameter count across different Mixers, resulting in a smaller dimension \( D \) compared to attention models.

Considering various scenarios, it is evident that PatchAD exhibits a relative advantage in the scaling of parameter counts compared to other models. However, it is essential to emphasize that the actual module designs of different models are quite complex, which may ultimately lead to discrepancies in performance.

\subsubsection{Complexity Analysis}
This part begins with an analysis of the time complexity of PatchAD, followed by a comparison with other model.

Time Complexity of PatchAD:
Each layer of PatchAD involves four MLP Mixers operating on different tensor dimensions. Let \( C \) denote the number of channels, \( N \) the number of patches, \( P \) the patch size, \( D \) the hidden dimension, and \( L \) the number of layers. 

The time complexity of PatchAD primarily arises from the various Mixers, detailed as follows:
\begin{enumerate}
    \item Channel Mixer: Processes the channel dimension \( C \), yielding \( O(C^2 D(N + P)) \).  
    \item Inter Mixer: Operates on the inter-patch dimension \( N \), leading to \( O(CDN^2) \). 
    \item Intra Mixer: Functions on the intra-patch dimension \( P \), resulting in \( O(CDP^2) \).  
    \item MixRep Mixer: Unifies representations along the hidden dimension \( D \), contributing \( O(CD^2(N + P)) \).  
\end{enumerate}
Thus, the overall time complexity for each layer can be expressed as:
\begin{equation}
    O_{T}=O\left(C^2 D(N + P) + CDN^2 + CDP^2 + CD^2(N + P)\right).
\end{equation}
For \( L \) layers, this results in \( L \times \) the aforementioned terms, \textit{i.e.}, \( O_{T}^{(L)}=L \cdot O_{T} \). Given that PatchAD employs a multi-layer parallel computation approach, the calculations for different layers can theoretically proceed simultaneously, allowing us to disregard the \( L \) factor. Consequently, the final complexity remains \( O_T \).

Next, we analyze the time complexity of contemporary attention models. Similarly, we primarily consider their core complexity sources, namely the attention module and the MLP component. 

The time complexity contributed by the attention module is given by:
\[
O_{attn}=O(LC(N+P)^2D),
\]
while the MLP module contributes:
\[
O_{mlp}=O(LC(N+P)D^2).
\]
Thus, the overall complexity can be expressed as:
\begin{equation}
    \begin{aligned}
        O_1&=O_{attn}+O_{mlp}\\
        &=O(LC(N+P)^2D)+O(LC(N+P)D^2).
    \end{aligned}
\end{equation}

When considering the parallel optimization employed by PatchAD, it is evident that PatchAD possesses a distinct advantage in terms of time complexity. We will now analyze the comparison of time complexities between the two models when parallel optimization strategies are not utilized.

Assuming both models have the same number of layers, the coefficient \( L \) can be disregarded. When \( T = N \cdot P \gg D \), meaning the temporal window length significantly exceeds the model dimension, the attention model is primarily influenced by the square of the time length, resulting in a complexity of \( O(LC(N+P)^2D) \). In contrast, PatchAD segments the temporal sequence into varying sizes, leading to a primary complexity of \( O\left(LC^2 D(N + P) + LCDN^2 \right) \). Thus, when the temporal window length is excessively long, the advantages of PatchAD become particularly pronounced. 

Conversely, if \( T \sim D \), the complexity of the attention model can be approximated as \( O(LCT^3) \); similarly, the complexity of PatchAD can also be approximated as \( O(LCT^3) \). Therefore, when disregarding the impact of constant factors, the time complexities of both models are essentially comparable.

In summary, when parallel acceleration optimizations are commonly employed, PatchAD's time complexity clearly outperforms that of other models. Additionally, when confronted with lengthy time series, PatchAD's advantages become even more pronounced. In other scenarios, such as when the temporal window is relatively short, the time complexities of both models are approximately similar.

\subsection{Results of Parameter Sensitivity}
All tables (\ref{tab:addlabel1},\ref{tab:addlabel2},\ref{tab:addlabel3},\ref{tab:addlabel4},\ref{tab:addlabel5},\ref{tab:addlabel6}) in this subsection present the raw data from the parameter sensitivity experiments.

\begin{table}[htbp]
  \centering
  \caption{Parameter studies on project head constraint results (patch size=[3,5], window size=105). All results are in \%.}
    \resizebox{1\linewidth}{!}{
    \begin{tabular}{c|cccc|cccc|cccc}
    \toprule
    \textbf{Dataset} & \multicolumn{4}{c|}{\textbf{MSL}} & \multicolumn{4}{c|}{\textbf{SMAP}} & \multicolumn{4}{c}{\textbf{PSM}} \\
    \midrule
    \textbf{Contraint} & \textbf{Acc} & \textbf{P} & \textbf{R} & \textbf{F1} & \textbf{Acc} & \textbf{P} & \textbf{R} & \textbf{F1} & \textbf{Acc} & \textbf{P} & \textbf{R} & \textbf{F1} \\
    \midrule
    \textbf{0.0} & 98.51  & 91.67  & 94.49  & 93.06  & 99.01  &	93.55  & 99.05  & 96.22  & 98.73  & 97.42  & 98.03  & 97.72  \\
    \textbf{0.1} & 98.92  & 91.96  & 98.33  & 95.03  & 99.04  & 94.43  & 98.33  & 96.34  & 98.72  & 97.35  & 98.04  & 97.70  \\
    \textbf{0.2} & 98.92  & 92.05  & 98.20  & 95.02  & 99.11  & 94.42  & 98.91  & 96.61  & 98.91  & 97.43  & 98.66  & 98.04  \\
    \textbf{0.3} & 98.89  & 92.03  & 97.93  & 94.88  & 99.14  & 94.41  & 99.11  & 96.71  & 98.89  & 97.42  & 98.59  & 98.00  \\
    \textbf{0.4} & 98.86  & 92.09  & 97.54  & 94.73  & 99.12  & 94.45  & 98.97  & 96.66  & 98.71  & 97.42  & 97.94  & 97.68  \\
    \textbf{0.5} & 98.96  & 92.12  & 98.60  & 95.25  & 99.02  & 94.40  & 98.17  & 96.25  & 98.73  & 97.42  & 98.03  & 97.72  \\
    \textbf{0.6} & 98.92  & 92.11  & 98.20  & 95.06  & 99.01  & 94.39  & 98.07  & 96.19  & 98.92  & 97.45  & 98.70  & 98.07  \\
    \textbf{0.7} & 98.84  & 91.91  & 97.59  & 94.67  & 99.07  & 94.43  & 98.55  & 96.44  & 98.76  & 97.39  & 98.14  & 97.76  \\
    \textbf{0.8} & 98.91  & 92.01  & 98.20  & 95.00  & 99.10  & 94.40  & 98.81  & 96.55  & 98.92  & 97.42  & 98.73  & 98.07  \\
    \textbf{0.9} & 98.78  & 91.99  & 96.90  & 94.38  & 98.96  & 94.37  & 97.70  & 96.01  & 98.92  & 97.47  & 98.64  & 98.05  \\
    \bottomrule
    \end{tabular}%
    }
  \label{tab:addlabel1}%
\end{table}%

\begin{table}[htbp]
  \centering
  \caption{Parameter studies on anomaly threshold $\sigma$ results (patch size=[3,5], window size=105). All results are in \%.}
    \resizebox{1\linewidth}{!}{
    \begin{tabular}{c|cccc|cccc|cccc}
\toprule
\textbf{Dataset} & \multicolumn{4}{c|}{\textbf{MSL}} & \multicolumn{4}{c|}{\textbf{SMAP}} & \multicolumn{4}{c}{\textbf{PSM}} \\
\midrule
\textbf{AR} & \textbf{Acc} & \textbf{P} & \textbf{R} & \textbf{F1} & \textbf{Acc} & \textbf{P} & \textbf{R} & \textbf{F1} & \textbf{Acc} & \textbf{P} & \textbf{R} & \textbf{F1} \\
\midrule
\textbf{0.5} & 97.93  & 95.39  & 84.39  & 89.55  & 99.15  & 96.46  & 96.91  & 96.68  & 98.45  & 98.61  & 95.77  & 97.17  \\
\textbf{0.6} & 98.27  & 94.70  & 88.54  & 91.52  & 99.15  & 95.84  & 97.58  & 96.70  & 98.77  & 98.33  & 97.20  & 97.76  \\
\textbf{0.7} & 98.17  & 93.76  & 88.54  & 91.07  & 99.09  & 95.22  & 97.77  & 96.48  & 98.70  & 98.07  & 97.20  & 97.64  \\
\textbf{0.8} & 98.45  & 93.19  & 92.04  & 92.61  & 99.15  & 94.73  & 98.86  & 96.75  & 98.95  & 97.94  & 98.29  & 98.11  \\
\textbf{0.9} & 98.45  & 92.36  & 92.93  & 92.64  & 99.01  & 94.07  & 98.49  & 96.23  & 98.92  & 97.68  & 98.45  & 98.06  \\
\textbf{1.0} & 98.84  & 91.80  & 97.73  & 94.67  & 99.01  & 94.05  & 98.49  & 96.22  & 98.70  & 97.30  & 98.03  & 97.66  \\
\textbf{1.1} & 98.73  & 91.22  & 97.36  & 94.19  & 98.85  & 92.76  & 98.69  & 95.63  & 98.68  & 97.06  & 98.20  & 97.63  \\
\textbf{1.2} & 98.64  & 90.45  & 97.36  & 93.78  & 98.76  & 92.19  & 98.69  & 95.33  & 98.62  & 96.85  & 98.20  & 97.52  \\
\bottomrule
\end{tabular}%
}

  \label{tab:addlabel2}%
\end{table}%

\begin{table}[htbp]
  \centering
  \caption{Parameter studies on $d_{model}$ size results (patch size=[3,5], window size=105). All results are in \%.}
    \resizebox{1\linewidth}{!}{
    \begin{tabular}{c|cccc|cccc|cccc}
    \toprule
    \textbf{Dataset} & \multicolumn{4}{c|}{\textbf{MSL}} & \multicolumn{4}{c|}{\textbf{SMAP}} & \multicolumn{4}{c}{\textbf{PSM}} \\
    \midrule
    {$d_{model}$} & \textbf{Acc} & \textbf{P} & \textbf{R} & \textbf{F1} & \textbf{Acc} & \textbf{P} & \textbf{R} & \textbf{F1} & \textbf{Acc} & \textbf{P} & \textbf{R} & \textbf{F1} \\
    \midrule
    \textbf{10} & 98.62  & 91.81  & 95.42  & 93.58  & 99.04  & 94.42  & 98.28  & 96.31  & 98.87  & 97.38  & 98.56  & 97.97  \\
    \textbf{20} & 98.92  & 92.08  & 98.20  & 95.04  & 98.98  & 94.33  & 97.88  & 96.07  & 98.73  & 97.37  & 98.05  & 97.71  \\
    \textbf{30} & 98.89  & 92.06  & 97.94  & 94.91  & 99.10  & 94.43  & 98.80  & 96.57  & 98.87  & 97.43  & 98.52  & 97.97  \\
    \textbf{40} & 98.91  & 92.03  & 98.20  & 95.02  & 99.13  & 94.49  & 98.95  & 96.67  & 98.73  & 97.42  & 98.03  & 97.72  \\
    \textbf{50} & 98.93  & 92.15  & 98.20  & 95.08  & 99.11  & 94.45  & 98.89  & 96.62  & 98.90  & 97.44  & 98.63  & 98.03  \\
    \textbf{60} & 98.91  & 92.03  & 98.20  & 95.02  & 99.13  & 94.44  & 98.98  & 96.66  & 98.94  & 97.48  & 98.74  & 98.10  \\
    \textbf{70} & 98.87  & 92.08  & 97.67  & 94.79  & 99.05  & 94.40  & 98.43  & 96.37  & 98.97  & 97.48  & 98.83  & 98.15  \\
    \textbf{80} & 98.77  & 91.99  & 96.70  & 94.29  & 99.03  & 94.38  & 98.24  & 96.27  & 98.89  & 97.44  & 98.58  & 98.00  \\
    \textbf{90} & 98.92  & 92.06  & 98.20  & 95.03  & 99.05  & 94.38  & 98.42  & 96.35  & 98.89  & 97.46  & 98.56  & 98.01  \\
    \bottomrule
    \end{tabular}%
    }
  \label{tab:addlabel3}%
\end{table}%

\begin{table}[htbp]
  \centering
  \caption{Parameter studies on encoder layers $L$ results (patch size=[3,5], window size=105). All results are in \%.}
    \resizebox{1\linewidth}{!}{
    \begin{tabular}{c|cccc|cccc|cccc}
\toprule
\textbf{Dataset} & \multicolumn{4}{c|}{\textbf{MSL}} & \multicolumn{4}{c|}{\textbf{SMAP}} & \multicolumn{4}{c}{\textbf{PSM}} \\
\midrule
\textbf{Encoder layer} & \textbf{Acc} & \textbf{P} & \textbf{R} & \textbf{F1} & \textbf{Acc} & \textbf{P} & \textbf{R} & \textbf{F1} & \textbf{Acc} & \textbf{P} & \textbf{R} & \textbf{F1} \\
\midrule
\textbf{1} & 98.60  & 91.72  & 95.27  & 93.46  & 98.76  & 94.29  & 96.11  & 95.19  & 98.10  & 97.28  & 95.82  & 96.54  \\
\textbf{2} & 98.88  & 91.96  & 97.93  & 94.85  & 99.01  & 94.42  & 98.09  & 96.22  & 98.79  & 97.44  & 98.20  & 97.82  \\
\textbf{3} & 98.92  & 92.07  & 98.20  & 95.03  & 99.10  & 94.41  & 98.79  & 96.55  & 98.84  & 97.47  & 98.36  & 97.91  \\
\textbf{4} & 98.88  & 92.19  & 97.67  & 94.85  & 99.11  & 94.42  & 98.91  & 96.61  & 98.95  & 97.41  & 98.83  & 98.12  \\
\textbf{5} & 98.83  & 91.95  & 97.41  & 94.60  & 99.13  & 94.46  & 99.02  & 96.69  & 98.87  & 97.44  & 98.52  & 97.98  \\
\bottomrule
\end{tabular}%
}
  \label{tab:addlabel4}%
\end{table}%

\begin{table}[htbp]
  \centering
  \caption{Parameter studies on window size results (window size=105). All results are in \%.}
    \resizebox{1\linewidth}{!}{
    \begin{tabular}{c|cccc|cccc|cccc}
    \toprule
    \textbf{Dataset} & \multicolumn{4}{c|}{\textbf{MSL}} & \multicolumn{4}{c|}{\textbf{SMAP}} & \multicolumn{4}{c}{\textbf{PSM}} \\
    \midrule
    \textbf{Window size} & \textbf{Acc} & \textbf{P} & \textbf{R} & \textbf{F1} & \textbf{Acc} & \textbf{P} & \textbf{R} & \textbf{F1} & \textbf{Acc} & \textbf{P} & \textbf{R} & \textbf{F1} \\
    \midrule
    \textbf{30} & 98.11  & 91.47  & 90.52  & 90.99  & 98.91  & 94.38  & 97.23  & 95.78  & 98.88  & 97.48  & 98.51  & 97.99  \\
    \textbf{45} & 97.98  & 91.23  & 89.44  & 90.33  & 98.93  & 95.55  & 96.14  & 95.84  & 98.67  & 97.30  & 97.92  & 97.61  \\
    \textbf{60} & 98.33  & 92.05  & 92.16  & 92.10  & 98.86  & 94.18  & 97.09  & 95.62  & 98.75  & 97.59  & 97.93  & 97.76  \\
    \textbf{75} & 98.51  & 91.79  & 94.31  & 93.03  & 98.93  & 94.27  & 97.60  & 95.90  & 98.72  & 97.35  & 98.06  & 97.70  \\
    \textbf{90} & 98.71  & 91.92  & 96.24  & 94.03  & 99.06  & 94.34  & 98.55  & 96.40  & 98.92  & 97.44  & 98.71  & 98.07  \\
    \textbf{105} & 98.92  & 92.05  & 98.20  & 95.02  & 99.08  & 94.44  & 98.60  & 96.48  & 98.66  & 97.38  & 97.79  & 97.59  \\
    \textbf{120} & 98.86  & 92.10  & 97.55  & 94.75  & 99.13  & 94.47  & 98.98  & 96.67  & 98.79  & 97.53  & 98.13  & 97.83  \\
    \textbf{135} & 98.65  & 91.85  & 95.67  & 93.72  & 99.07  & 94.46  & 98.50  & 96.44  & 98.88  & 97.38  & 98.60  & 97.99  \\
    \textbf{150} & 98.71  & 91.83  & 96.30  & 94.02  & 99.12  & 94.48  & 98.91  & 96.64  & 98.66  & 97.43  & 97.72  & 97.58  \\
    \textbf{165} & 98.61  & 91.85  & 95.27  & 93.53  & 99.09  & 94.50  & 98.65  & 96.53  & 98.73  & 97.44  & 97.98  & 97.71  \\
    \textbf{180} & 98.53  & 91.73  & 94.55  & 93.12  & 99.01  & 94.35  & 98.18  & 96.23  & 98.70  & 97.55  & 97.78  & 97.66  \\
    \textbf{195} & 98.77  & 92.02  & 96.70  & 94.31  & 98.96  & 94.35  & 97.74  & 96.01  & 98.87  & 97.44  & 98.50  & 97.97  \\
    \textbf{210} & 98.81  & 91.97  & 97.17  & 94.50  & 98.83  & 94.30  & 96.71  & 95.49  & 98.92  & 97.45  & 98.69  & 98.06  \\
    \bottomrule
    \end{tabular}%
    }
  \label{tab:addlabel5}%
\end{table}%

\begin{table}[htbp]
  \centering
  \caption{Parameter studies on multi-scale patch size results (window size=105). All results are in \%.}
    \resizebox{1\linewidth}{!}{
    \begin{tabular}{c|cccc|cccc|cccc}
    \toprule
    \textbf{Dataset} & \multicolumn{4}{c|}{\textbf{MSL}} & \multicolumn{4}{c|}{\textbf{SMAP}} & \multicolumn{4}{c}{\textbf{PSM}} \\
    \midrule
    \textbf{Patch size} & \textbf{Acc} & \textbf{P} & \textbf{R} & \textbf{F1} & \textbf{Acc} & \textbf{P} & \textbf{R} & \textbf{F1} & \textbf{Acc} & \textbf{P} & \textbf{R} & \textbf{F1} \\
    \midrule
    \textbf{[1]} & 98.49  & 91.73  & 94.12  & 92.91  & 99.09  & 94.47  & 98.69  & 96.53  & 98.66  & 97.21  & 97.99  & 97.60  \\
    \textbf{[3]} & 98.57  & 91.67  & 95.08  & 93.34  & 99.08  & 94.44  & 98.61  & 96.48  & 98.84  & 97.56  & 98.26  & 97.91  \\
    \textbf{[5]} & 98.34  & 91.70  & 92.66  & 92.18  & 99.02  & 94.40  & 98.20  & 96.26  & 98.83  & 97.44  & 98.37  & 97.91  \\
    \textbf{[7]} & 98.45  & 91.96  & 93.41  & 92.68  & 99.13  & 94.46  & 99.04  & 96.70  & 98.88  & 97.43  & 98.56  & 97.99  \\
    \textbf{[1, 3]} & 98.63  & 91.87  & 95.42  & 93.61  & 99.11  & 94.48  & 98.86  & 96.62  & 98.78  & 97.46  & 98.16  & 97.81  \\
    \textbf{[1, 5]} & 98.63  & 91.88  & 95.42  & 93.61  & 99.07  & 94.43  & 98.58  & 96.46  & 98.72  & 97.37  & 98.04  & 97.70  \\
    \textbf{[1, 7]} & 98.63  & 91.89  & 95.42  & 93.62  & 99.04  & 94.43  & 98.32  & 96.33  & 98.71  & 97.39  & 97.99  & 97.69  \\
    \textbf{[3, 5]} & 98.78  & 91.99  & 96.90  & 94.38  & 99.07  & 94.39  & 98.63  & 96.46  & 98.91  & 97.43  & 98.66  & 98.04  \\
    \textbf{[3, 7]} & 98.68  & 91.92  & 95.93  & 93.88  & 99.09  & 94.40  & 98.76  & 96.53  & 98.87  & 97.41  & 98.55  & 97.97  \\
    \textbf{[5, 7]} & 98.61  & 91.86  & 95.27  & 93.53  & 99.08  & 94.44  & 98.63  & 96.49  & 98.85  & 97.40  & 98.46  & 97.93  \\
    \textbf{[1, 3, 5]} & 98.85  & 92.01  & 97.54  & 94.69  & 99.12  & 94.47  & 98.92  & 96.64  & 98.82  & 97.37  & 98.39  & 97.88  \\
    \textbf{[1, 3, 7]} & 98.74  & 92.00  & 96.46  & 94.18  & 99.09  & 94.40  & 98.71  & 96.50  & 98.79  & 97.43  & 98.21  & 97.82  \\
    \textbf{[1, 5, 7]} & 98.92  & 92.05  & 98.20  & 95.02  & 99.04  & 94.38  & 98.36  & 96.33  & 98.67  & 97.36  & 97.87  & 97.61  \\
    \textbf{[3, 5, 7]} & 98.56  & 91.76  & 94.89  & 93.30  & 99.08  & 94.42  & 98.61  & 96.46  & 98.87  & 97.45  & 98.52  & 97.98  \\
    \textbf{[1, 3, 5, 7]} & 98.62  & 91.78  & 95.42  & 93.56  & 99.08  & 94.43  & 98.61  & 96.48  & 98.92  & 97.43  & 98.71  & 98.06  \\
    \bottomrule
    \end{tabular}%
    }
  \label{tab:addlabel6}%
\end{table}%


\fi

\end{document}